\PassOptionsToPackage{dvipsnames,table}{xcolor}
\documentclass[]{./meta_ai/fairmeta}

\usepackage{microtype}
\usepackage{graphicx}
\usepackage{subcaption}
\usepackage{booktabs}

\usepackage{xspace}
\usepackage{makecell}
\usepackage{nicefrac}
\usepackage{tikz-cd}
\usepackage[framemethod=TikZ]{mdframed}
\usepackage{soul}
\usepackage{multirow}
\usepackage{scalerel}
\usepackage{algorithm}
\usepackage{algpseudocode}
\usepackage{caption}
\usepackage{wrapfig}
\usepackage[titletoc,page]{appendix}
\usepackage{titletoc}
\usepackage{tabularx}

\newmdenv[backgroundcolor=metabg, roundcorner=5pt, skipabove=7pt, linewidth=0pt, innertopmargin=4pt]{metaframe}

\usepackage{amsmath,amsfonts,bm,amsthm,amssymb,amsfonts}

\usepackage{jmath}
\usepackage{xspace}
\usepackage{calc}
\usepackage{accents}
\usepackage{xparse,xpatch}
\usepackage{thmtools}

\usepackage[colorinlistoftodos,disable]{todonotes}

\newcommand{\remi}[1]{\todo[inline,color=orange!10,caption={}]{\scriptsize{\textbf{RM: }#1}}}

\newcommand{\vtdflong}{TD-Conditional Flow Matching\xspace}
\newcommand{\vtdfshort}{\textsc{td-cfm}\xspace}
\newcommand{\ctdflong}{Coupled TD-Conditional Flow Matching\xspace}
\newcommand{\ctdfshort}{\textsc{td-cfm(c)}\xspace}
\newcommand{\tdflong}{TD${}^2$-Conditional Flow Matching\xspace}
\newcommand{\tdfshort}{\textsc{td${}^2$-cfm}\xspace}

\newcommand{\vtddshort}{\textsc{td-dd}\xspace}

\newcommand{\tddshort}{\textsc{td${}^2$-dd}\xspace}

\newcommand{\tdganshort}{\textsc{gan}\xspace}
\newcommand{\tdvaeshort}{\textsc{vae}\xspace}

\newcommand{\onestepterm}[1]{\accentset{\rightarrow}{#1}}
\newcommand{\bootterm}[1]{\accentset{\curvearrowright}{#1}}

\NewDocumentCommand{\fixlimits}{e{^_}}{%
  \IfValueT{#1}{^{\mkern4mu#1}}%
  \IfValueT{#2}{_{#2}}%
}
\xapptocmd{\bootterm}{\fixlimits}{}{}
\xapptocmd{\onestepterm}{\fixlimits}{}{}

\newcommand{\deq}{\mathrel{\mathop{:}}=} 
\newcommand{\ghmdist}{m} 
\newcommand{\ghmdistbar}{\widetilde{m}} 
\newcommand{\ghmprobpath}{m}
\newcommand{\ghmvecfield}{\tilde{v}}

\newcommand{\sm}{m}
\newcommand{\Tpi}[1]{\left(\mathcal{T}^\pi #1\right)}

\newcommand{\Ppi}[1]{\left(P^\pi #1\right)}

\newcommand{\setfont}[1]{\mathsf{#1}}

\newcommand{\scorefn}{s}

\newtheorem{lemma}{Lemma}
\newtheorem{definition}{Definition}
\newtheorem{theorem}{Theorem}
\newtheorem{proposition}{Proposition}

\newcommand{\owntag}[1]{\stepcounter{equation}\tag{#1\theequation}}

\definecolor{hl}{RGB}{205, 232, 248}
\definecolor{best}{RGB}{199, 221, 236}

\newcommand{\mhl}[1]{\sethlcolor{hl}\hl{#1}}

\usepackage{etoolbox}
\makeatletter
\patchcmd{\hyper@makecurrent}{%
    \ifx\Hy@param\Hy@chapterstring
        \let\Hy@param\Hy@chapapp
    \fi
}{%
    \iftoggle{inappendix}{%
        \@checkappendixparam{chapter}%
        \@checkappendixparam{section}%
        \@checkappendixparam{subsection}%
        \@checkappendixparam{subsubsection}%
        \@checkappendixparam{paragraph}%
        \@checkappendixparam{subparagraph}%
    }{}%
}{}{\errmessage{failed to patch}}

\newcommand*{\@checkappendixparam}[1]{%
    \def\@checkappendixparamtmp{#1}%
    \ifx\Hy@param\@checkappendixparamtmp
        \let\Hy@param\Hy@appendixstring
    \fi
}
\makeatletter

\newtoggle{inappendix}
\togglefalse{inappendix}

\apptocmd{\appendix}{\toggletrue{inappendix}}{}{\errmessage{failed to patch}}
\apptocmd{\subappendices}{\toggletrue{inappendix}}{}{\errmessage{failed to patch}}

\makeatletter
\renewcommand{\sectionautorefname}{\protect\textsection\@gobble}
\renewcommand{\subsectionautorefname}{\protect\textsection\@gobble}
\renewcommand{\subsubsectionautorefname}{\protect\textsection\@gobble}
\makeatother

\makeatletter
\addto\extrasenglish{%
  \renewcommand{\sectionautorefname}{\protect\textsection\@gobble}%
  \renewcommand{\subsectionautorefname}{\protect\textsection\@gobble}%
  \renewcommand{\subsubsectionautorefname}{\protect\textsection\@gobble}%
}
\makeatother

\def\Vhrulefill{\leavevmode\leaders\hrule height 0.7ex depth \dimexpr0.4pt-0.7ex\hfill\kern0pt}

\title{Temporal Difference Flows}

\author[2,3,*]{Jesse Farebrother}
\author[1]{Matteo Pirotta}
\author[1]{Andrea Tirinzoni}
\author[1]{R\'emi Munos}
\author[1]{Alessandro Lazaric}
\author[1]{\\Ahmed Touati}

\affiliation[1]{FAIR at Meta}
\affiliation[2]{Mila -- Qu\'ebec AI Institute}
\affiliation[3]{McGill University}

\contribution[*]{Work done at Meta}

\abstract{Predictive models of the future are fundamental for an agent's ability to reason and plan.
A common strategy learns a world model and unrolls it step-by-step at inference, where small errors can rapidly compound.
Geometric Horizon Models (GHMs) offer a compelling alternative by directly making predictions of future states, avoiding cumulative inference errors.
While GHMs can be conveniently learned by a generative analog to temporal difference (TD) learning, existing methods are negatively affected by bootstrapping predictions at train time and struggle to generate high-quality predictions at long horizons.
This paper introduces Temporal Difference Flows (TD-Flow), which leverages the structure of a novel Bellman equation on probability paths alongside flow-matching techniques to learn accurate GHMs at over $5\times$ the horizon length of prior methods.
Theoretically, we establish a new convergence result and primarily attribute TD-Flow's efficacy to reduced gradient variance during training. We further show that similar arguments can be extended to diffusion-based methods. 
Empirically, we validate TD-Flow across a diverse set of domains on both generative metrics and downstream tasks including policy evaluation.
Moreover, integrating TD-Flow with recent behavior foundation models for planning over pre-trained policies demonstrates substantial performance gains, underscoring its promise for long-horizon decision-making.
}

\begin{document}

\maketitle

\section{Introduction}\label{sec:intro}
Predictive modeling lies at the heart of intelligent decision-making, enabling agents to reason and plan in complex environments.
In Reinforcement Learning (RL), this predictive capability has traditionally been achieved through world models that capture the transition structure of the environment.
These models have enabled significant advances across numerous domains --- from robotics manipulation employing model-predictive control \citep{sikchi21learning,hafner23mastering,hansen22tdmpc,hansen24tdmpc2}, to sample-efficient exploration strategies \citep{schmidhuber91apossibility,stadie16incentivizing,pathak17curiosity}, and sophisticated planning algorithms \citep{silver16mastering,silver17go,schrittwieser20mastering}.
However, while world models have demonstrated impressive results, they face fundamental limitations when deployed for long-horizon reasoning.
The standard approach of unrolling predictions step-by-step leads to compounding errors, as small inaccuracies in each prediction accumulate and propagate forward in time \citep{talvitie14model,jafferjee20dyna,lambert22error}.
This ``curse of horizon'' presents a significant challenge for applications requiring 
reliable~long-range~predictions.

An alternative approach is to learn a generative model of future states directly, avoiding compounding errors during inference.
These models, usually referred to as Geometric Horizon Models~\citep[GHM;][]{thakoor22ghm} or $\gamma$-models~\citep{janner20gmodel}, are learned by leveraging the temporal difference structure of the successor measure \citep{blier21successor}.
However, their reliance on bootstrapped predictions during training can lead to instability and growing inaccuracy over long horizons.
As a result, current methods struggle to make accurate predictions beyond $20$-$50$ steps, also limiting their utility for long-term decision-making.
In this paper, we show that while state-of-the-art generative methods like flow matching \citep{lipman2022flow} and denoising diffusion \citep{ho20ddpm} cannot be directly applied to learn long-horizon GHMs, their iterative nature can be leveraged to better exploit the temporal difference structure of the problem.
This insight yields a new class of methods that provably converges to the successor measure while reducing the variance of their sample-based gradient estimates, enabling stable long-horizon predictions.
Empirically, our approach produces significantly more accurate GHMs at all horizons, consistently outperforming state-of-the-art algorithms across domains and metrics, including prediction accuracy, value function estimation, and generalized~policy~improvement.

\section{Background}\label{sec:background}
In the following, we use capital letters to denote random variables, sans-serif fonts for sets (e.g., $\setfont{A}$), and $\mathscr{P}(\setfont{A})$ to denote the space of probability measures over a measurable set $\setfont{A}$.

\textbf{Markov Decision Process}\,
We consider a reward-free discounted Markov decision process $\mathcal{M} = \left( \setfont{S}, \setfont{A}, P, \gamma \right)$, which characterizes the dynamics of a sequential decision-making problem.
At each step, the agent selects an action $a \in \setfont{A}$ in state $s \in \setfont{S}$ according to its policy $\pi : \setfont{S} \to \setfont{A}$.
This action influences the transition to the next state $s' \in \setfont{S}$, governed by the transition kernel $P : \setfont{S} \times \setfont{A} \to \mathscr{P}(\setfont{S})$, which defines a probability measure over successor states.
The discount factor $\gamma \in [0, 1)$ can be interpreted as implying a process that either continues with probability $\gamma$ or terminates with probability $1 - \gamma$.
This interpretation naturally defines a geometric distribution of future states the agent will occupy, where states reached after $k$ steps are discounted by $\gamma^k$.

\textbf{Successor Measure}\,
The normalized \emph{successor measure}~\citep{dayan93sr,blier21successor} of a policy $\pi$ describes the discounted distribution of future states visited by $\pi$ starting from an initial state-action pair $(s, a)$.
For the measurable subset $\setfont{X} \subseteq \setfont{S}$ the successor measure $\sm^\pi(\setfont{X} \mid s, a)$ represents the probability that future states fall within $\setfont{X}$, geometrically discounted by $\gamma$ according to the time of visitation. Formally, it is defined as:
\begin{equation*}%
\sm^\pi(\,\setfont{X} \mid s,a\,)= (1 - \gamma) \sum_{k=0}^\infty \gamma^k\, \mathrm{Pr}(S_{k+1} \in \setfont{X} \mid S_0 = s,\,A_0 = a,\, \pi),
\end{equation*}
where $\Pr(\cdot\!\mid S_0,A_0,\pi)$ denotes the probability of state-action sequences $(S_k,A_k)_{k
\geq 0}$ generated from $(S_0, A_0)$ following $S_k \sim P(\cdot\!\mid\!S_{k-1},A_{k-1})$ and
$A_k = \pi(S_k)$.
The successor measure encapsulates the long-term dynamics of $\pi$, enabling value estimation for any reward function $r : \setfont{S} \to \mathbb{R}$.
Specifically, the value of taking action $a \in \setfont{A}$ in state $s \in \setfont{S}$ is the expected reward under states visited by $\pi$ amplified by the effective horizon $(1 - \gamma)^{-1}$:
\begin{equation}\label{eq:sm-vf}
Q^\pi(s, a) = (1 - \gamma)^{-1} \E_{X \sim \sm^\pi(\cdot\mid s, a)}{r(X)} \, .
\end{equation}
Moreover, $m^\pi$ is the fixed point of the Bellman operator ${\cal T}^\pi : \mathscr{P}(\setfont{S})^{\setfont{S}\times\setfont{A}} \to \mathscr{P}(\setfont{S})^{\setfont{S}\times\setfont{A}}$~\citep{thakoor22ghm}:
\begin{align}\label{eq:bellman}
    \sm^\pi(\cdot \mid s, a) &= \Tpi{m^\pi}(\cdot \mid s, a) \\
    &\deq (1-\gamma) P(\cdot \mid s, a) + \gamma \Ppi{m^\pi}(\cdot \mid s, a)\nonumber \, .
\end{align}
The operator $P^\pi$ applied to $\sm$ mixes the one-step kernel with the successor measure, accounting for transitioning from $(s, a)$ to a new state-action pair $(s', \pi(s'))$ and querying the successor measure $\sm(\cdot\mid s,' \pi(s'))$ 
\remi{$\sm(\cdot\mid s,' \pi(s'))$ instead of $\sm^\pi(\cdot\mid s,' \pi(s'))$?}
thereafter:
\begin{equation*}
    \Ppi{\sm}( \mathrm{d}x \mid s, a)
    = \int_{s'} P(\mathrm{d}s' \mid s, a) \, \sm(\mathrm{d}x \mid s', \pi(s'))\,.
\end{equation*}

\textbf{Geometric Horizon Model}\,
A \emph{Geometric Horizon Model} \citep[GHM;][]{thakoor22ghm} or $\gamma$-model \citep{janner20gmodel} is a generative model of the \emph{normalized} successor measure.
To learn the parametric model $\ghmdistbar(\cdots;\theta) \approx \sm^\pi$ we can minimize a Monte-Carlo cross-entropy objective over source states from the empirical~distribution~$\rho$~as,
$$
\argmin_\theta \,\, 
\E_{S \sim \rho,\,X \sim \sm^{\pi}(\cdot\,|\,S, \pi(S))}{-\log{ \ghmdistbar(X\,|\,S,A;\theta))}} \,.
$$
In order to sample from $m^\pi$ we deploy policy $\pi$ for $t \sim \text{Geometric}(1 - \gamma)$ steps resulting in state $X = S_{t}$.
Akin to other Monte-Carlo methods in RL, this approach is problematic when learning from off-policy data, often leading to high-variance estimators that rely on importance sampling \citep{precup01offpolicy}.

Alternatively, we can leverage the Bellman equation~\eqref{eq:bellman} to construct an off-policy iterative method for estimating $\sm^\pi$.
Given initial weights $\theta^{(0)}$, each iteration updates $\theta$ by minimizing the following temporal-difference cross-entropy objective over transitions that need not come from policy $\pi$,%
\begin{align}%
   \theta^{\scaleto{(n+1)\mathstrut}{6.5pt}} =\,
    \argmin_{\theta}\, 
    \E_{(S, A) \sim \rho,\,X \sim \Tpi{\ghmdistbar^{(n)}}(\cdot \mid S, A)}{-\log \ghmdistbar(X \mid S, A;\theta)}.\label{eq:iterative.td}
\end{align}
In the equation above and throughout the paper, we adopt the shorthand $\ghmdistbar^{(n)} = \ghmdistbar(\cdots; \theta^{(n)})$. To generate samples $X \sim \Tpi{\ghmdistbar^{(n)}}(\cdot\mid S,A)$ we first draw a successor state $S' \sim P(\cdot\mid S,A)$; then with probability $1 - \gamma$, we return $S'$; otherwise, with probability $\gamma$, we return a \emph{bootstrapped sample} drawn from $\ghmdistbar^{(n)}(\cdot\mid S', \pi(S'))$.

Several probabilistic models have been applied to this problem, including generative adversarial networks \citep[e.g.,][]{janner20gmodel,wiltzer24dsm}, normalizing flows~\citep[e.g.,][]{janner20gmodel}, and variational auto-encoders~\citep[e.g.,][]{thakoor22ghm,tomar2024video}. %
We now turn our attention to a class of generative models based on the flow-matching framework specifically designed to leverage the underlying structure of the Bellman equation \eqref{eq:bellman}, enabling more effective generative models of the successor measure.

\begin{figure*}
    \centering
    \includegraphics[width=0.9855\linewidth]{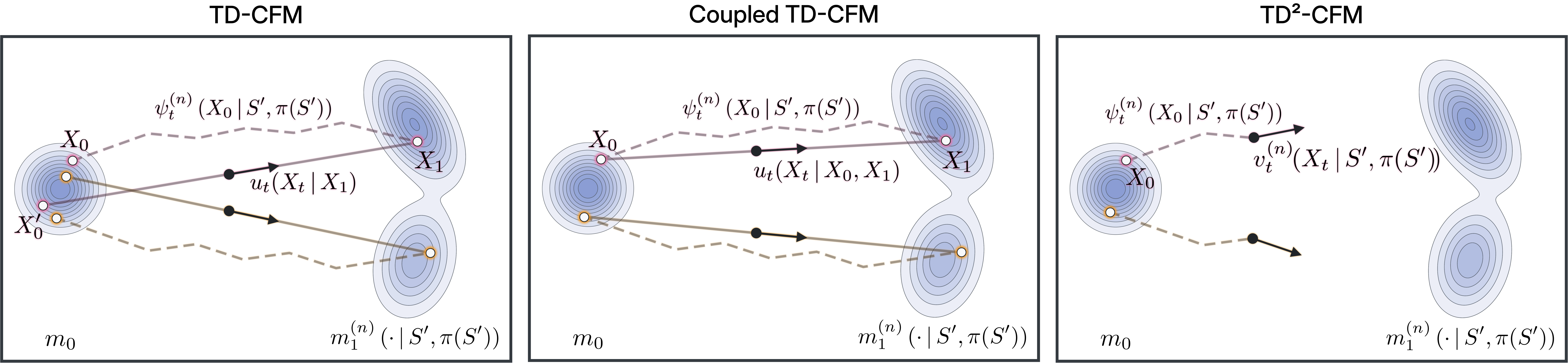}
    \caption{
    Visual depiction of TD-Flow variants. Samples are mapped from $m_0$ to the target distribution $m_1^{(n)}$ through the neural ODE $\psi^{(n)}_t$. Dashed lines depict the neural ODE trajectory; solid lines show the conditional probability path $u_t$. \textbf{(Left)} \vtdfshort maps $X_0$ to $X_1$ before creating a separate conditional path between $X'_0$ and $X_1$, resulting in crossing paths. \textbf{(Middle)} \ctdfshort directly couples $X_0$ used to generate $X_1$ when constructing the conditional probability path. \textbf{(Right)} \tdfshort solves the neural ODE up to time $t$ to directly obtain the target velocity $\ghmvecfield_t$.
    }
    \label{fig:methods}
\end{figure*}
\section{Temporal Difference Flows}\label{sec:td-flow}
Flow Matching \citep[FM;][]{lipman2022flow,lipman2024flow,liu2022flow,albergo23building} constructs a time-dependent probability path $\ghmprobpath_t : \setfont{S} \times \setfont{A} \to \mathscr{P}(\setfont{S})$ for $t\in [0,1]$ that evolves smoothly from the source distribution $m_0 = p_0 \in \mathscr{P}(\setfont{S})$ to the target distribution $\ghmprobpath_1 \approx \sm^\pi$.
This evolution is governed by a vector field $v_t : \setfont{S} \times \setfont{S} \times \setfont{A} \to \setfont{S}$, which dictates the instantaneous movement of samples along~$m_t$.
The relationship between this vector field and the resulting probability path is established through a time-dependent flow $\psi_t : \setfont{S} \times \setfont{S} \times \setfont{A} \to \setfont{S}$, defined as the solution to the following ordinary differential equation initial value problem:
$$
\frac{\mathrm{d}}{\mathrm{d}t} \psi_t(x\mid s,a) = v_t\big(\psi_t(x\mid s, a) \mid s, a\big),\,\, \psi_0(x\mid s, a) = x
\,\,\Longleftrightarrow\,\,
\psi_t(x \,|\,s, a) = x + \int_0^t v_\tau\big(\psi_\tau(x|s,a)\,|\,s, a\big)\, \mathrm{d}\tau \,.
$$
We say that $v_t$ generates $m_t$ if its flow $\psi_t$ satisfies $X_t := \psi_t(X_0 \mid S, A) \sim m_t(\cdot \mid S, A)$ for $X_0 \sim m_0$. In words, the flow $\psi_t$ pushes samples forward through time, ensuring they are distributed according to $m_t$ at time $t$.
To learn this transformation, we can minimize the squared $L^2$ distance between a parameterized vector field $\ghmvecfield_t(\cdots;\theta)$ and the true vector field $v_t$ over $t \sim \mathcal{U}([0, 1])$, yielding the  Monte-Carlo Flow Matching (\textsc{mc-fm})~loss,
\begin{equation}
\begin{gathered}
\ell_{\mathrm{\textsc{mc-fm}}}(\theta) = \mathbb{E}_{\rho,t,X_t}
\Big[
\big\|
\ghmvecfield_t(X_t \,|\, S, A;\theta) - v_t(X_t\,|\,S,A)
\big{\|}^2
\Big]\,, \\
\text{where}\,X_t \sim \sm_t(\cdot\,|\,S, A) \,.
\end{gathered}
\owntag{\textsc{mc-fm}; }\label{eq:fm-marginal}
\end{equation}
Despite its conceptual simplicity, direct optimization of the flow matching objective above proves challenging due to the inaccessibility of the true probability path $m_t$ and its associated vector field $v_t$.

Alternatively, \citet{lipman2022flow} shows that we can sidestep this problem entirely by introducing additional conditioning information. Instead of directly modeling the probability path $m_t$ we can introduce a random variable $Z$ and define a \emph{conditional path} on $Z$ as $p_{t\mid Z} : \setfont{S} \times \setfont{Z} \to \mathscr{P}(\setfont{S})$ 
\citep{lipman2024flow,tong24improving}.
The conditional velocity field $u_{t\mid Z} : \setfont{S} \times \setfont{Z} \to \setfont{S}$ that generates $p_{t\mid Z}$ can now be computed in closed form for many simple choices of $Z$ and $p_{t\mid Z}$.
One such choice is taking $Z = X_1$ and performing a linear Gaussian interpolation from $X_0 \to X_1$ resulting in $p_{t\mid 1}(\cdot \,|\, X_1) = \mathcal{N}(\cdot \mid t X_1,\, (1 - t)^2 I)$ with the corresponding vector field given by $u_{t\mid 1}(x\,|\,X_1) = (X_1 - x) / (1 - t)$.
Armed with the ability to sample from $p_{t\mid 1}$ and to compute $u_{t\mid 1}$, we can directly learn $\ghmvecfield_t$ by optimizing the Monte-Carlo \emph{Conditional} Flow Matching (\textsc{mc-cfm}) objective:
\begin{equation}
\begin{gathered}
\ell_{\mathrm{\textsc{mc-cfm}}}(\theta) = \mathbb{E}_{\rho,t,Z,X_t}
\Big[
\big\|
\ghmvecfield_t(X_t\mid S,A; \theta) - u_{t\mid Z}(X_t \mid Z)
\big{\|}^2
\Big] \,, \\
\text{where}\,Z = X_1 \sim \sm^\pi(\cdot\mid S, A) \,,
X_t \sim p_{t\mid Z}(\cdot\mid Z) \, .
\end{gathered}
\owntag{\textsc{mc-cfm}; }\label{eq:mc-cfm}
\end{equation}
Remarkably, both \eqref{eq:fm-marginal} and \eqref{eq:mc-cfm} share the same gradient and converge to the same solution.

\begin{metaframe}
\begin{proposition}[\citealt{lipman2024flow}]\label{prop:fm.cfm.equivalence}
Given a conditional probability path $p_{t|Z}$ and vector field $u_{t\mid Z}$ with their associated marginal counterparts $p_t(x)$ and $v_t(x)$, we have
\begin{align*}
\nabla_\theta\, \ell_{\mathrm{\textsc{mc-fm}}}(\theta) = \nabla_\theta\, \ell_{\mathrm{\textsc{mc-cfm}}}(\theta).
\end{align*}
\end{proposition}
\end{metaframe}

\textbf{TD-CFM}\, 
While \eqref{eq:mc-cfm} requires direct access to samples from the target distribution $m^\pi$, we can instead learn from an offline dataset $\rho$ containing only one-step transitions $(S, A, S')$ through an iterative process similar to \eqref{eq:iterative.td}.
Starting with initial parameters $\theta^{(0)}$, at each iteration, we minimize the \vtdflong (\vtdfshort) loss $\ell_{\mathrm{\vtdfshort}}$ --- an extension of \eqref{eq:mc-cfm} that differs only in its sampling procedure:
\begin{equation}
\begin{gathered}
X_0 \sim p_0 \\ Z = X_1 \sim (1 - \gamma)\, \delta_{S'} + \gamma\, \delta_{ \widetilde{\psi}^{(n)}_1\left(X_0 \,|\, S', \pi(S')\right)
} \,.
\end{gathered}
\owntag{\vtdfshort; }\label{eq:td-cfm}
\end{equation}
In this procedure, with probability $1 - \gamma$, we return the successor state $S'$. Otherwise, with probability $\gamma$ we sample from the neural ordinary differential equation \citep{chen18neuralode} $\widetilde{\psi}^{(n)}_t$ with corresponding vector field $\ghmvecfield^{(n)}_t\left(X_t\mid S',\pi(S')\right)$ from $X_0 \sim p_0$ to produce a sample $X_1 \sim \ghmdistbar^{(n)}(\cdot\mid S',\pi(S'))$.

\textbf{Coupled TD-CFM}\, Although \eqref{eq:td-cfm} offers a principled way of learning the flow from noise to data, an increasingly popular strategy to improve flow matching methods is to correlate noise and data whenever a ``natural'' coupling is available~\citep[e.g.,][]{liu2022flow,shi23dsbm,pooladian23multisample,tong24improving,valentin24sbf}.
Motivated by this idea, we observe that the process used to generate $X_1$ described above already provides a direct coupling between $X_0$ and $X_1$.
We can leverage this coupling by conditioning the probability path $p_{t\mid Z}$ on both endpoints, i.e., $Z = (X_0, X_1)$, rather than just conditioning on $Z = X_1$ as in \vtdfshort.
As illustrated in \autoref{fig:methods}, this coupling helps align $X_t$ with the path generated by $\widetilde{\psi}^{(n)}_t$, potentially simplifying the regression problem.
This procedure gives rise to the \ctdflong (\ctdfshort) loss $\ell_{\mathrm{\ctdfshort}}$ which now extends $\ell_{\mathrm{\vtdfshort}}$, again, differing only in its sampling procedure:
\begin{equation}
\begin{gathered}
X_0 \sim p_0\\
X_1 \sim (1 - \gamma)\, \delta_{S'} + \gamma\, \delta_{\widetilde{\psi}^{(n)}_1 \left(X_0\mid S', \pi(S')\right)} \\
Z = (X_0, X_1) \, .
\end{gathered}
\owntag{\ctdfshort; }\label{eq:td-cfm(c)} \\
\end{equation}
A convenient approach to specifying the conditional path $p_{t\mid Z}$ is to define 
$X_t = \phi_t(X_0, X_1) = \alpha_t X_1 + \beta_t X_0$ as the affine interpolant between $X_0$ and $X_1$, with the interpolation coefficients satisfying the boundary conditions $\alpha_0 = \beta_1 = 0$, $\alpha_1 = \beta_0 = 1$, and monotonicity constraints $\dot{\alpha}_t>0, -\dot{\beta}_t>0$, where the over-dot denotes the time derivative. From this definition, the conditional vector field arises as the time derivative of this interpolant defined as $u_{t\mid 0,1}(X_t \mid X_0, X_1) = \dot\phi_t(X_0, X_1) = \dot\alpha_t X_1 + \dot\beta_t X_0$.
A simple choice of the interpolation coefficients that yields a linear (straight-line) conditional path is given by $\beta_t = 1 - \alpha_t = 1 - t$.

\textbf{TD$\mathbf{{}^2}$-CFM}\, While \eqref{eq:td-cfm(c)} improves upon \eqref{eq:td-cfm} by accounting for the coupling between bootstrapped samples and their generating noise, both methods rely upon fitting an ad-hoc conditional vector field $u_{t\mid Z}$ that generates the surrogate conditional path $p_{t \mid Z}$.
To formulate a more structured approach, we exploit the linearity of the Bellman equation, as detailed in the following result.
\begin{metaframe}
\begin{restatable}{lemma}{lemvfonestepbootstrap}\label{lem:vf-onestep-bootstrap}
Let $\onestepterm{p}_t$ be a probability path for $P$ generated by vector field $\onestepterm{v}_t$ and $\bootterm{p}_t^{(n)}$ be a probability path for $P^\pi m_1^{(n)}$ generated by $\bootterm{v}_t^{(n)}$  such that $\onestepterm{p}_0 = \bootterm{p}_0^{(n)} = m_0$. For any $t\in[0,1]$ and $(s,a)$ let\,\footnotemark%
\begin{align*}
    v_t^{(n+1)}(\cdot \mid s, a)
    =
    \argmin_{v\,:\, \mathbb{R}^d\rightarrow \mathbb{R}^d} (1-\gamma) 
    &\mathbb{E}_{\onestepterm{X}_t \sim \onestepterm{p}_t(\cdot | s, a)}\Big[\big \| v(\onestepterm{X}_t) - \onestepterm{v}_t(\onestepterm{X}_t \mid s, a)\big \|^2\Big] \\
    +\,\, \gamma 
    &\mathbb{E}_{\bootterm{X}_t \sim \bootterm{p}_t^{(n)}(\cdot | s,a)}\Big[\big \| v(\bootterm{X}_t) - \bootterm{v}_t^{(n)}(\bootterm{X}_t \mid s,a)\big \|^2\Big].
\end{align*}
Then $v_t^{(n+1)}$ induces a probability path $m_t^{(n+1)}$ such that $m_0^{(n+1)} = m_0$ and $m_1^{(n+1)} = \mathcal{T}^\pi m_1^{(n)}$.
\end{restatable}
\end{metaframe}
\footnotetext{Notice here that the minimization is over the space of all functions and not the parameterized vector fields $\tilde{v}_t(\cdots;\theta)$.}
This result shows that it is possible to use two independent probability paths for the two terms in the sampling process induced by the Bellman operator. For the first term, we can use a standard CFM approach for $Z = X_1$ with conditional path $\onestepterm{p}_{t\mid 1}$ and vector field $\onestepterm{u}_{t\mid 1}$, which induces the marginal,
\begin{align*}
\onestepterm{v}_t(x\mid s,a) = \int  \onestepterm{u}_{t\mid 1}(x \mid x_1) \frac{\onestepterm{p}_{t\mid 1}(x \mid x_1) P(\mathrm{d}x_1\mid s,a)}{ \onestepterm{p}_t(x\mid s,a)},
\end{align*}
where $\onestepterm{p}_t(x|s,a) = \int \onestepterm{p}_{t|1}(x|s')P(\mathrm{d}s'|s,a)$.
For the second term, we can leverage the GHM $m_t^{(n)}$ learned at the previous iteration to construct the marginal,
\begin{align*}
\bootterm{v}_t^{(n)}(x\mid s,a) = \int v_{t}^{(n)}(x \mid s',a') \frac{m_{t}^{(n)}(x \mid s',a') P(\mathrm{d}s'\mid s,a)}{ \bootterm{p}_t^{(n)}(x\mid s,a)},
\end{align*}
where $\bootterm{p}_t^{(n)}(x\mid s,a) = \int m_t^{(n)}(x\mid s',a')P(\mathrm{d}s'\mid s,a)$, and $a'=\pi(s')$. This shows that $m_t^{(n)}$ plays the role of a conditional probability path for the bootstrapping term and $v_t^{(n)}$ is its associated conditional vector field. We can then use the equivalence between \textsc{fm} and \textsc{cfm} in \autoref{prop:fm.cfm.equivalence} to replace the marginal probability paths and vector fields in \autoref{lem:vf-onestep-bootstrap} with their conditional counterparts to obtain the loss:
\begin{gather}
\onestepterm{\ell}(\theta)
=
\mathbb{E}_{\rho,t,Z,\onestepterm{X}_t}
\Big[
\big\|
\ghmvecfield_t(\onestepterm{X}_t\mid S, A; \theta) - \onestepterm{u}_{t\mid Z}(\onestepterm{X}_t\mid Z)
\big{\|}^2
\Big]\,, \tag*{} \\
\text{where}\, Z = X_1 \sim P(\cdot \mid S, A),\,
\onestepterm{X}_t \sim \onestepterm{p}_{t\mid Z}(\cdot\,|\,Z) \,, \tag*{} \\[0.5\baselineskip]
\bootterm{\ell}(\theta)
=
\mathbb{E}_{\rho,t,\bootterm{X}_t}
\Big[ 
\big{\|}
\ghmvecfield_t(\bootterm{X}_t\mid S, A; \theta) - \ghmvecfield^{(n)}_t(\bootterm{X}_t\mid S',\pi(S')
\big{\|}^2
\Big]\,, \tag*{} \\
\text{where}\, X_0 \sim p_0,\, S'\sim P(\cdot\mid S,A),\,
\bootterm{X}_t = \widetilde{\psi}_t^{(n)}(X_0\mid S', \pi(S')) \,, \tag*{} \\[0.5\baselineskip]
\ell_{\mathrm{\tdfshort}}(\theta) = (1 - \gamma)
\onestepterm{\ell}(\theta)
+ \gamma
\bootterm{\ell}(\theta) \,.
\owntag{\tdfshort; }\label{eq:td2fm}
\end{gather}
Since we now bootstrap the previous estimate not only in the sampling process but also in the objective function, we refer to this method as \tdflong (\tdfshort).
The right panel of \autoref{fig:methods} depicts the process of obtaining the bootstrapped vector field $\ghmvecfield_t^{(n)}$ for \tdfshort.
We provide further implementation details and pseudo-code for all the aforementioned TD-Flow methods in \autoref{app:algorithms}.
Next, we extend our $\textsc{td}{}^2$ result to the class of denoising diffusion models.

\subsection{Extension to Diffusion Models}\label{ssec:diffusion}
Denoising Diffusion models \citep{sohl2015deep,ho20ddpm} build a diffusion process starting from a data sample $X_0 \sim q_0 = m^\pi(\cdot\mid S, A)$\footnote{Different to flow matching, time is inverted in diffusion models and ranges from 0 to $T$.} and corrupting it via a stochastic differential equation (SDE),
\begin{equation}
    \mathrm{d}X_t = f(t)\, X_t\, \mathrm{d}t + g(t)\, \mathrm{d}W_t\,,
    \label{eq:forward.sde}
\end{equation}
where $t \in [0, T]$ for some time horizon $T$, $f,g: [0, T] \rightarrow \R$ is drift and diffusion term, and $W_t \in \R^d$ is a standard Brownian motion. The forward process of the linear SDE~\eqref{eq:forward.sde} has an analytic Gaussian kernel $q_{t\mid 0}(\cdot \mid X_0) = \mathcal{N}(\cdot \mid \alpha_t X_0, \sigma_t^2 I)$, where $\alpha_t$ and $\sigma_t$ can be computed in closed form.
To sample from the target data distribution $q_0$, we can solve the reverse SDE~\citep{song2019generative} from time $T$ to $0$:
\begin{equation}
   \mathrm{d}X_t = \Big(f(t)\, X_t - g(t) \,\nabla_{X_t} \log q_t(X_t \mid S, A) \Big)\mathrm{d}t + g(t)\, \mathrm{d}\widebar{W}_t \,,
   \label{eq:reverse.sde}
\end{equation}
where $\widebar{W}_t$ is the reverse-time Brownian motion and $q_t$ is the marginal distribution of both the forward \eqref{eq:forward-SDE} and reverse \eqref{eq:reverse-SDE} process. 
To simulate~\eqref{eq:reverse.sde}, we can train a parametrized score function $\tilde{\scorefn}_t(x \mid s, a;\theta)$ to approximate $\nabla_{x_t} \log q_t(x_t \mid s, a)$ using the denoising diffusion / score matching objective~\citep{vincent2011connection}: 
\begin{gather}
\ell_{\mathrm{\textsc{dd}}}(\theta) = \mathbb{E}_{\rho,t,X_0, X_t}
\Big[
\big\|
\tilde{\scorefn}_t(X_t \mid S, A; \theta) - \nabla_{X_t} \log q_{t|0}(X_t \mid X_0)
\big{\|}^2
\Big]\,, \tag*{} \\
\text{where } X_0 \sim m^\pi(\cdot \mid S, A),\, X_t \sim q_{t\mid 0}(\cdot \mid X_0) \,.
\owntag{\textsc{dd}; } \label{eq:dd}
\end{gather}
\textbf{Temporal Difference Diffusion}\,
Following the blueprint in \autoref{sec:td-flow}, we define an iterative process starting from $\tilde{\scorefn}^{(0)} = \tilde{\scorefn}(\cdots; \theta^{(0)})$ and minimize at each iteration the Temporal-Difference Denoising Diffusion ({\vtddshort}) loss:
 \begin{equation}
 \begin{gathered}
\ell_{\mathrm{\vtddshort}}(\theta) = \mathbb{E}_{\rho,t,X_0,X_t}
\Big[
\big\| \tilde{\scorefn}(X_t\mid S,A; \theta) - \nabla_x \log q_{t\mid 0}(X_t \mid X_0)
\big{\|}^2
\Big]\,, \\
\text{where}\, X_0 \sim \Tpi{\ghmdistbar_{0\mid T}^{(n)}}(\cdot\mid S,A), X_t \sim q_{t|0}(\cdot\mid X_0) \, .
\owntag{\textsc{td-dd}; }\label{eq:td-dd}
\end{gathered}
\end{equation}
Once again, to sample $X_0 \sim \Tpi{\ghmdistbar^{(n)}_{0|T}}(\cdot\mid S,A)$, we proceed as follows: with probability $1 - \gamma$, we draw a successor state $S' \sim P(\cdot\mid S, A)$; conversely, with probability $\gamma$, we sample from the bootstrapped model by solving the reverse SDE with score function $\tilde{\scorefn}^{(n)}$, initiated from $X_T$.
Following an approach analogous to \autoref{lem:vf-onestep-bootstrap}, we demonstrate in \autoref{app:diffusion} that we can employ two distinct diffusion processes for the two terms involved in the Bellman operator, which consequently leads to the {\tddshort} objective:
\begin{gather}
\onestepterm{\ell}(\theta)
= \mathbb{E}_{\rho,t,\onestepterm{X}_t}
\Big[
\big\|
\tilde{\scorefn}_t(\onestepterm{X}_t\mid S, A; \theta) - \nabla_{\onestepterm{X}_t} q_{t\mid 0}(\onestepterm{X}_t\mid S')
\big{\|}^2
\Big] \, ,
\tag*{} \\
\text{where}\, \onestepterm{X}_t \sim q_{t\mid 0}(\cdot\mid S')\,, \tag*{} \\[0.5\baselineskip]
\bootterm{\ell}(\theta)
=
\mathbb{E}_{\rho,t,\bootterm{X}_t}
\Big[
\big\|
\tilde{\scorefn}_t(\bootterm{X}_t\mid S, A; \theta) - \tilde{\scorefn}^{(n)}_t(\bootterm{X}_t\mid S',\pi(S')
\big{\|}^2
\Big]\,, \tag*{} \\
\text{where}\, X_T \sim q_T,\, \bootterm{X}_t \sim q^{(n)}_{t\mid T}(\cdot\mid S', \pi(S'))\,, \tag*{} \\[0.5\baselineskip]
\ell_{\mathrm{\tddshort}}(\theta) = (1 - \gamma)
\onestepterm{\ell}(\theta)
+ \gamma
\bootterm{\ell}(\theta) \, .
\owntag{\tddshort; }\label{eq:td2dd}
\end{gather}

\section{Theoretical Analysis}\label{ssec:theory}

We now study the learning dynamics of an idealized version of the TD-Flow methods, assuming that the flow-matching loss is minimized exactly at each iteration. Under this assumption, at each iteration we compute a probability path $m_t^{(n)}$ such that $m_1^{(n)} = \mathcal T^\pi m_1^{(n-1)}$, which immediately implies that $m_1^{(n)} \rightarrow m^\pi$ by the contraction property of $\mathcal T^\pi$. The following result shows that the overall probability paths $m_t^{(n)}$ follow a similar process. All proofs are deferred to \autoref{app:proofs}.
\begin{metaframe}
\begin{restatable}{theorem}{thmprobpathoperator}\label{thm:prob.path.operator}
    For any $n \geq 1$, the probability paths generated by \vtdfshort, \ctdfshort, or \tdfshort satisfy
    $$m_t^{(n+1)}(x \mid s,a) = \left(\mathcal{B}_t^\pi m_t^{(n)}\right)(x\mid s,a),\;\; \forall\, t\in[0,1]$$
    where $\mathcal{B}_t^\pi m := (1-\gamma) P_t + \gamma P^\pi m$ and $P_t(x | s,a) := \int p_{t\mid 1}( x \mid x_1) P(x_1 | s,a)\mathrm{d}x_1$. For any $t \in [0,1]$, the operator $\mathcal{B}_t^\pi$ is a $\gamma$-contraction in 1-Wasserstein distance, that is, for any couple of probability paths $p_t, q_t$,
    \begin{equation*}
        \sup_{s,a}\,\, W_1\left(\left(\mathcal{B}_t^\pi p_t\right)(\cdot \mid s, a), \left(\mathcal{B}_t^\pi q_t\right)(\cdot \mid s, a)\right)
        \leq \gamma \sup_{s, a} W_1\left(p_t(\cdot \mid s, a), q_t(\cdot \mid s, a)\right).
    \end{equation*}
\end{restatable}
\end{metaframe}
\autoref{thm:prob.path.operator} shows that all TD-flow methods fundamentally implement the same update where the probability path at $t\in[0,1]$ is obtained by applying a Bellman-like operator $\mathcal{B}_t$ to the previous iteration. This operator is a $\gamma$-contraction as $\mathcal{T}^\pi$, directly implying the following result.
\begin{metaframe}
\begin{restatable}{corollary}{corconvergence}\label{cor:convergence}
    Let $\{m_t^{(n)}\}_{n \geq 0}$ be the sequence of probability paths produced by \vtdfshort, \ctdfshort, or \tdfshort starting from an arbitrary vector field $v_t^{(0)}$. Then,
    \begin{align*}
        \lim_{n \rightarrow \infty} m_t^{(n)} = \widebar{m}_t = \mathcal{B}_t \widebar{m}_t,
    \end{align*}
    where $\widebar{m}_t$ is the unique fixed point of $\mathcal{B}_t$, and $\widebar{m}_t = m^{\mathrm{\textsc{mc}}}_t$, where $m^{\mathrm{\textsc{mc}}}_t(\cdot\mid s,a) = \int p_{t|1}(\cdot\mid x_1)\, m^\pi(x_1\mid s,a)$ is the probability path of the Monte-Carlo approach in~\eqref{eq:mc-cfm}.
\end{restatable}
\end{metaframe}
This corollary shows that the fixed point of $\mathcal{B}_t$ coincides with the probability path generated in Monte-Carlo Conditional Flow Matching~\eqref{eq:mc-cfm}, which assumes direct access to samples of $m^\pi$. 
An important subtlety in \autoref{thm:prob.path.operator} is that all algorithms apply the same operator for $n\geq 1$, but the result holds for $n=0$ only for \tdfshort. This means that even starting from the same $\theta^{(0)}$, the three algorithms may generate different sequences $\{m_t^{(n)}\}_{n \geq 0}$, while still converging to $\widebar{m}_t$. In
\textcolor{metablue}{Theorems}~\ref{th:vtdf-bellman}~and~\ref{th:ctdf-bellman}
, we show we can reconcile \ctdfshort\ and \vtdfshort\ with \tdfshort\ under a mild assumption on the form of the initial vector field.

While \autoref{thm:prob.path.operator} analyzes an idealized version of the algorithms, in practice gradients are estimated from samples and the following analysis reveals important differences in their variance. 
We introduce the (unbiased) sample-based gradients for each of the algorithms,
\begin{gather*}
\mathbb{E}\big[g_{\mathrm{\vtdfshort}}(Y_\mathrm{\vtdfshort})\big] = \nabla_\theta\, \ell_{\mathrm{\vtdfshort}}(\theta),\\
\mathbb{E}\big[g_{\mathrm{\ctdfshort}}(Y_\mathrm{\ctdfshort})\big] = \nabla_\theta\, \ell_{\mathrm{\ctdfshort}}(\theta)\\
\mathbb{E}\big[g_{\mathrm{\tdfshort}}(Y_\mathrm{\tdfshort})\big] = \nabla_\theta\, \ell_{\mathrm{\tdfshort}}(\theta),
\end{gather*}
where $Y$ summarizes the random variables involved in the loss definitions in~\eqref{eq:td-cfm},~\eqref{eq:td-cfm(c)}, and~\eqref{eq:td2fm} (see \autoref{app:variance.analysis} for a formal definition of the gradients). We want to compare the total variance of the gradient estimates $\sigma^2 = \mathrm{Tr}\big( \mathrm{Cov}_Y \left [ \,g(Y)\, \right] \big)$, where $\mathrm{Tr}$ denotes the trace.
\begin{metaframe}
\begin{restatable}{theorem}{thmvartdfvtdfmain}\label{th:var-tdf-vtdf-main}
For any $n\geq 1$ and $t\in[0,1]$, assume that $m^{(n)}_t(x \mid s, a) = \int p_{t|1}(x \mid x_1) m^{(n)}_1(x_1 \mid s, a) \mathrm{d}x_1$, then
\begin{equation*}
    \sigma^2_{\mathrm{\vtdfshort}} =  \sigma^2_{\mathrm{\tdfshort}}
    + \gamma^2\, \mathbb{E}_\rho \left[ \mathrm{Tr}\left( \mathrm{Cov}_{X_1 \mid S, A, X_t} \left[ \nabla_\theta\, v_t(X_t\mid S, A;\theta)^\top u_{t\mid1}(X_t \mid X_1)\right] \right)\right].
\end{equation*} 
\end{restatable}
\end{metaframe}
\begin{metaframe}
\begin{restatable}{theorem}{thmvarctdfmain}\label{th:var-ctdf-main}
For any $n\geq 1$ and $t\in[0,1]$, assume that $m^{(n)}_t(x \mid s, a) = \int p_{t|0,1}(x \mid x_0, x_1) m^{(n)}_{0, 1}(x_0, x_1 \mid s, a)  \mathrm{d}x_0 \mathrm{d}x_1$\,\footnotemark, then we obtain 
\begin{equation*}
    \sigma^2_{\mathrm{\ctdfshort}}  =  \sigma^2_{\mathrm{\tdfshort}}
    + \gamma^2 \mathbb{E}_\rho \left[ \mathrm{Tr}\left( \mathrm{Cov}_{Z \mid S, A, X_t} \left[ \nabla_\theta\, v_t(X_t\mid S, A;\theta)^\top u_{t\mid Z}(X_t\mid Z)\right] \right)\right],
\end{equation*}  
where $Z=(X_0, X_1)$. 
Furthermore, if we use straight conditional paths, \textit{i.e.}, $X_t = tX_1 + (1-t)X_0$, and the linear interpolant $X_t$ does not intersect for any $s,a,s'$, then $\sigma^2_{\mathrm{\ctdfshort}} = \sigma^2_{\mathrm{\tdfshort}}$.
\end{restatable}
\end{metaframe}
\footnotetext{$m_{0,1}^{(n)}(x_0, x_1 | s,a) = m_{0}(x_0) \delta_{\psi^{(n)}_1(x_0 \mid s, a)}(x_1)$ is the joint distribution of $(X_0, X_1)$, \textit{i.e} the endpoints of the ODE.}

In both results, the probability path $m^{(n)}_t$ from the previous iteration must be identical for the algorithms being compared.
The analysis reveals that \vtdfshort and \ctdfshort suffer from a larger variance compared to \tdfshort, which uses the vector field $v^{(n)}$ both to sample $X_t$ and as a target for the regression problem.
This variance gap is ``discounted'' by $\gamma^2$, which suggests that the performance of these algorithms would be similar for problems with small horizons but would increase as $\gamma \to 1$.
The extra variance in both cases stems from samples generated by the algorithm (i.e., they do not depend on the transitions available in the dataset).
In this sense, we can refer to it as \emph{computational variance}, and in principle, it could be reduced by increasing the number of samples $X_0$, $X_1$, and $X_t$ used in gradient computation.
While the variance of \vtdfshort and \ctdfshort cannot be directly compared, we expect that constructing $X_t$ from $X_0$ and $X_1$ (instead of $X_1$ only) will tend to reduce its variance. 
Specifically, when $X_t$ is obtained by linear interpolation between $X_0$ and $X_1$, and it does not generate crossing paths, the variance of \ctdfshort reduces to the one of \tdfshort.

\section{Experiments}\label{sec:experiments}
We now present a series of experiments to assess the efficacy of our TD-based flow and diffusion approaches with baselines employing Generative Adversarial Networks \citep{goodfellow14gan} and $\beta$-Variational Auto-Encoders \citep{higgins17beta}.
Following the methodology from ~\citet{touati23zeroshot, pirotta24bfmil}, we benchmark $22$ tasks spanning $4$ domains (Maze, Walker, Cheetah, Quadruped) from the DeepMind Control Suite \citep{tunyasuvunakool20dmc}.
For a single policy, we evaluate how well each method models its i) successor measure and ii) value function.
While lower errors in estimating the successor measure are expected to lead to better value estimation, this is not always the case since modeling errors may disproportionally affect states with negligible rewards.
Additionally, motivated by our theoretical results, we explore how the probability path's design affects our proposed methods' relative performance.

Finally, we examine the scalability of our approach by learning a generative model of the successor measure across a large class of parameterized policies derived from the Forward-Backward (FB) representation \citep{touati21fwdbwd,touati23zeroshot}, a non-generative model of the successor measure.
We conclude by demonstrating how $\textsc{td}{}^2$ methods enable more effective planning for task-relevant policies when performing Generalized Policy Improvement \citep[GPI;][]{barreto2017sf}, far surpassing the capabilities of FB alone.

\subsection{Empirical Evaluation of Geometric Horizon Models}\label{sec:experiments.singlepol}

Before benchmarking, we must first obtain a policy to evaluate.
We follow the approach taken in \citet{thakoor22ghm} and pre-train a set of deterministic policies -- one for each task -- using TD3 \citep{fujimoto18td3}. 
The final policy obtained from this pre-training phase is now fixed for the remainder of our experiments.
GHM training proceeds in an off-policy manner where we learn the successor measure of a TD3 policy using transition data from the ExoRL dataset \citep{yarats22exorl}; specifically, we use a dataset of $10$M transitions collected by a random network distillation policy \citep{burda19exploration}.
All GHM methods are trained for $3$M gradient steps using the AdamW optimizer \citep{loshchilov19decoupled} with a batch size of $1024$ and weight decay of $0.001$.
We maintain a target network using an exponential moving average of the training parameters with a step size of $0.001$.
Special care was taken to match the capacity of the neural networks between methods with a UNet-style architecture employed for all flow and diffusion methods, while the GAN and VAE baselines use an MLP with residual connections for all their respective networks.
Full details for the training methodology, network architecture, and hyper-parameters can be found in \autoref{app:experimental.details}.
\begin{wrapfigure}{r}{0.5225\textwidth}  %
    \vspace{1.5mm}
    \centering
    \includegraphics[width=\linewidth]{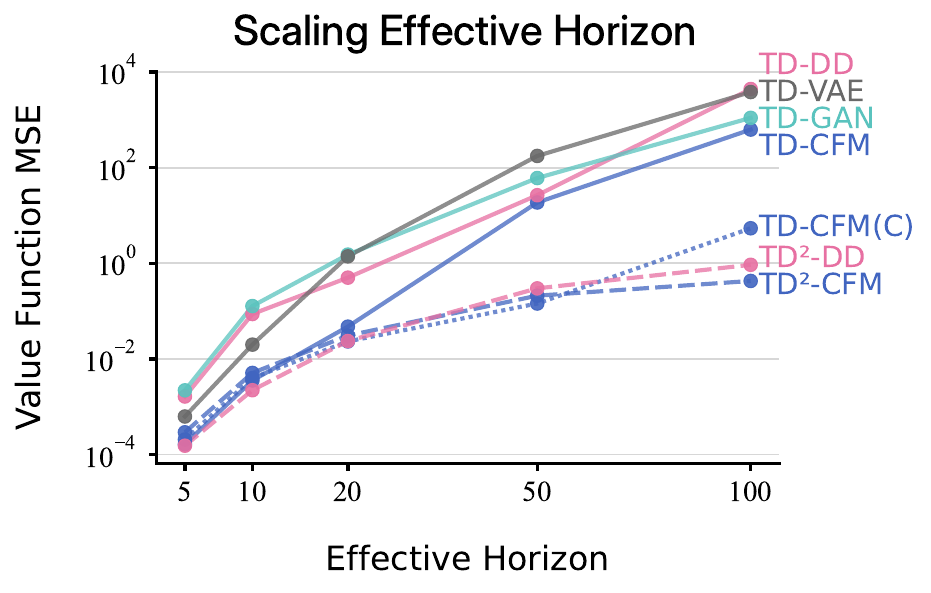}
    \caption{Value-Function prediction error as a function of the effective horizon $(1-\gamma)^{-1}$ for $\gamma \in \{0.8, 0.9, 0.95, 0.98, 0.99\}$ on the \textsc{Pointmass} loop task. \textsc{td}${}^2$ methods show impressive robustness to increasingly long-horizon predictions.}
    \label{fig:gamma-scaling-mse}
    \vspace{-4mm}
\end{wrapfigure}
\leavevmode

We implement all conditional flow matching methods (\vtdfshort, \ctdfshort, \tdfshort) with the Optimal Transport Gaussian conditional path from \citet{lipman2022flow}.
When constructing our bootstrap targets, we sample from the neural ODE using the Midpoint solver with a constant step size of $t / 10$ for a maximum of $10$ steps. For \tdfshort, we sample $t \sim \mathcal{U}([0, 1])$; otherwise, we integrate to $t=1$ and construct $X_t$ using the conditional path.
For Denoising Diffusion methods (\vtddshort, \tddshort), we train a DDPM \citep{ho20ddpm} by discretizing $\beta \in (0.1, 20)$ using $T = 1,000$ steps.
We construct diffusion bootstrapped targets using $20$ steps of the DDIM \citep{song21ddim} sampler. For \vtddshort, we solve to $t=0$ and regress towards the noise that re-corrupted our sample.
Alternatively, \tddshort directly regresses towards the noise prediction from the target network at a randomly selected noise level.
The first baseline we consider is a GHM instantiated as a Generative Adversarial Network \citep{goodfellow14gan} similar to the one found in \citet{janner20gmodel}.
We follow the best practices from \citet{huang24gan} with the primary modification being a relativistic discriminator \citep{jolicoeur-martineau19rgan} equipped with a zero-centered gradient penalty on both real and fake samples.
For our second baseline, we implement a $\beta$-VAE \citep{higgins17beta} following the practices outlined in \citet{thakoor22ghm}.

To evaluate the quality of our models, we first generate samples from the ground truth successor measure $\sm^\pi$ according to the following procedure.
We first randomly sample $64$ source states $S_0$ from the initial state distribution and execute policy $\pi$ for $1,000$ steps. Along each trajectory, we resample $2048$ states with replacement according to the stopping time $t \sim \text{Geometric}(1 - \gamma)$.
For the same $64$ source states, we generate a matching set of $2048$ samples from each GHM.
Now in possession of these two sets of samples, we evaluate the: 1) log-likelihood of the true samples for models with tractable densities (i.e., diffusion and flow methods); 2) Earth Mover's Distance \citep[EMD;][]{RubnerTG00}, which quantifies the minimal transport cost between the two empirical distributions; and 3) mean-squared error of a Monte-Carlo estimate of the true value function $Q^\pi$ and the value function derived from GHM samples using \eqref{eq:sm-vf}. Full details can be found in \autoref{app:evaluation}.

Having established our training framework, baselines, and evaluation protocol, we proceed to investigate a key prediction from our theoretical analysis.
Our variance analysis suggests that our TD-Flow framework should enable more stable training across extended temporal horizons.
To validate this hypothesis, we train each GHM for $3$ seeds on the loop task in the Maze domain while varying the effective horizon $(1 - \gamma)^{-1}$ across five values: $\{ 5, 10, 20, 50, 100 \}$. %
\autoref{fig:gamma-scaling-mse} illustrates the relationship between value function MSE and the effective horizon.
The results demonstrate that \textsc{td}${}^2$-based methods maintain consistent performance even as the effective horizon increases, while alternative approaches show significant performance degradation.
Notably, at an effective horizon of $100$, \textsc{td}${}^2$-based methods maintain their accuracy and achieve performance improvements of nearly four orders of magnitude compared to their naive implementations.
These results empirically support for our initial hypothesis, with the stability of \textsc{td}${}^2$ methods aligning with our predictions.

\begin{wrapfigure}{r}{0.4975\textwidth}  %
    \vspace{-3mm}
    \centering
    \begin{minipage}{\linewidth}
        \centering
        \captionsetup{type=table}
        \caption{Evaluation results comparing our \textsc{td}-based methods along with \tdganshort and \tdvaeshort baselines for a single-policy. Results are computed over $19$ tasks from $4$ domains and further averaged across $3$ seeds. For each metric we \mhl{highlight} the best performing methods.  }\label{tab:single_policy_value.main}
        \resizebox{\linewidth}{!}{
        \begin{tabular}{lcccc}
\toprule
&
{\fontsize{11}{13}\selectfont \textbf{Method}} &
{\fontsize{11}{13}\selectfont \textbf{EMD}  \raisebox{0.3ex}{\scalebox{0.7}{$\downarrow$}}}  &
{\fontsize{11}{13}\selectfont \textbf{Norm NLL} \raisebox{0.3ex}{\scalebox{0.7}{$\downarrow$}}} &
{\fontsize{11}{13}\selectfont \textbf{MSE(V)} \raisebox{0.3ex}{\scalebox{0.7}{$\downarrow$}}} \\
\hline%
\multirow{8}{*}{\rotatebox{90}{\textsc{Cheetah}}} & \vtddshort & $20.22$ (\small\textcolor{gray}{$0.26$}) & $2.824$ (\small\textcolor{gray}{$0.195$}) & $454.49$ (\small\textcolor{gray}{$131.97$}) \\
 & \tddshort & $14.14$ (\small\textcolor{gray}{$1.08$}) & $0.806$ (\small\textcolor{gray}{$0.016$}) & $189.15$ (\small\textcolor{gray}{$23.63$}) \\
 \cline{2-5}
 & \vtdfshort & $12.26$ (\small\textcolor{gray}{$0.02$}) & $0.886$ (\small\textcolor{gray}{$0.024$}) & $228.77$ (\small\textcolor{gray}{$2.20$}) \\
 & \ctdfshort & \cellcolor{best} $10.51$ (\small\textcolor{gray}{$0.06$}) & $0.447$ (\small\textcolor{gray}{$0.020$}) & $140.78$ (\small\textcolor{gray}{$18.72$}) \\
 & \tdfshort & \cellcolor{best} $10.57$ (\small\textcolor{gray}{$0.07$}) & \cellcolor{best} $0.422$ (\small\textcolor{gray}{$0.014$}) & \cellcolor{best} $135.22$ (\small\textcolor{gray}{$19.79$}) \\
 \cline{2-5}
 & \tdganshort & $23.97$ (\small\textcolor{gray}{$0.46$}) & --- & $2463.22$ (\small\textcolor{gray}{$628.05$}) \\
 & \tdvaeshort & $83.77$ (\small\textcolor{gray}{$0.41$}) & --- & $1284.27$ (\small\textcolor{gray}{$37.62$}) \\
    \hline
\multirow{8}{*}{\rotatebox{90}{\textsc{Pointmass}}} & \vtddshort & $0.149$ (\small\textcolor{gray}{$0.001$}) & $2.974$ (\small\textcolor{gray}{$0.100$}) & $1245.20$ (\small\textcolor{gray}{$29.27$}) \\
 & \tddshort & $0.027$ (\small\textcolor{gray}{$0.001$}) & $0.761$ (\small\textcolor{gray}{$0.082$}) & $11.13$ (\small\textcolor{gray}{$3.09$}) \\
 \cline{2-5}
 & \vtdfshort & $0.062$ (\small\textcolor{gray}{$0.003$}) & $0.554$ (\small\textcolor{gray}{$0.033$}) & $355.56$ (\small\textcolor{gray}{$82.83$}) \\
 & \ctdfshort & $0.022$ (\small\textcolor{gray}{$0.002$}) & $-0.696$ (\small\textcolor{gray}{$0.094$}) & $11.89$ (\small\textcolor{gray}{$3.16$}) \\
 & \tdfshort & \cellcolor{best} $0.021$ (\small\textcolor{gray}{$0.000$}) & \cellcolor{best} $-0.843$ (\small\textcolor{gray}{$0.027$}) & \cellcolor{best} $8.74$ (\small\textcolor{gray}{$2.09$}) \\
 \cline{2-5}
 & \tdganshort & $0.203$ (\small\textcolor{gray}{$0.037$}) & --- & $1257.26$ (\small\textcolor{gray}{$112.86$}) \\
 & \tdvaeshort & $0.410$ (\small\textcolor{gray}{$0.036$}) & --- & $1821.89$ (\small\textcolor{gray}{$69.78$}) \\
    \hline
\multirow{8}{*}{\rotatebox{90}{\textsc{Quadruped}}}  & \vtddshort & $28.33$ (\small\textcolor{gray}{$0.33$}) & $1.908$ (\small\textcolor{gray}{$0.041$}) & $1490.75$ (\small\textcolor{gray}{$444.49$}) \\
 & \tddshort & $22.64$ (\small\textcolor{gray}{$2.47$}) & $0.861$ (\small\textcolor{gray}{$0.028$}) & $159.03$ (\small\textcolor{gray}{$14.64$}) \\
 \cline{2-5}
 & \vtdfshort & $15.73$ (\small\textcolor{gray}{$0.06$}) & $1.056$ (\small\textcolor{gray}{$0.002$}) & $525.06$ (\small\textcolor{gray}{$28.90$}) \\
 & \ctdfshort & \cellcolor{best} $14.38$ (\small\textcolor{gray}{$0.03$}) & $0.488$ (\small\textcolor{gray}{$0.003$}) & $155.25$ (\small\textcolor{gray}{$5.58$}) \\
 & \tdfshort & \cellcolor{best} $14.51$ (\small\textcolor{gray}{$0.05$}) & \cellcolor{best} $0.379$ (\small\textcolor{gray}{$0.011$}) & \cellcolor{best} $141.77$ (\small\textcolor{gray}{$3.10$}) \\
 \cline{2-5}
 & \tdganshort & $36772.12$ (\small\textcolor{gray}{$13898.25$}) & --- & $2634.69$ (\small\textcolor{gray}{$798.38$}) \\
 & \tdvaeshort & $60.27$ (\small\textcolor{gray}{$0.28$}) & --- & $1156.33$ (\small\textcolor{gray}{$36.52$}) \\
    \hline
\multirow{8}{*}{\rotatebox{90}{\textsc{Walker}}} & \vtddshort & $20.58$ (\small\textcolor{gray}{$0.24$}) & $2.649$ (\small\textcolor{gray}{$0.137$}) & $382.40$ (\small\textcolor{gray}{$458.63$}) \\
 & \tddshort & $12.09$ (\small\textcolor{gray}{$0.12$}) & $0.537$ (\small\textcolor{gray}{$0.060$}) & $39.04$ (\small\textcolor{gray}{$6.08$}) \\
 \cline{2-5}
 & \vtdfshort & $13.53$ (\small\textcolor{gray}{$0.11$}) & $0.713$ (\small\textcolor{gray}{$0.028$}) & $225.27$ (\small\textcolor{gray}{$42.43$}) \\
 & \ctdfshort & \cellcolor{best} $11.91$ (\small\textcolor{gray}{$0.02$}) & $0.219$ (\small\textcolor{gray}{$0.016$}) & $30.71$ (\small\textcolor{gray}{$3.44$}) \\
 & \tdfshort & \cellcolor{best} $11.92$ (\small\textcolor{gray}{$0.10$}) & \cellcolor{best} $0.104$ (\small\textcolor{gray}{$0.001$}) & \cellcolor{best} $28.35$ (\small\textcolor{gray}{$6.10$}) \\
 \cline{2-5}
 & \tdganshort & $24.51$ (\small\textcolor{gray}{$0.89$}) & --- & $3690.65$ (\small\textcolor{gray}{$1117.94$}) \\
 & \tdvaeshort & $111.73$ (\small\textcolor{gray}{$2.53$}) & --- & $2457.61$ (\small\textcolor{gray}{$16.25$}) \\
\bottomrule
\end{tabular}

        }
    \end{minipage}

    \vspace{2em}

    \begin{minipage}{\linewidth}
        \centering
        \captionsetup{type=table}
        \caption{Performance difference between \ctdfshort and \tdfshort for curved and straight conditional paths. Lower is better with negative values indicating a net improvement by employing a curved paths.}\label{tab:curved-path}
        \resizebox{.95\linewidth}{!}{
            \begin{tabular}{cccc}
\toprule 
{\fontsize{11}{13}\selectfont \textbf{Method}} &
{\fontsize{11}{13}\selectfont \textbf{EMD}  \raisebox{0.3ex}{\scalebox{0.7}{$\downarrow$}}}  &
{\fontsize{11}{13}\selectfont \textbf{Norm NLL} \raisebox{0.3ex}{\scalebox{0.7}{$\downarrow$}}} &
{\fontsize{11}{13}\selectfont \textbf{MSE(V)} \raisebox{0.3ex}{\scalebox{0.7}{$\downarrow$}}} \\
\hline
\ctdfshort & $14.08$ (\small\textcolor{gray}{$12.42$}) & $1.79$ (\small\textcolor{gray}{$1.98$}) & $310.45$ (\small\textcolor{gray}{$258.94$}) \\
\tdfshort & $0.09$ (\small\textcolor{gray}{$0.09$}) & $-0.01$ (\small\textcolor{gray}{$0.04$}) & $-3.36$ (\small\textcolor{gray}{$7.76$}) \\
\bottomrule
\end{tabular}

        }
        \vspace{-2mm}
    \end{minipage}
\end{wrapfigure}
\leavevmode
In the following, we shift our attention to a more in-depth analysis of the largest horizon of $100$ ($\gamma = 0.99$).
For each algorithm, we train a GHM for $3$ independent seeds for all domains and tasks.
\autoref{tab:single_policy_value.main} reports aggregate performance across our full suite of metrics. For each domain and metric, we highlight results in a $1\%$ range with respect to the best-performing method.
The results demonstrate a clear pattern of superior performance for \textsc{td}${}^2$-based algorithms: \tdfshort achieves significant improvements over \vtdfshort with a $10\times$ reduction in value-function MSE, $1.5\times$ reduction in EMD, and $3\times$ reduction in log-likelihood, averaged across all four domains.
In line with our theoretical predictions, the coupled variant of \vtdfshort performs comparably to \tdfshort, given straight conditional paths.
While a comparison between flow matching and diffusion is not at the core of this paper, in our experiments, flow matching generally improves performance over diffusion across all metrics.
We posit this is primarily due to noise in the diffusion process adversely impacting an already noisy prediction problem at large horizons.

Given the comparable performance between \ctdfshort and \tdfshort with straight conditional paths, we next examine how these methods behave with alternative path geometries.
Our theoretical analysis suggests an important distinction: \tdfshort should maintain its effectiveness with non-straight paths, while the performance of \ctdfshort should degrade.
To test this prediction, we maintain the methodology above while replacing conditional path in \eqref{eq:td2fm} with the following curved path $p_{t\mid1}(\cdot\,|\, X_1) = \mathcal{N}(\cdot\,|\,\alpha_t X_1, \beta_t^2)$ with coefficients $\alpha_t = \sin\left(\frac{\pi}{2}t\right)$ and $\beta_t = \cos\left(\frac{\pi}{2}t\right)$. The corresponding conditional vector field is now given by $u_{t|1}(X_t|X_1) = \big(\dot{\alpha}_t - \frac{\alpha_t}{\beta_t}\big) X_1 + \frac{\dot{\beta}_t}{\beta_t}X_t$.
Additionally, for \ctdfshort we condition the curved path above on $X_0$ and $X_1$ resulting in the conditional vector field $u_{t\mid0,1}(X_t\mid X_0, X_1) = \frac{\pi}{2} \big( \beta_t X_1 - \alpha_t X_0 \big)$.
\autoref{tab:curved-path} illustrates the performance difference relative to the straight path results (\autoref{tab:single_policy_value.main}) averaged across all domains and tasks.
The results strongly support our theoretical prediction: \tdfshort not only maintained but surprisingly improved performance compared to the linear path. In contrast, \ctdfshort showed significant performance degradation, confirming our hypothesis about its limitations with non-straight paths.

\subsection{Planning via Generalized Policy Improvement}

\begin{figure*}[t]
    \centering
    \includegraphics[width=.975\linewidth]{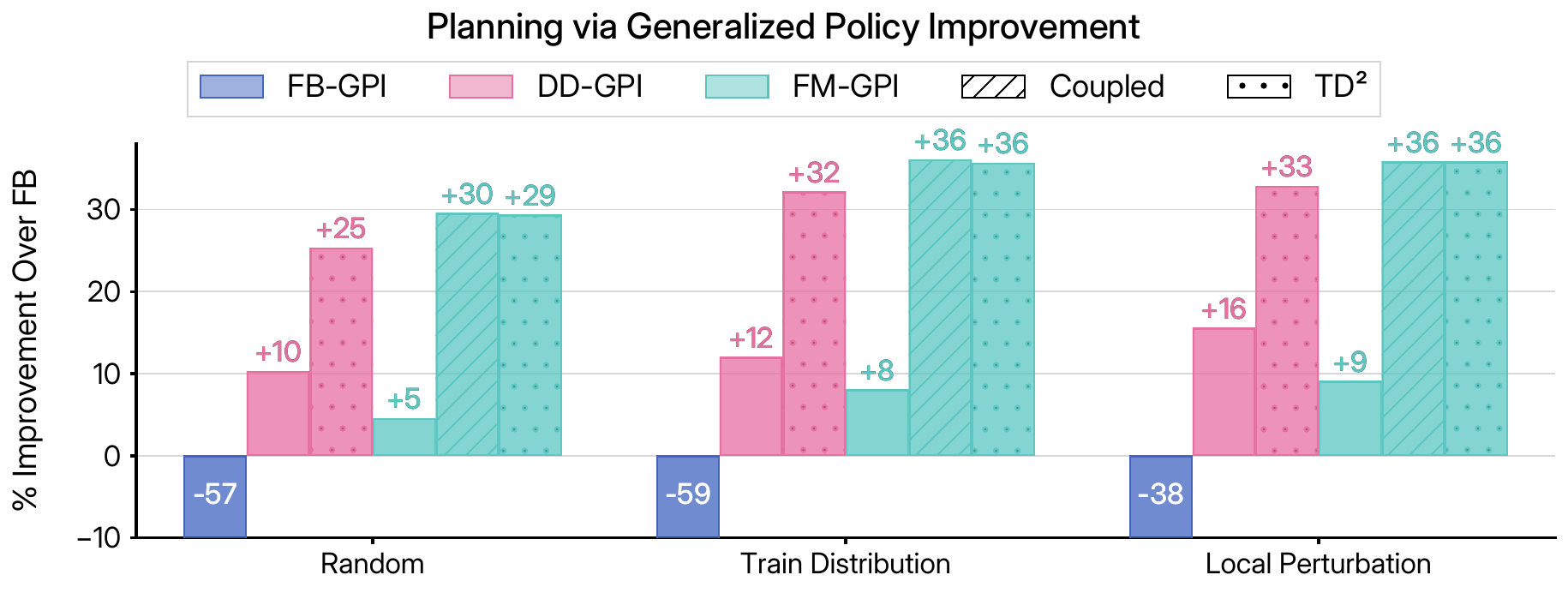}
    \caption{Performance improvement over the zero-shot Forward Backward \citep[FB;][]{touati21fwdbwd} policies when planning with Generalized Policy Improvement \citep[GPI;][]{barreto2017sf}. \textsc{FB-GPI} performs GPI over the FB value-function $Q^{\pi_w}$. \textsc{DD-GPI} and \textsc{FM-GPI} perform GPI with the value function implied by the GHM $m^{\pi_w}$ for our diffusion-based and flow-based methods, respectively. Results are averaged over 22 tasks across 4 domains.}
    \label{fig:gpi.percentage}
\end{figure*}

We now turn our attention towards training policy-conditioned GHMs which can be utilized for test-time planning.
To accomplish this, we first pre-train a Forward Backward \citep[FB;][]{touati21fwdbwd, touati23zeroshot} representation using the same dataset of $10$M transitions as described in \autoref{sec:experiments.singlepol}. This pre-training yields a class of $w$-conditioned policies $\pi_w$, where each $w \in \setfont{W} = \setfont{S}^{d-1}(\sqrt{d})$ represents an embedding of a reward function situated on a $d$-dimensional hypersphere with radius $\sqrt{d}$. We then train the GHM $m^{\pi_w}$ conditioned on the policy by incorporating the embedding $w$ directly into the model's input. All GHM methods are trained for $8$M gradient steps, maintaining the same parameters used in \autoref{sec:experiments.singlepol}, with the exception of a higher weight decay coefficient of $0.01$. For additional insights into the accuracy of the policy-conditioned GHMs, we direct the reader to \autoref{app:additional.results}. Overall, we observed similar trends to those seen in our single-policy experiments.

Given that both FB and $w$-conditioned GHM models enable estimation of a policy's value function $Q^{\pi_w}$, we can utilize this information to perform Generalized Policy Improvement \citep[GPI;][]{barreto2017sf} during evaluation.
Specifically, at each time step $t$, we choose an action $a_t = \pi_{w_t}(s_t)$, where $w_t$ is derived as follows:
\begin{equation}\label{eq:gpi}
w_t \in \argmax_{w\sim D(\setfont{W})}\,\, \underbrace{(1 - \gamma)^{-1} \E_{X \sim \sm^{\pi_w}(\cdot|s_t,\pi_w(s_t)))}{r(X)}}_{Q^{\pi_w}(s_t, \pi_w(s_t))}\,.
\end{equation}
Here $D(\setfont{W})$ is a sampling distribution over $\setfont{W}$.
We consider three such distributions: 
\emph{i)} \textit{Random}: uniform distribution over $\setfont{W}$; \emph{ii)} \textit{Local Perturbation}: we perturb the embedding $w_r$ of the task reward $r$ by the uniform distribution; \emph{iii)} \textit{Train Distribution}: we sample $w$ from the training distribution used by FB.
To approximate \eqref{eq:gpi}, we sample $255$ embeddings from $D(\setfont{W})$ and explicitly include the task embedding $w_r$, resulting in a maximization over $256$ policies.
To estimate the action-value function, we average the reward over $128$ states sampled from $m^{\pi_w}$.
Performance is measured by averaging returns over $100$ episodes, each lasting $1000$ steps.

\autoref{fig:gpi.percentage} illustrates the average percentage of improvement for each algorithm and $w$-sampling strategy relative to the performance of the FB policy $\pi_{w_r}$ for the task reward $r$.
We refer to \autoref{app:additional.results} for a more detailed view of these results.
All TD-based GHM approaches lead to a significant improvement over the base FB policy, with \ctdfshort{} and \tdfshort{} providing $\approx30\%+$ improvement with all sampling approaches.
\tddshort{} also leads to significant performance gains but is still dominated by the flow matching methods.
Notably, FB-based GPI not only fails to improve performance but actually deteriorates it on average with significant degradation observed in three out of four domains (detailed results available in \autoref{app:additional.results}).
When comparing different distributions $D(\setfont{W})$, we observe that while FB-GPI's performance fluctuates considerably, GHM methods maintain their robustness across distributions, showing only minor variation.
These results underscore the ability of our improved GHMs to make long-term predictions enabling powerful planning capabilities.

\section{Discussion}\label{sec:discussion}
In this paper, we introduced temporal difference flows, a novel generative modeling approach that significantly advances long-horizon predictive models of state.
By leveraging the successor measure's temporal difference structure both in its sampling procedure and learning objective, \tdfshort and \tddshort effectively address challenges associated with modeling long-range state dynamics.
The methods developed in this paper provide a robust theoretical and empirical foundation that demonstrates the advantages of our framework across a range of tasks, metrics, and domains.
We envision numerous exciting applications emerging from this work, particularly around imitation learning \citep{wu2025diffusing,jain25nonadversarial}, planning \citep{sutton91dyna,thakoor22ghm,zhu24gom}, and off-policy evaluation \citep{precup00eligibility,precup01offpolicy,nachum19dualdice,fujimoto21sr}.
Furthermore, recent work on consistency models \citep{song23consistency,yang24consistency} and self-distillation \citep{frans24onestep} suggests promising avenues for tackling the computational burden of sampling --- a limitation common to the family of iterative generative models that our approach builds upon.

\section*{Acknowledgements}

We thank Yaron Lipman for informative discussions on flow matching and for providing valuable feedback on our work.
Additionally, we thank Yann Ollivier for his keen insights and thoughtful suggestions throughout the development of this work.
Finally, our work was made possible thanks to contributions from the Python community. In particular, we made extensive use of Numpy \citep{harris20numpy}, Matplotlib \citep{hunter07mpl}, Seaborn \citep{waskom2021seaborn}, einops \citep{rogozhnikov22einops}, and Mujoco \citep{todorov2012mujoco}.

\bibliography{bookends}

\begin{thebibliography}{94}
\providecommand{\natexlab}[1]{#1}
\providecommand{\url}[1]{\texttt{#1}}
\expandafter\ifx\csname urlstyle\endcsname\relax
  \providecommand{\doi}[1]{doi: #1}\else
  \providecommand{\doi}{doi: \begingroup \urlstyle{rm}\Url}\fi

\bibitem[Albergo and Vanden{-}Eijnden(2023)]{albergo23building}
Michael~S. Albergo and Eric Vanden{-}Eijnden.
\newblock Building normalizing flows with stochastic interpolants.
\newblock In \emph{International Conference on Learning Representations,
  ({ICLR})}, 2023.

\bibitem[Anderson(1982)]{anderson1982reverse}
Brian~DO Anderson.
\newblock Reverse-time diffusion equation models.
\newblock \emph{Stochastic Processes and their Applications}, 12\penalty0
  (3):\penalty0 313--326, 1982.

\bibitem[Ba et~al.(2016)Ba, Kiros, and Hinton]{ba2016ln}
Jimmy Ba, Jamie Kiros, and Geoffrey~E. Hinton.
\newblock Layer normalization.
\newblock \emph{CoRR}, abs/1607.06450, 2016.

\bibitem[Barreto et~al.(2017)Barreto, Dabney, Munos, Hunt, Schaul, Silver, and
  van Hasselt]{barreto2017sf}
Andr{\'e} Barreto, Will Dabney, R{\'e}mi Munos, Jonathan~J. Hunt, Tom Schaul,
  David Silver, and Hado van Hasselt.
\newblock Successor features for transfer in reinforcement learning.
\newblock In \emph{Neural Information Processing Systems (NeurIPS)}, 2017.

\bibitem[Barreto et~al.(2020)Barreto, Hou, Borsa, Silver, and
  Precup]{andre2020fast}
Andr{\'e} Barreto, Shaobo Hou, Diana Borsa, David Silver, and Doina Precup.
\newblock Fast reinforcement learning with generalized policy updates.
\newblock \emph{Proceedings of the National Academy of Sciences (PNAS)},
  117\penalty0 (48):\penalty0 30079--30087, 2020.

\bibitem[Blier et~al.(2021)Blier, Tallec, and Ollivier]{blier21successor}
L{\'e}onard Blier, Corentin Tallec, and Yann Ollivier.
\newblock Learning successor states and goal-dependent values: A mathematical
  viewpoint.
\newblock \emph{CoRR}, abs/2101.07123, 2021.

\bibitem[Borsa et~al.(2019)Borsa, Barreto, Quan, Mankowitz, van Hasselt, Munos,
  Silver, and Schaul]{borsa19usfa}
Diana Borsa, Andr{\'e} Barreto, John Quan, Daniel~J. Mankowitz, Hado van
  Hasselt, R{\'e}mi Munos, David Silver, and Tom Schaul.
\newblock Universal successor features approximators.
\newblock In \emph{International Conference on Learning Representations
  (ICLR)}, 2019.

\bibitem[Burda et~al.(2019)Burda, Edwards, Storkey, and
  Klimov]{burda19exploration}
Yuri Burda, Harrison Edwards, Amos~J. Storkey, and Oleg Klimov.
\newblock Exploration by random network distillation.
\newblock In \emph{International Conference on Learning Representations
  ({ICLR})}, 2019.

\bibitem[Cetin et~al.(2024)Cetin, Touati, and Ollivier]{cetin2024finer}
Edoardo Cetin, Ahmed Touati, and Yann Ollivier.
\newblock Finer behavioral foundation models via auto-regressive features and
  advantage weighting.
\newblock \emph{CoRR}, abs/2412.04368, 2024.

\bibitem[Chen et~al.(2018)Chen, Rubanova, Bettencourt, and
  Duvenaud]{chen18neuralode}
Tian~Qi Chen, Yulia Rubanova, Jesse Bettencourt, and David Duvenaud.
\newblock Neural ordinary differential equations.
\newblock In \emph{Neural Information Processing Systems ({NeurIPS})}, 2018.

\bibitem[Dayan(1993)]{dayan93sr}
Peter Dayan.
\newblock Improving generalization for temporal difference learning: The
  successor representation.
\newblock \emph{Neural Computation}, 1993.

\bibitem[De~Bortoli et~al.(2024)De~Bortoli, Korshunova, Mnih, and
  Doucet]{valentin24sbf}
Valentin De~Bortoli, Iryna Korshunova, Andriy Mnih, and Arnaud Doucet.
\newblock Schrödinger bridge flow for unpaired data translation.
\newblock In \emph{Neural Information Processing Systems (NeurIPS)}, 2024.

\bibitem[Dinh et~al.(2015)Dinh, Krueger, and Bengio]{dnih15nice}
Laurent Dinh, David Krueger, and Yoshua Bengio.
\newblock {NICE}: Non-linear independent components estimation.
\newblock In \emph{International Conference on Learning Representations (ICLR),
  Workshop Track Proceedings}, 2015.

\bibitem[Dinh et~al.(2017)Dinh, Sohl-Dickstein, and Bengio]{dnih17nvp}
Laurent Dinh, Jascha Sohl-Dickstein, and Samy Bengio.
\newblock Density estimation using real nvp.
\newblock In \emph{International Conference on Learning Representations
  (ICLR)}, 2017.

\bibitem[Farebrother et~al.(2023)Farebrother, Greaves, Agarwal, Le~Lan,
  Goroshin, Castro, and Bellemare]{farebrother2023pvn}
Jesse Farebrother, Joshua Greaves, Rishabh Agarwal, Charline Le~Lan, Ross
  Goroshin, Pablo~Samuel Castro, and Marc~G. Bellemare.
\newblock Proto-value networks: Scaling representation learning with auxiliary
  tasks.
\newblock In \emph{International Conference on Learning Representations
  (ICLR)}, 2023.

\bibitem[Flamary et~al.(2021)Flamary, Courty, Gramfort, Alaya, Boisbunon,
  Chambon, Chapel, Corenflos, Fatras, Fournier, Gautheron, Gayraud, Janati,
  Rakotomamonjy, Redko, Rolet, Schutz, Seguy, Sutherland, Tavenard, Tong, and
  Vayer]{flamary2021pot}
R{\'e}mi Flamary, Nicolas Courty, Alexandre Gramfort, Mokhtar~Z. Alaya,
  Aur{\'e}lie Boisbunon, Stanislas Chambon, Laetitia Chapel, Adrien Corenflos,
  Kilian Fatras, Nemo Fournier, L{\'e}o Gautheron, Nathalie~T.H. Gayraud,
  Hicham Janati, Alain Rakotomamonjy, Ievgen Redko, Antoine Rolet, Antony
  Schutz, Vivien Seguy, Danica~J. Sutherland, Romain Tavenard, Alexander Tong,
  and Titouan Vayer.
\newblock Pot: Python optimal transport.
\newblock \emph{Journal of Machine Learning Research}, 22\penalty0
  (78):\penalty0 1--8, 2021.

\bibitem[Frans et~al.(2025)Frans, Hafner, Levine, and Abbeel]{frans24onestep}
Kevin Frans, Danijar Hafner, Sergey Levine, and Pieter Abbeel.
\newblock One step diffusion via shortcut models.
\newblock In \emph{International Conference on Learning Representations
  ({ICLR})}, 2025.

\bibitem[Fujimoto et~al.(2018)Fujimoto, van Hoof, and Meger]{fujimoto18td3}
Scott Fujimoto, Herke van Hoof, and David Meger.
\newblock Addressing function approximation error in actor-critic methods.
\newblock In \emph{International Conference on Machine Learning (ICML)}, 2018.

\bibitem[Fujimoto et~al.(2021)Fujimoto, Meger, and Precup]{fujimoto21sr}
Scott Fujimoto, David Meger, and Doina Precup.
\newblock A deep reinforcement learning approach to marginalized importance
  sampling with the successor representation.
\newblock In \emph{International Conference on Machine Learning ({ICML})},
  2021.

\bibitem[Ghosh et~al.(2023)Ghosh, Bhateja, and Levine]{ghosh23icvf}
Dibya Ghosh, Chethan~Anand Bhateja, and Sergey Levine.
\newblock Reinforcement learning from passive data via latent intentions.
\newblock In \emph{International Conference on Machine Learning (ICML)}, 2023.

\bibitem[Goodfellow et~al.(2014)Goodfellow, Pouget-Abadie, Mirza, Xu,
  Warde-Farley, Ozair, Courville, and Bengio]{goodfellow14gan}
Ian Goodfellow, Jean Pouget-Abadie, Mehdi Mirza, Bing Xu, David Warde-Farley,
  Sherjil Ozair, Aaron Courville, and Yoshua Bengio.
\newblock Generative adversarial nets.
\newblock In \emph{Neural Information Processing Systems (NeurIPS)}, 2014.

\bibitem[Grathwohl et~al.(2019)Grathwohl, Chen, Bettencourt, Sutskever, and
  Duvenaud]{grathwohl19ffjord}
Will Grathwohl, Ricky T.~Q. Chen, Jesse Bettencourt, Ilya Sutskever, and David
  Duvenaud.
\newblock {FFJORD:} free-form continuous dynamics for scalable reversible
  generative models.
\newblock In \emph{International Conference on Learning Representations
  ({ICLR})}, 2019.

\bibitem[Hafner et~al.(2023)Hafner, Pasukonis, Ba, and
  Lillicrap]{hafner23mastering}
Danijar Hafner, Jurgis Pasukonis, Jimmy Ba, and Timothy~P. Lillicrap.
\newblock Mastering diverse domains through world models.
\newblock \emph{CoRR}, abs/2301.04104, 2023.

\bibitem[Hansen et~al.(2022)Hansen, Su, and Wang]{hansen22tdmpc}
Nicklas Hansen, Hao Su, and Xiaolong Wang.
\newblock Temporal difference learning for model predictive control.
\newblock In \emph{International Conference on Machine Learning ({ICML})},
  2022.

\bibitem[Hansen et~al.(2024)Hansen, Su, and Wang]{hansen24tdmpc2}
Nicklas Hansen, Hao Su, and Xiaolong Wang.
\newblock {TD-MPC2}: Scalable, robust world models for continuous control.
\newblock In \emph{International Conference on Learning Representations
  (ICLR)}, 2024.

\bibitem[Harris et~al.(2020)Harris, K.~Jarrod, van~der Walt, Gommers, Virtanen,
  Cournapeau, Wieser, Taylor, Berg, Smith, Kern, Picus, Hoyer, van Kerkwijk,
  Brett, Haldane, Fern\'andez~del R\'io, Wiebe, Peterson, G{\'e}rard-Marchant,
  Sheppard, Reddy, Weckesser, Abbasi, Gohlke, and Oliphant]{harris20numpy}
Charles~R. Harris, Millman K.~Jarrod, St{\'e}fan~J. van~der Walt, Ralf Gommers,
  Pauli Virtanen, David Cournapeau, Eric Wieser, Julian Taylor, Sebastian Berg,
  Nathaniel~J. Smith, Robert Kern, Matti Picus, Stephan Hoyer, Marten~H. van
  Kerkwijk, Matthew Brett, Allan Haldane, Jaime Fern\'andez~del R\'io, Mark
  Wiebe, Pearu Peterson, Pierre G{\'e}rard-Marchant, Kevin Sheppard, Tyler
  Reddy, Warren Weckesser, Hameer Abbasi, Christoph Gohlke, and Travis~E.
  Oliphant.
\newblock Array programming with numpy.
\newblock \emph{Nature}, 585\penalty0 (7825):\penalty0 357--362, 2020.

\bibitem[Higgins et~al.(2017)Higgins, Matthey, Pal, Burgess, Glorot, Botvinick,
  Mohamed, and Lerchner]{higgins17beta}
Irina Higgins, Loic Matthey, Arka Pal, Christopher Burgess, Xavier Glorot,
  Matthew Botvinick, Shakir Mohamed, and Alexander Lerchner.
\newblock beta-vae: Learning basic visual concepts with a constrained
  variational framework.
\newblock In \emph{International Conference on Learning Representations
  (ICLR)}, 2017.

\bibitem[Ho et~al.(2020)Ho, Jain, and Abbeel]{ho20ddpm}
Jonathan Ho, Ajay Jain, and Pieter Abbeel.
\newblock Denoising diffusion probabilistic models.
\newblock In \emph{Neural Information Processing Systems ({NeurIPS})}, 2020.

\bibitem[Huang et~al.(2024)Huang, Gokaslan, Kuleshov, and Tompkin]{huang24gan}
Nick Huang, Aaron Gokaslan, Volodymyr Kuleshov, and James Tompkin.
\newblock The gan is dead; long live the gan! a modern gan baseline.
\newblock In \emph{Neural Information Processing Systems (NeurIPS)}, 2024.

\bibitem[Hunter(2007)]{hunter07mpl}
J.~D. Hunter.
\newblock Matplotlib: A 2d graphics environment.
\newblock \emph{Computing in Science \& Engineering}, 9\penalty0 (3):\penalty0
  90--95, 2007.

\bibitem[Jafferjee et~al.(2020)Jafferjee, Imani, Talvitie, White, and
  Bowling]{jafferjee20dyna}
Taher Jafferjee, Ehsan Imani, Erin Talvitie, Martha White, and Michael Bowling.
\newblock Hallucinating value: {A} pitfall of dyna-style planning with
  imperfect environment models.
\newblock \emph{CoRR}, abs/2006.04363, 2020.

\bibitem[Jain et~al.(2023)Jain, Lehnert, Rish, and Berseth]{jain23maximum}
Arnav~Kumar Jain, Lucas Lehnert, Irina Rish, and Glen Berseth.
\newblock Maximum state entropy exploration using predecessor and successor
  representations.
\newblock In \emph{Neural Information Processing Systems ({NeurIPS})}, 2023.

\bibitem[Jain et~al.(2025)Jain, Wiltzer, Farebrother, Rish, Berseth, and
  Choudhury]{jain25nonadversarial}
Arnav~Kumar Jain, Harley Wiltzer, Jesse Farebrother, Irina Rish, Glen Berseth,
  and Sanjiban Choudhury.
\newblock Non-adversarial inverse reinforcement learning via successor feature
  matching.
\newblock In \emph{International Conference on Learning Representations
  (ICLR)}, 2025.

\bibitem[Janner et~al.(2020)Janner, Mordatch, and Levine]{janner20gmodel}
Michael Janner, Igor Mordatch, and Sergey Levine.
\newblock Gamma-models: Generative temporal difference learning for
  infinite-horizon prediction.
\newblock In \emph{Neural Information Processing Systems (NeurIPS)}, 2020.

\bibitem[Jolicoeur-Martineau(2019)]{jolicoeur-martineau19rgan}
Alexia Jolicoeur-Martineau.
\newblock The relativistic discriminator: a key element missing from standard
  gan.
\newblock In \emph{International Conference on Learning Representations
  (ICLR)}, 2019.

\bibitem[Kingma and Ba(2015)]{kingma15adam}
Diederik~P. Kingma and Jimmy Ba.
\newblock Adam: A method for stochastic optimization.
\newblock In \emph{International Conference on Learning Representations
  (ICLR)}, 2015.

\bibitem[Kingma and Welling(2014)]{kingma13autoencoding}
Diederik~P. Kingma and Max Welling.
\newblock Auto-encoding variational bayes.
\newblock In \emph{International Conference on Learning Representations
  (ICLR)}, 2014.

\bibitem[Lambert et~al.(2022)Lambert, Pister, and Calandra]{lambert22error}
Nathan Lambert, Kristofer Pister, and Roberto Calandra.
\newblock Investigating compounding prediction errors in learned dynamics
  models.
\newblock \emph{CoRR}, abs/2203.09637, 2022.

\bibitem[Le~Lan et~al.(2022)Le~Lan, Tu, Oberman, Agarwal, and
  Bellemare]{lelan22generalization}
Charline Le~Lan, Stephen Tu, Adam Oberman, Rishabh Agarwal, and Marc~G.
  Bellemare.
\newblock On the generalization of representations in reinforcement learning.
\newblock In \emph{International Conference on Artificial Intelligence and
  Statistics (AISTATS)}, 2022.

\bibitem[Le~Lan et~al.(2023{\natexlab{a}})Le~Lan, Greaves, Farebrother,
  Rowland, Pedregosa, Agarwal, and Bellemare]{lelan23subspace}
Charline Le~Lan, Joshua Greaves, Jesse Farebrother, Mark Rowland, Fabian
  Pedregosa, Rishabh Agarwal, and Marc~G. Bellemare.
\newblock A novel stochastic gradient descent algorithm for learning principal
  subspaces.
\newblock In \emph{International Conference on Artificial Intelligence and
  Statistics (AISTATS)}, 2023{\natexlab{a}}.

\bibitem[Le~Lan et~al.(2023{\natexlab{b}})Le~Lan, Tu, Rowland, Harutyunyan,
  Agarwal, Bellemare, and Dabney]{lelan23bootstrap}
Charline Le~Lan, Stephen Tu, Mark Rowland, Anna Harutyunyan, Rishabh Agarwal,
  Marc~G. Bellemare, and Will Dabney.
\newblock Bootstrapped representations in reinforcement learning.
\newblock In \emph{International Conference on Machine Learning (ICML)},
  2023{\natexlab{b}}.

\bibitem[Lipman et~al.(2023)Lipman, Chen, Ben{-}Hamu, Nickel, and
  Le]{lipman2022flow}
Yaron Lipman, Ricky T.~Q. Chen, Heli Ben{-}Hamu, Maximilian Nickel, and Matthew
  Le.
\newblock Flow matching for generative modeling.
\newblock In \emph{International Conference on Learning Representations
  ({ICLR})}, 2023.

\bibitem[Lipman et~al.(2024)Lipman, Havasi, Holderrieth, Shaul, Le, Karrer,
  Chen, Lopez{-}Paz, Ben{-}Hamu, and Gat]{lipman2024flow}
Yaron Lipman, Marton Havasi, Peter Holderrieth, Neta Shaul, Matt Le, Brian
  Karrer, Ricky T.~Q. Chen, David Lopez{-}Paz, Heli Ben{-}Hamu, and Itai Gat.
\newblock Flow matching guide and code.
\newblock \emph{CoRR}, abs/2412.06264, 2024.

\bibitem[Liu et~al.(2023)Liu, Gong, and Liu]{liu2022flow}
Xingchao Liu, Chengyue Gong, and Qiang Liu.
\newblock Flow straight and fast: Learning to generate and transfer data with
  rectified flow.
\newblock In \emph{International Conference on Learning Representations
  (ICLR)}, 2023.

\bibitem[Loshchilov and Hutter(2019)]{loshchilov19decoupled}
Ilya Loshchilov and Frank Hutter.
\newblock Decoupled weight decay regularization.
\newblock In \emph{International Conference on Learning Representations
  ({ICLR})}, 2019.

\bibitem[Machado et~al.(2018)Machado, Rosenbaum, Guo, Liu, Tesauro, and
  Campbell]{machado18eigopts}
Marlos~C. Machado, Clemens Rosenbaum, Xiaoxiao Guo, Miao Liu, Gerald Tesauro,
  and Murray Campbell.
\newblock Eigenoption discovery through the deep successor representation.
\newblock In \emph{International Conference on Learning Representations
  (ICLR)}, 2018.

\bibitem[Machado et~al.(2020)Machado, Bellemare, and Bowling]{machado20count}
Marlos~C. Machado, Marc~G. Bellemare, and Michael Bowling.
\newblock Count-based exploration with the successor representation.
\newblock In \emph{AAAI Conference on Artificial Intelligence}, 2020.

\bibitem[Machado et~al.(2023)Machado, Barreto, Precup, and
  Bowling]{machado23sr}
Marlos~C. Machado, Andr{\'e} Barreto, Doina Precup, and Michael Bowling.
\newblock Temporal abstraction in reinforcement learning with the successor
  representation.
\newblock \emph{Journal of Machine Learning Research (JMLR)}, 24:\penalty0
  80:1--80:69, 2023.

\bibitem[Misra(2019)]{misra19mish}
Diganta Misra.
\newblock Mish: A self regularized non-monotonic neural activation function.
\newblock \emph{CoRR}, abs/1908.08681, 2019.

\bibitem[Nachum et~al.(2019)Nachum, Chow, Dai, and Li]{nachum19dualdice}
Ofir Nachum, Yinlam Chow, Bo~Dai, and Lihong Li.
\newblock Dualdice: Behavior-agnostic estimation of discounted stationary
  distribution corrections.
\newblock In \emph{Neural Information Processing Systems ({NeurIPS})}, 2019.

\bibitem[Park et~al.(2024)Park, Kreiman, and Levine]{park24hilbert}
Seohong Park, Tobias Kreiman, and Sergey Levine.
\newblock Foundation policies with hilbert representations.
\newblock In \emph{International Conference on Machine Learning ({ICML})},
  2024.

\bibitem[Pathak et~al.(2017)Pathak, Agrawal, Efros, and
  Darrell]{pathak17curiosity}
Deepak Pathak, Pulkit Agrawal, Alexei~A. Efros, and Trevor Darrell.
\newblock Curiosity-driven exploration by self-supervised prediction.
\newblock In \emph{International Conference on Machine Learning (ICML)}, 2017.

\bibitem[Perez et~al.(2018)Perez, Strub, de~Vries, Dumoulin, and
  Courville]{perez18film}
Ethan Perez, Florian Strub, Harm de~Vries, Vincent Dumoulin, and Aaron~C.
  Courville.
\newblock Film: Visual reasoning with a general conditioning layer.
\newblock In \emph{AAAI Conference on Artificial Intelligence}, 2018.

\bibitem[Pirotta et~al.(2024)Pirotta, Tirinzoni, Touati, Lazaric, and
  Ollivier]{pirotta24bfmil}
Matteo Pirotta, Andrea Tirinzoni, Ahmed Touati, Alessandro Lazaric, and Yann
  Ollivier.
\newblock Fast imitation via behavior foundation models.
\newblock In \emph{International Conference on Learning Representations
  (ICLR)}, 2024.

\bibitem[Pooladian et~al.(2023)Pooladian, Ben-Hamu, Domingo-Enrich, Amos,
  Lipman, and Chen]{pooladian23multisample}
Aram-Alexandre Pooladian, Heli Ben-Hamu, Carles Domingo-Enrich, Brandon Amos,
  Yaron Lipman, and Ricky T.~Q. Chen.
\newblock Multisample flow matching: Straightening flows with minibatch
  couplings.
\newblock In \emph{International Conference on Machine Learning (ICML)}, 2023.

\bibitem[Precup et~al.(2000)Precup, Sutton, and Singh]{precup00eligibility}
Doina Precup, Richard~S. Sutton, and Satinder Singh.
\newblock Eligibility traces for off-policy policy evaluation.
\newblock In \emph{International Conference on Machine Learning {(ICML})},
  2000.

\bibitem[Precup et~al.(2001)Precup, Sutton, and Dasgupta]{precup01offpolicy}
Doina Precup, Richard~S. Sutton, and Sanjoy Dasgupta.
\newblock Off-policy temporal difference learning with function approximation.
\newblock In \emph{International Conference on Machine Learning (ICML)}, 2001.

\bibitem[Rezende and Mohamed(2015)]{rezende15nf}
Danilo Rezende and Shakir Mohamed.
\newblock Variational inference with normalizing flows.
\newblock In \emph{International Conference on Machine Learning ({ICML})},
  2015.

\bibitem[Rogozhnikov(2022)]{rogozhnikov22einops}
Alex Rogozhnikov.
\newblock Einops: Clear and reliable tensor manipulations with einstein-like
  notation.
\newblock In \emph{International Conference on Learning Representations
  (ICLR)}, 2022.

\bibitem[Ronneberger et~al.(2015)Ronneberger, Fischer, and
  Brox]{ronneberger15unet}
Olaf Ronneberger, Philipp Fischer, and Thomas Brox.
\newblock U-net: Convolutional networks for biomedical image segmentation.
\newblock In \emph{Medical Image Computing and Computer-Assisted Intervention
  (MICCAI)}, volume 9351, pages 234--241, 2015.

\bibitem[Rubner et~al.(2000)Rubner, Tomasi, and Guibas]{RubnerTG00}
Yossi Rubner, Carlo Tomasi, and Leonidas~J. Guibas.
\newblock The earth mover's distance as a metric for image retrieval.
\newblock \emph{International Journal of Computer Vision}, 40\penalty0
  (2):\penalty0 99--121, 2000.

\bibitem[Schmidhuber(1991)]{schmidhuber91apossibility}
J\"urgen Schmidhuber.
\newblock A possibility for implementing curiosity and boredom in
  model-building neural controllers.
\newblock In \emph{International Conference on Simulation of Adaptive
  Behavior}, 1991.

\bibitem[Schramm and Boularias(2024)]{schramm24belldiff}
Liam Schramm and Abdeslam Boularias.
\newblock Bellman diffusion models.
\newblock \emph{CoRR}, abs/2407.12163, 2024.

\bibitem[Schrittwieser et~al.(2020)Schrittwieser, Antonoglou, Hubert, Simonyan,
  Sifre, Schmitt, Guez, Lockhart, Hassabis, Graepel, Lillicrap, and
  Silver]{schrittwieser20mastering}
Julian Schrittwieser, Ioannis Antonoglou, Thomas Hubert, Karen Simonyan,
  Laurent Sifre, Simon Schmitt, Arthur Guez, Edward Lockhart, Demis Hassabis,
  Thore Graepel, Timothy Lillicrap, and David Silver.
\newblock Mastering atari, go, chess and shogi by planning with a learned
  model.
\newblock \emph{Nature}, 588\penalty0 (7839):\penalty0 604--609, 2020.

\bibitem[Shi et~al.(2023)Shi, De~Bortoli, Campbell, and Doucet]{shi23dsbm}
Yuyang Shi, Valentin De~Bortoli, Andrew Campbell, and Arnaud Doucet.
\newblock Diffusion schrödinger bridge matching.
\newblock In \emph{Neural Information Processing Systems (NeurIPS)}, 2023.

\bibitem[Sikchi et~al.(2021)Sikchi, Zhou, and Held]{sikchi21learning}
Harshit Sikchi, Wenxuan Zhou, and David Held.
\newblock Learning off-policy with online planning.
\newblock In \emph{Conference on Robot Learning (CoRL)}, 2021.

\bibitem[Silver et~al.(2016)Silver, Huang, Maddison, Guez, Sifre, van~den
  Driessche, Schrittwieser, Antonoglou, Panneershelvam, Lanctot, Dieleman,
  Grewe, Nham, Kalchbrenner, Sutskever, Lillicrap, Leach, Kavukcuoglu, Graepel,
  and Hassabis]{silver16mastering}
David Silver, Aja Huang, Chris~J. Maddison, Arthur Guez, Laurent Sifre, George
  van~den Driessche, Julian Schrittwieser, Ioannis Antonoglou, Vedavyas
  Panneershelvam, Marc Lanctot, Sander Dieleman, Dominik Grewe, John Nham, Nal
  Kalchbrenner, Ilya Sutskever, Timothy~P. Lillicrap, Madeleine Leach, Koray
  Kavukcuoglu, Thore Graepel, and Demis Hassabis.
\newblock Mastering the game of go with deep neural networks and tree search.
\newblock \emph{Nature}, 529\penalty0 (7587):\penalty0 484–489, 2016.

\bibitem[Silver et~al.(2017)Silver, Schrittwieser, Simonyan, Antonoglou, Huang,
  Guez, Hubert, Baker, Lai, Bolton, Chen, Lillicrap, Hui, Sifre, van~den
  Driessche, Graepel, and Hassabis]{silver17go}
David Silver, Julian Schrittwieser, Karen Simonyan, Ioannis Antonoglou, Aja
  Huang, Arthur Guez, Thomas Hubert, Lucas Baker, Matthew Lai, Adrian Bolton,
  Yutian Chen, Timothy Lillicrap, Fan Hui, Laurent Sifre, George van~den
  Driessche, Thore Graepel, and Demis Hassabis.
\newblock Mastering the game of go without human knowledge.
\newblock \emph{Nature}, 550\penalty0 (7676):\penalty0 354--359, 2017.

\bibitem[Sohl-Dickstein et~al.(2015)Sohl-Dickstein, Weiss, Maheswaranathan, and
  Ganguli]{sohl2015deep}
Jascha Sohl-Dickstein, Eric Weiss, Niru Maheswaranathan, and Surya Ganguli.
\newblock Deep unsupervised learning using nonequilibrium thermodynamics.
\newblock In \emph{International Conference on Machine Learning ({ICML})},
  2015.

\bibitem[Song et~al.(2021{\natexlab{a}})Song, Meng, and Ermon]{song21ddim}
Jiaming Song, Chenlin Meng, and Stefano Ermon.
\newblock Denoising diffusion implicit models.
\newblock In \emph{International Conference on Learning Representations
  ({ICLR})}, 2021{\natexlab{a}}.

\bibitem[Song and Ermon(2019)]{song2019generative}
Yang Song and Stefano Ermon.
\newblock Generative modeling by estimating gradients of the data distribution.
\newblock In \emph{Neural Information Processing Systems ({NeurIPS})}, 2019.

\bibitem[Song et~al.(2021{\natexlab{b}})Song, Sohl{-}Dickstein, Kingma, Kumar,
  Ermon, and Poole]{song21sde}
Yang Song, Jascha Sohl{-}Dickstein, Diederik~P. Kingma, Abhishek Kumar, Stefano
  Ermon, and Ben Poole.
\newblock Score-based generative modeling through stochastic differential
  equations.
\newblock In \emph{International Conference on Learning Representations
  ({ICLR})}, 2021{\natexlab{b}}.

\bibitem[Song et~al.(2023)Song, Dhariwal, Chen, and
  Sutskever]{song23consistency}
Yang Song, Prafulla Dhariwal, Mark Chen, and Ilya Sutskever.
\newblock Consistency models.
\newblock In \emph{International Conference on Machine Learning ({ICML})},
  2023.

\bibitem[Stadie et~al.(2016)Stadie, Levine, and Abbeel]{stadie16incentivizing}
Bradly~C. Stadie, Sergey Levine, and Pieter Abbeel.
\newblock Incentivizing exploration in reinforcement learning with deep
  predictive models.
\newblock In \emph{International Conference on Learning Representations
  (ICLR)}, 2016.

\bibitem[Sutton(1991)]{sutton91dyna}
Richard~S. Sutton.
\newblock Dyna, an integrated architecture for learning, planning, and
  reacting.
\newblock \emph{ACM SIGART}, 2\penalty0 (4):\penalty0 160–163, 1991.

\bibitem[Talvitie(2014)]{talvitie14model}
Erin Talvitie.
\newblock Model regularization for stable sample rollouts.
\newblock In \emph{Conference on Uncertainty in Artificial Intelligence
  ({UAI})}, 2014.

\bibitem[Thakoor et~al.(2022)Thakoor, Rowland, Borsa, Dabney, Munos, and
  Barreto]{thakoor22ghm}
Shantanu Thakoor, Mark Rowland, Diana Borsa, Will Dabney, R{\'e}mi Munos, and
  Andr{\'e} Barreto.
\newblock Generalised policy improvement with geometric policy composition.
\newblock In \emph{International Conference on Machine Learning (ICML)}, 2022.

\bibitem[Tirinzoni et~al.(2025)Tirinzoni, Touati, Farebrother, Guzek,
  Kanervisto, Xu, Lazaric, and Pirotta]{tirinzoni25zeroshot}
Andrea Tirinzoni, Ahmed Touati, Jesse Farebrother, Mateusz Guzek, Anssi
  Kanervisto, Yingchen Xu, Alessandro Lazaric, and Matteo Pirotta.
\newblock Zero-shot whole-body humanoid control via behavioral foundation
  models.
\newblock In \emph{International Conference on Learning Representations
  ({ICLR})}, 2025.

\bibitem[Todorov et~al.(2012)Todorov, Erez, and Tassa]{todorov2012mujoco}
Emanuel Todorov, Tom Erez, and Yuval Tassa.
\newblock Mujoco: A physics engine for model-based control.
\newblock In \emph{International Conference on Intelligent Robots and Systems
  ({IROS})}, 2012.

\bibitem[Tomar et~al.(2024)Tomar, Hansen-Estruch, Bachman, Lamb, Langford,
  Taylor, and Levine]{tomar2024video}
Manan Tomar, Philippe Hansen-Estruch, Philip Bachman, Alex Lamb, John Langford,
  Matthew~E Taylor, and Sergey Levine.
\newblock Video occupancy models.
\newblock \emph{CoRR}, abs/2407.09533, 2024.

\bibitem[Tong et~al.(2024)Tong, Fatras, Malkin, Huguet, Zhang, Rector-Brooks,
  Wolf, and Bengio]{tong24improving}
Alexander Tong, Kilian Fatras, Nikolay Malkin, Guillaume Huguet, Yanlei Zhang,
  Jarrid Rector-Brooks, Guy Wolf, and Yoshua Bengio.
\newblock Improving and generalizing flow-based generative models with
  minibatch optimal transport.
\newblock In \emph{Transactions on Machine Learning Research (TMLR)}, 2024.

\bibitem[Touati and Ollivier(2021)]{touati21fwdbwd}
Ahmed Touati and Yann Ollivier.
\newblock Learning one representation to optimize all rewards.
\newblock In \emph{Neural Information Processing Systems (NeurIPS)}, 2021.

\bibitem[Touati et~al.(2023)Touati, Rapin, and Ollivier]{touati23zeroshot}
Ahmed Touati, J{\'e}r{\'e}my Rapin, and Yann Ollivier.
\newblock Does zero-shot reinforcement learning exist?
\newblock In \emph{International Conference on Learning Representations
  (ICLR)}, 2023.

\bibitem[Tunyasuvunakool et~al.(2020)Tunyasuvunakool, Muldal, Doron, Liu,
  Bohez, Merel, Erez, Lillicrap, Heess, and Tassa]{tunyasuvunakool20dmc}
Saran Tunyasuvunakool, Alistair Muldal, Yotam Doron, Siqi Liu, Steven Bohez,
  Josh Merel, Tom Erez, Timothy Lillicrap, Nicolas Heess, and Yuval Tassa.
\newblock dm{\_}control: Software and tasks for continuous control.
\newblock \emph{Software Impacts}, 6:\penalty0 100022, 2020.

\bibitem[van~den Oord et~al.(2017)van~den Oord, Vinyals, and
  Kavukcuoglu]{oord17neural}
Aäron van~den Oord, Oriol Vinyals, and Koray Kavukcuoglu.
\newblock Neural discrete representation learning.
\newblock In \emph{Neural Information Processing Systems (NeurIPS)}, 2017.

\bibitem[Vincent(2011)]{vincent2011connection}
Pascal Vincent.
\newblock A connection between score matching and denoising autoencoders.
\newblock \emph{Neural Computation}, 23\penalty0 (7):\penalty0 1661--1674,
  2011.

\bibitem[Waskom(2021)]{waskom2021seaborn}
Michael~L. Waskom.
\newblock Seaborn: Statistical data visualization.
\newblock \emph{Journal of Open Source Software}, 6\penalty0 (60):\penalty0
  3021, 2021.

\bibitem[Wiltzer et~al.(2024{\natexlab{a}})Wiltzer, Farebrother, Gretton, and
  Rowland]{wiltzer24mvdrl}
Harley Wiltzer, Jesse Farebrother, Arthur Gretton, and Mark Rowland.
\newblock Foundations of multivariate distributional reinforcement learning.
\newblock In \emph{Neural Information Processing Systems (NeurIPS)},
  2024{\natexlab{a}}.

\bibitem[Wiltzer et~al.(2024{\natexlab{b}})Wiltzer, Farebrother, Gretton, Tang,
  Barreto, Dabney, Bellemare, and Rowland]{wiltzer24dsm}
Harley Wiltzer, Jesse Farebrother, Arthur Gretton, Yunhao Tang, André Barreto,
  Will Dabney, Marc~G. Bellemare, and Mark Rowland.
\newblock A distributional analogue to the successor representation.
\newblock In \emph{International Conference on Machine Learning (ICML)},
  2024{\natexlab{b}}.

\bibitem[Wu et~al.(2025)Wu, Chen, Swamy, Brantley, and Sun]{wu2025diffusing}
Runzhe Wu, Yiding Chen, Gokul Swamy, Kiant{\'e} Brantley, and Wen Sun.
\newblock Diffusing states and matching scores: A new framework for imitation
  learning.
\newblock In \emph{International Conference on Learning Representations
  ({ICLR})}, 2025.

\bibitem[Yang et~al.(2024)Yang, Zhang, Zhang, Liu, Xu, Zhang, Meng, Ermon, and
  Cui]{yang24consistency}
Ling Yang, Zixiang Zhang, Zhilong Zhang, Xingchao Liu, Minkai Xu, Wentao Zhang,
  Chenlin Meng, Stefano Ermon, and Bin Cui.
\newblock Consistency flow matching: Defining straight flows with velocity
  consistency.
\newblock \emph{CoRR}, abs/2407.02398, 2024.

\bibitem[Yarats et~al.(2022)Yarats, Brandfonbrener, Liu, Laskin, Abbeel,
  Lazaric, and Pinto]{yarats22exorl}
Denis Yarats, David Brandfonbrener, Hao Liu, Michael Laskin, Pieter Abbeel,
  Alessandro Lazaric, and Lerrel Pinto.
\newblock Don’t change the algorithm, change the data: Exploratory data for
  offline reinforcement learning.
\newblock \emph{CoRR}, abs/2201.13425, 2022.

\bibitem[Zhang et~al.(2021)Zhang, Chen, Zhao, Xiong, Qin, and
  Liu]{zhang21multidim}
Pushi Zhang, Xiaoyu Chen, Li~Zhao, Wei Xiong, Tao Qin, and Tie-Yan Liu.
\newblock Distributional reinforcement learning for multi-dimensional reward
  functions.
\newblock In \emph{Neural Information Processing Systems (NeurIPS)}, 2021.

\bibitem[Zhu et~al.(2024)Zhu, Wang, Han, Du, and Gupta]{zhu24gom}
Chuning Zhu, Xinqi Wang, Tyler Han, Simon~S. Du, and Abhishek Gupta.
\newblock Distributional successor features enable zero-shot policy
  optimization.
\newblock In \emph{Neural Information Processing Systems ({NeurIPS})}, 2024.

\end{thebibliography}

\clearpage
\newpage
\beginappendix

\vspace{-0.75cm}
\begin{appendices}

\startcontents[sections]
\printcontents[sections]{l}{1}{\setcounter{tocdepth}{2}}

\section{Related Work}\label{sec:related-work}

The Successor Representation \citep{dayan93sr} was originally proposed for tabular MDPs and was later generalized to continuous state spaces with the Successor Measure \citep{blier21successor}.
Successor Features \citep{barreto2017sf,andre2020fast} extends these ideas by instead modeling the evolution of multi-dimensional features assuming rewards decompose linearly over these features.
Prior works have leveraged these methods for zero-shot policy evaluation \citep{dayan93sr,barreto2017sf,wiltzer24dsm}, zero-shot policy optimization \citep{borsa19usfa,touati21fwdbwd,touati23zeroshot,park24hilbert,zhu24gom,cetin2024finer,tirinzoni25zeroshot}, imitation learning \citep{pirotta24bfmil,jain25nonadversarial}, exploration \citep{machado20count,jain23maximum}, representation learning \citep{lelan22generalization,lelan23subspace,lelan23bootstrap,farebrother2023pvn,ghosh23icvf}, and building temporal abstractions \citep{machado18eigopts,machado23sr}.

\citet{janner20gmodel} originally proposed a method to learn a generative model of the successor measure with modeling techniques spanning from Generative Adversarial Networks \citep{goodfellow14gan} to Normalizing Flows \citep{dnih15nice,rezende15nf} like RealNVP \citep{dnih17nvp}.
Followup work \citep[e.g.,][]{thakoor22ghm,tomar2024video} explored other generative modeling techniques including various types of auto-encoders \citep[e.g.,][]{higgins17beta,oord17neural}.
Also of note is recent work learning generative models of multi-dimensional cumulants including features \citep{wiltzer24mvdrl,zhu24gom} and multi-variate reward functions \citep{zhang21multidim}.
Prior work by \citet{wiltzer24dsm} sought to deal with the instability of long-horizon predictions in GHMs by employing an $n$-step mixture distribution where they sample $t \sim \text{Geometric}(1 - \gamma)$ and bootstrap if $t > n$; otherwise returning the state at time $t$ along the trajectory. Without resorting to importance sampling this approach is limited to the on-policy setting.
Finally, most closely related to our work is that of \citet{schramm24belldiff} who provide a preliminary and limited derivation of what we term \tddshort.
In contrast, our work not only rigorously formalizes and significantly extends these ideas but also integrates them into the more general flow-matching framework \citep{lipman2022flow,lipman2024flow}, additionally incorporating extensions to score-matching \citep{song21sde,song21sde} and diffusion \citep{sohl2015deep,ho20ddpm}.
Moreover, we conduct an extensive empirical analysis, demonstrating the efficacy of our approach --- an aspect notably absent from \citet{schramm24belldiff}.

\section{Extension to Score Matching and Diffusion Models}\label{app:diffusion}
This section extends our framework to score matching and denoising diffusion models. We leverage the unification of these methods under stochastic differential equations \citep{song21sde} introducing an analogous class of Temporal Difference Diffusion methods.

\subsection{Background}

Both score-based generative modeling~\citep{song2019generative} and diffusion probabilistic modeling~\citep{sohl2015deep,ho20ddpm} can be unified under the framework of stochastic differential equations (SDE) introduced in~\citet{song21sde}. Unlike in flow-matching, time is inverted in diffusion models and ranges from time 0 to $T$.
Given the data distribution $q_0$ and prior simple distribution $q_T$ (the ``noise'' distribution), we construct a diffusion process $\{ X_t \}_{t \in [0, T]}$ such that $X_0 \sim q_0$ and $X_T \sim q_T$. This diffusion can be modeled as the solution to an Ito SDE: 
\begin{equation} \label{eq:forward-SDE}
    \mathrm{d}X_t = f(t)\, X_t\, \mathrm{d}t + g(t)\, \mathrm{d}W_t \,\mid\, X_0 \sim q_0\,,
\end{equation}
where $W_t$ is a standard Brownian motion and $f: [0, T] \rightarrow \mathbb{R}^d$ is scalar function called the drift coefficient, and $g: [0, T] \rightarrow \mathbb{R}$ is scalar function  known as diffusion coefficient.

Generating samples from $X_0 \sim q_0$ consists in sampling $X_T \sim q_T$ and reversing the forward-SDE process in~\eqref{eq:forward-SDE}. A known result from \citet{anderson1982reverse} states that the reverse of a diffusion process is also a diffusion process, running backward in time and given by the reverse-time SDE:
\begin{equation}~\label{eq:reverse-SDE}
\mathrm{d}X_t = \Big( f(t)\,X_t - g(t)^2\, \nabla_{X_t} \log q_t(X_t) \Big) \mathrm{d}t + g(t)\, \mathrm{d}\widebar{W}_t \,\mid\, X_T \sim q_T\,,
\end{equation}
where $\widebar{W}_t$ is a Brownian motion when time flows backwards from $T$ to 0, $\mathrm{d}t$ is an infinitesimal negative timestep and $q_t$ is the marginal distribution of $X_t$. Therefore, once we learn the score of the marginal distribution $\nabla_x \log q_t(x)$, we can sample from $q_0$ by simulating the reverse diffusion 
process~\eqref{eq:reverse-SDE}.

To estimate $\nabla_x \log q_t(x)$, we can train a time-dependent score-based model $\tilde{\scorefn}(\cdots; \theta): [0, T] \times \mathbb{R}^d \rightarrow \mathbb{R}^d$ via the denoising diffusion / score matching objective~\citep{vincent2011connection, song2019generative}:
\begin{equation}
\ell_{\mathrm{\textsc{dd}}}(\theta) = \mathbb{E}_{t \sim \mathcal{U}([0, 1]), X_0 \sim q_0} \mathbb{E}_{X_t \sim q_{t\mid 0}(\cdot \mid X_0)} \Big[ \big\| \tilde{\scorefn}_t(X_t;\theta) - \nabla_{X_t} \log q_{t\mid 0}(X_t \mid X_0)\big{\|}^2 \Big]\,.
\end{equation}
For $\ell_{\mathrm{\textsc{dd}}}$ to be tractable, we need to know the conditional probability $q_{t\mid 0}$. Usually, specific choices of the drift and diffusion coefficients $f_t$ and $g_t$ are used such that  $q_{t\mid 0}$ is always a Gaussian distribution $\mathcal{N}(\cdot \mid \alpha_t x_0, \sigma^2_t)$, where the mean $\alpha_t$ and variance $\sigma^2_t$ can be computed in closed-form. The global minimizer of $\ell_{\mathrm{\textsc{dd}}}(\theta)$ denoted by $\scorefn_t^{\star}(x)$ is equal to the score function $\nabla_x \log q_t(x)$, thanks to the following proposition: 

\begin{metaframe}
\begin{proposition}[\citealt{vincent2011connection}] \label{prop: sm equivalence} Let $q_t(x) = \int q_0(x_0) q_{t|0}(x | x_0)\, \mathrm{d}x_0$, then we have:
\begin{equation}
    \nabla_\theta\, \ell_{\mathrm{\textsc{dd}}}(\theta) = \nabla_\theta\, \mathbb{E}_{t, X_t \sim q_t} \Big[ \big\| \tilde{\scorefn}_t(X_t; \theta) - \nabla_{X_t} \log q_{t}(X_t)\big{\|}^2 \Big] \, .
\end{equation}
\end{proposition}
\end{metaframe}

\subsection{Temporal Difference Diffusion}
 To learn a predictive model of $m^\pi$ using diffusion from an offline dataset, we follow a similar approach to what we presented in \autoref{sec:td-flow} and we define an iterative process starting from initial weights $\theta^{(0)}$ and at each iteration minimizing the Temporal-Difference Denoising Diffusion (\vtddshort) loss:

 \begin{equation}
 \begin{gathered}
\ell_{\mathrm{\vtddshort}}(\theta) = \mathbb{E}_{\rho,t,X_0,X_t}
\Big[
\big\|
\tilde{\scorefn}_t(X_t\mid S,A; \theta) - \nabla_x \log q_{t\mid 0}(X_t \mid X_1)
\big{\|}^2
\Big]\,, \\
\text{where}\,,\,\, X_0 \sim \Tpi{\ghmdistbar_{0\mid T}^{(n)}}(\cdot\mid S, A),\,
X_t \sim q_{t\mid 0}(\cdot\mid X_0) \,.
\end{gathered}
\owntag{\text{\vtddshort}; }
\end{equation}
In order to sample $X_0 \sim \Tpi{\ghmdistbar^{(n)}_{0|T}}(\cdot\,|\,s,a)$, with probability $1 - \gamma$, we return the successor state $S' \sim P(\cdot \mid S, A)$. Otherwise, with probability $\gamma$ we solve the following reverse-time SDE from $X_T$ using the score $\tilde{\scorefn}_t^{(n)}$,
\begin{equation}
\mathrm{d}X_t = \Big( f(t)\, X_t - g(t)^2 \tilde{\scorefn}_t^{(n)}(X_t \mid S, A) \Big)\, \mathrm{d}t + g(t) \mathrm{d}\widebar{W}_t \, .
\end{equation}
Minimizing $\ell_{\mathrm{\vtddshort}}(\theta)$ leads to score function $\tilde{\scorefn}^{(n+1)}_t(s\mid s,a)$ generating a marginal probability $q_t^{(n+1)}$ that approximates $\mathcal{T}^\pi q_0^{(n)}$ at $t=0$.

Following the $\tdfshort$ blueprint, we can further exploit the structure of the target bootstrapped distribution to design an improved diffusion process that converts Gaussian noise to $\mathcal{T}^\pi q^{(n)}_0$.
First, we show below that the mixture of a diffusion process is also a diffusion process with modified drift and diffusion functions.
\begin{metaframe}
\begin{lemma}\label{lemma:mixture_sde}
    Consider two diffusion processes with drift functions $\onestepterm{f}$ and $\bootterm{f}$, sharing the same diffusion coefficient $g$:
    \begin{align*}
        \mathrm{d}X_t & = \onestepterm{f}_t(X_t)\, \mathrm{d}t + g(t)\, \mathrm{d}W \\
        \mathrm{d}X_t & = \bootterm{f}_t(X_t)\, \mathrm{d}t + g(t)\, \mathrm{d}W \, .
    \end{align*}
Let $\onestepterm{q}_t$ and $\bootterm{q}_t$ be their marginal distribution, then the diffusion process corresponding to the mixture marginal distribution $q_t = (1-\gamma) \onestepterm{q}_t + \gamma \bootterm{q}_t$ is: 
        \begin{align*}
        \mathrm{d}X_t & = \frac{(1-\gamma)\onestepterm{q}_t \onestepterm{f}_t + \gamma  \bootterm{q}_t \bootterm{f}_t}{(1-\gamma) \onestepterm{q}_t + \gamma \bootterm{q}_t }(X_t)\,  \mathrm{d}t + g(t)\, \mathrm{d}W \,.
    \end{align*}
\end{lemma}
\end{metaframe}
\begin{proof}
    The marginal probabilities $\onestepterm{p}$ and  $\bootterm{p}$ are characterized by the Fokker-Planck equations: 
    \begin{align*}
        \frac{\partial \onestepterm{p}_t}{\partial t} & = -  \mathrm{div} (\onestepterm{p}_t \onestepterm{f}_t) + \frac{g_t^2}{2} \Delta \onestepterm{p}_t \\
        \frac{\partial \bootterm{p}_t}{\partial t} & = -  \mathrm{div} (\bootterm{p}_t \bootterm{f}_t) + \frac{g_t^2}{2} \Delta \bootterm{p}_t \\
    \end{align*}
where $\mathrm{div}$ is the divergence operator and $\Delta = \mathrm{div} \nabla$ is the Laplace operator.
Therefore,
\begin{align*}
    \frac{\partial p_t}{\partial t} & = (1-\gamma)\frac{\partial \onestepterm{p}_t}{\partial t} + \gamma \frac{\partial \bootterm{p}_t}{\partial t}  \\
    & = -  \mathrm{div} (\onestepterm{p}_t \onestepterm{f}_t) + \frac{g_t^2}{2} \Delta \onestepterm{p}_t -  \mathrm{div} (\bootterm{p}_t \bootterm{f}_t) + \frac{g_t^2}{2} \Delta \bootterm{p}_t \\
    & = - \mathrm{div} \left((1-\gamma) \onestepterm{p}_t \onestepterm{f}_t + \gamma \bootterm{p}_t \bootterm{f}_t \right) + \frac{g_t^2}{2} \Delta \left( (1-\gamma) \onestepterm{p}_t + \gamma \bootterm{p}_t \right) \\
    & = \mathrm{div} \left(p_t \frac{(1-\gamma) \onestepterm{p}_t \onestepterm{f}_t + \gamma \bootterm{p}_t \bootterm{f}_t)}{(1-\gamma) \onestepterm{p}_t + \gamma \bootterm{p}_t} \right) + \frac{g_t^2}{2} \Delta p_t \,.
\end{align*}
The drift $\frac{(1-\gamma) \onestepterm{p}_t \onestepterm{f}_t + \gamma \bootterm{p}_t \bootterm{f}_t}{(1-\gamma) \onestepterm{p}_t + \gamma \bootterm{p}_t}$ and the diffusion coefficient $g_t$ satisfy the Fokker-Planck equation with the probability path $p_t$, and therefore their associated diffusion process generate $p_t$.
\end{proof}
\autoref{lemma:mixture_sde} can be easily extended to the case of a continuous mixture of diffusion processes.

This result shows that it is possible to use two independent diffusion processes for the two terms in the sampling process induced by the Bellman operator. For the first, we can use the standard noising diffusion~process:
\begin{align*}
\onestepterm{q}_t(x\mid s,a) = \int q_{t\mid 0}(x\mid s')P(\mathrm{d}s'\mid s,a)\,,
\end{align*}
where we sample $X_t \sim q_{t\mid 0}(\cdot \mid s')$ by simulating a simple forward diffusion process \eqref{eq:forward-SDE}. %
For the second term, we can leverage the GHM $\ghmdist_t^{(n)}$ at the previous iteration to construct the process, %
\begin{align*}
\bootterm{q}_t^{(n)}&(x\mid s,a) = \int \ghmdist_t^{(n)}(x\mid s',\pi(s'))\, P(\mathrm{d}s'\mid s,a) \,,
\end{align*}
where $\ghmdist_t^{(n)}(x\mid s',a')$ is the marginal probability of the reverse SDE induced by the score $\scorefn^{(n)}$,
\begin{equation*}
    \mathrm{d}X_t = \left( f(t)\, X_t - g(t)^2\, \scorefn_t^{(n)}(X_t \mid s, a) \right)\, \mathrm{d}t + g(t)\, \mathrm{d}\widebar{W}_t \, .
\end{equation*}
Additionally, $\bootterm{q}_t^{(n)}(x\mid s,a)$, as continuous mixture of diffusion's marginals $\ghmdist_t^{(n)}(x\mid s',\pi(s'))$ weighted by $P(s' \mid s, a)$, can be generated by the diffusion process,
\begin{equation*}
\begin{gathered}
    \mathrm{d}X_t = \big( f(t)\, X_t - g(t)^2  \, \bootterm{\scorefn}_t(X_t \mid s, a) \big)\,\mathrm{d}t + g(t)\, \mathrm{d}\widebar{W}_t,\,\text{where} \\[0.5\baselineskip]
    \bootterm{\scorefn}_t(x_t \mid s, a) = \frac{\int P(\mathrm{d}s' \mid s,a)\, q_t^{(n)}(x\mid s',\pi(s'))\, \scorefn_t^{(n)}(x_t \mid s', \pi(s'))}{\int P(\mathrm{d}s'\mid s,a)\, q_t^{(n)}(x\mid s',\pi(s'))} \, .
\end{gathered}
\end{equation*}

Given these two diffusion processes, the target probability $q^{(n+1)}_t = (1-\gamma) \onestepterm{q}_t + \gamma \bootterm{q}^{(n)}_t$ can be generated by the following reverse SDE,
\begin{equation*}
    \mathrm{d}X_t = \left( f(t) X_t - g(t)^2\, \scorefn^{(n+1)}_t(X_t \mid s, a) \right)\,\mathrm{d}t + g(t)\, \mathrm{d}\widebar{W}_t,
\end{equation*}
where $\scorefn^{(n+1)}_t(x \mid s, a) = \frac{(1-\gamma) \onestepterm{q}_t \nabla_{x} \log \onestepterm{q}_t+ \gamma \bootterm{q}^{(n)}_t \bootterm{\scorefn}^{(n)}_t }{(1-\gamma) \onestepterm{q}_t + \gamma \bootterm{q}^{(n)}_t}(x \mid s, a)$. Therefore, we can learn $\tilde{\scorefn}_t(\cdots; \theta)$ to approximate $\scorefn^{(n+1)}_t$ by  minimizing the loss,
\begin{align}
    \ell(\theta) 
    & = (1-\gamma) 
    \mathbb{E}_{\rho,t,X_t \sim \onestepterm{q}_t(\cdot \mid S, A)}\Big[\big \| \tilde{\scorefn}(X_t \mid S, A; \theta) - \nabla_{X_t} \log \onestepterm{q}_t(X_t \mid S, A)\big \|^2\Big] \\
    & \qquad + \gamma 
    \mathbb{E}_{\rho,t,X_t \sim \bootterm{q}_t^{(n)}(\cdot \mid S,A)}\Big[\big \| \tilde{\scorefn}(X_t \mid S, A; \theta) -  \bootterm{\scorefn}_t^{(n)}(X_t \mid S,A)\big \|^2\Big].\nonumber
\end{align}

We can simplify the first term via \autoref{prop: sm equivalence} (since $\onestepterm{q}_t(x|s,a) = \int q_{t|0}(x|s')P(\mathrm{d}s'|s,a)$), hence we have 
\begin{align*}
    \nabla_\theta\, \mathbb{E}_{\rho,t,X_t \sim \onestepterm{q}_t(\cdot \mid s, a)}\Big[\big \| \tilde{\scorefn}(X_t \mid s, a; \theta)& - \nabla_{X_t} \log \onestepterm{q}_t(X_t \mid S, A)\big \|^2\Big] = \\
    &\nabla_\theta\, \mathbb{E}_{\rho,t,X_t \sim q_{t\mid0}(\cdot \mid S')}\Big[\big \| \tilde{\scorefn}(X_t \mid S, A; \theta) - \nabla_{X_t} \log q_{t\mid 0}(X_t \mid S') \big \|^2\Big] \,.
\end{align*}
Moreover, using a similar argument for equivalence between the gradient of marginal and conditional flow-matching objectives, we can show that
\begin{align*}
    \nabla_\theta\, \mathbb{E}_{\rho,t,X_t \sim \bootterm{q}_t^{(n)}(\cdot \mid S,A)}\Big[\big \| \tilde{\scorefn}(X_t \mid S, A; \theta)& -  \bootterm{\scorefn}_t^{(n)}(X_t \mid S,A)\big \|^2\Big] = \\
    &\nabla_\theta\, \mathbb{E}_{\rho,t, X_T \sim q_T, X_t \sim q_{t\mid T}^{n}(\cdot \mid s,a)}\Big[\big \| \tilde{\scorefn}(X_t \mid S, A; \theta) -  \scorefn_t^{(n)}(X_t \mid S,A)\big \|^2\Big] \,.
\end{align*}
This leads us to the final \tddshort loss function,
\begin{align}
    \ell_{\mathrm{\tddshort}}(\theta) & = (1-\gamma) \mathbb{E}_{\rho,t,X_t \sim q_{t\mid 0}(\cdot\mid S')}
\Big[
\big\|
\tilde{\scorefn}_t(X_t\,|\,S, A; \theta) - \nabla_x \log p_{t\mid 0}(X_t \mid S')
\big{\|}^2
\Big] \\
& + \gamma \mathbb{E}_{\rho,t, X_t \sim q^{(n)}_{t\mid T}(\cdot \mid S', \pi(S'))}
\Big[
\big\|
\tilde{\scorefn}(X_t\mid S, A; \theta) - \tilde{\scorefn}^{(n)}_t(X_t\mid S',\pi(S')
\big{\|}^2
\Big]\,. \nonumber
\end{align}

\clearpage

\section{Experimental Details}\label{app:experimental.details}

\subsection{Evaluation}\label{app:evaluation}

\begin{wraptable}{r}{0.5\textwidth}
\vspace{-5mm}
\caption{Evaluation hyper-parameters for both single and multi-policy experiments.}
\label{table:eval-hparams}
\renewcommand{\arraystretch}{1.25}
\centering
\resizebox{\linewidth}{!}{
\begin{tabular}{@{}cll@{}}
     \toprule
     {\fontsize{11}{13}\selectfont \textbf{Evaluation}}
     & {\fontsize{11}{13}\selectfont \textbf{Hyperparameter}}
     & {\fontsize{11}{13}\selectfont \textbf{Value}} \\
     \midrule
     \multirow{4}{*}{EMD} & Number of states $s_0$ & $64$\\
     & Number of $m$-samples per state & $2048$\\
     & Number of episodes per state & $1$\\
     &Episode length & $1000$\\ \midrule
     \multirow{4}{*}{MSE(V)} & Number of state $s_0$ & $64$\\
     & Number of GHM-samples per state & $2048$\\
     & Number of episodes per state & $1$\\
     & Episode length & $1000$\\
     \midrule
     \multirow{3}{*}{GPI} & Number of $z$ samples & $256$\\
     & Number of GHM samples & $128$\\
     & Number of FB inference samples & $250,000$\\
     \bottomrule
\end{tabular}

}
\end{wraptable}
\par

Evaluating a GHM can be challenging, TD-based losses employing bootstrapping do not provide a good signal as to the quality of the learned model.
Instead, we opt to measure 1) the likelihood of a trajectory coming from the true discounted occupancy of a given policy, 2) the Earth Mover's Distance \citep[EMD;][]{RubnerTG00} between samples from the true occupancy and our GHM which provides an estimate of the distance between these two probability distributions, and 3) the value-function approximation error. 
In all cases, to obtain samples from the true discounted occupancy, we collect trajectories $\{ (s_0, s_1, \dots, s_T) \}_{i=1}^N$ from policy $\pi$ and subsequently resample states according to $t \sim \mathrm{Geometic}(1 - \gamma)$ for a particular discount factor $\gamma \in [0, 1)$.
Armed with samples from $m^{\pi}$ we compute the aforementioned metrics following the procedures stated below along with the parameter values outlined in \autoref{table:eval-hparams}.

\textbf{Normalized Negative Log-Likelihood.} 
To compute the log-likelihood of our flow matching and diffusion methods, we take advantage of the following change in variables formula \citep{dnih15nice,rezende15nf,chen18neuralode},
$$
\log{\left(\ghmdistbar(x_1 \,|\, s, a; \theta)\right)} = \log{\varphi(x_0)} + \int_{0}^1 \frac{\partial \log{\left( \ghmdistbar(x_t\,|\,s,a; \theta)\right)}}{\partial x_t}\,\, dt \, ,
$$
where $\varphi$ is the probability density function of a standard Gaussian distribution, which acts as the prior on $x_0$.
The change in log density over time can be written as the following differential equation called the instantaneous change of variables formula \citep[][Theorem 1]{chen18neuralode},
$$
\frac{\partial \log{\left( \ghmdistbar(x_t\,|\,s,a; \theta)\right)}}{\partial x_t} = -\Tr\left({\frac{\partial\, \ghmvecfield_t(x_t \,|\, s, a; \theta)}{\partial x_t}}\right) \, .
$$

We can now compute the log-likelihood for a sample $X \sim \ghmdist^{\pi}(\cdot\,|\, s, a)$ by integrating the total change in log-density backward in time from $x_1 = X$ to obtain $x_0$ which has tractable likelihood.
In practice, we solve the following coupled initial value problem using numerical integration \citep{grathwohl19ffjord},
\begin{equation}
\begin{gathered}\label{eq:llivp}
\begin{bmatrix}
x_0 \\
\log{\ghmdistbar(x_1\,|\,s,a; \theta)} - \log{\varphi(x_0)}
\end{bmatrix} =
\int_{1}^0 \begin{bmatrix}
    -\ghmvecfield_t(x_t \,|\, s, a; \theta) \\
    \Tr\left({\frac{\partial\, \ghmvecfield_t(x_t \,|\, s, a; \theta)}{\partial x_t}}\right)
\end{bmatrix} dt \,, \\
\,\,\text{where}\,\,
\begin{bmatrix}
    x_1 \\
    \log{\ghmdistbar(x\,|\,s,a; \theta)} - \log{\ghmdistbar(x_1\,|\,s,a; \theta)}
\end{bmatrix} =
\begin{bmatrix}
    X \\
    0
\end{bmatrix} \, .
\end{gathered}
\end{equation}
For all experiments we report the negative log-likelihood \emph{normalized by the dimension of the observation space}.

\textbf{Earth Mover's Distance} We compute the Earth Mover's Distance \citep[EMD;][]{RubnerTG00}, also known as the Wasserstein-1 distance, between $m=2048$ samples from the ground truth distribution $X \sim m^\pi(\cdot\,|\,S_k, A_k)$ and our learned GHM $\widetilde{X} \sim \ghmdistbar(\cdot\,|\,S_k, A_k; \theta)$ for a set of randomly sampled state-action pairs $\{ (S_k, A_k) \}_{k=1}^n$.
Intuitively, the EMD quantifies the minimum cost required to transform one distribution into another, where the cost is defined in terms of the Euclidean distance between states $X^{(i)}, X^{(j)}$. Formally, we have,
\begin{align*}
    \text{EMD}(\{ X^{(1)}, \dots, X^{(m)} \}, \{ \widetilde{X}^{(1)}, \dots, \widetilde{X}^{(m)} \}) = \min_{\xi \in \Xi} \sum_{i,j} \xi_{ij} \sum_{k=1}^d  \left( X^{(i)}_k - \widetilde{X}^{(j)}_k \right)^2 \,,
\end{align*}
where $\xi$ is a transport plan such that $\xi_{ij}$ specifies the proportion of mass moved from $X_i$ to $\widetilde{X}_j$.
We report the average EMD across $n=64$ source states using the Python Optimal Transport \citep{flamary2021pot} library.

\textbf{Value Function Mean Square Error (MSE(V)).} We compute the mean square error between a Monte-Carlo estimation $\widetilde{V}^\pi_{\mathrm{\textsc{mc}}}$ of the value function $V^{\pi}(s)$ and the estimation $\widetilde{V}_{\mathrm{\textsc{ghm}}}$ obtained using the learned model. We obtain $\widetilde{V}^\pi_{\mathrm{\textsc{mc}}}$ by collecting a trajectory $\{ (s_0, s_1, \dots, s_T) \}$ from policy $\pi$ and computing the discounted sum of rewards. We generate a single trajectory since both the policy and the environment are deterministic. The GHM estimate is given by \eqref{eq:sm-vf}, i.e., $$\widetilde{V}^\pi_{\mathrm{\textsc{ghm}}}(s) = (1 - \gamma)^{-1} \mathbb{E}_{\widetilde{X} \sim \ghmdistbar(\cdot\mid s, \pi(s))}\Big[\,r(\widetilde{X})\Big]\,.$$ Then, $\mathrm{MSE}(\widetilde{V}^\pi_{\mathrm{\textsc{mc}}},\widetilde{V}^\pi_{\mathrm{\textsc{ghm}}}) = \E_{S_0 \sim \nu}{(\widetilde{V}^\pi_{\mathrm{\textsc{ghm}}}(S_0)- \widetilde{V}^\pi_{\mathrm{\textsc{mc}}}(S_0))^2}$. We average our results over $64$ initial states $S_0$ sampled from the initial state distribution $\nu$.

\textbf{Planning with GPI.}
We evaluate planning performance by computing the average return over $100$ episodes, each lasting $1,000$ steps, for every task.
For the Forward-Backward representation \citep{touati21fwdbwd}, we directly follow the policy $\pi_{w_r}$ (thus $a_t = \pi_{w_r}(s_t)$) where $w_r = \E_{(S, R) \sim \rho}{B(s) \cdot R}$ is the zero-shot policy embedding inferred using $250,000$ transitions labeled with the task reward function $r$. Given that FB provides a direct way of estimating the value function of a policy (i.e., $Q^{\pi_w}_r(s,a) = F(s,a,w)^T z_r$), we can do planning in the policy embedding space by solving the following problem:
\[
w_t^{\mathrm{\textsc{fb-gpi}}}\in \argmax_{w \sim D(\setfont{W})} F(s_t, \pi_w(s_t),w)^T w_r.
\]
This optimization problem requires no generation except sampling from $D(\setfont{W})$. We approximate the max using $255$ samples from $D(\setfont{W})$ and additionally incorporating $w_r$ to ultimately maximize over $256$ policies.
On the other hand, for \textsc{ghm-gpi}, we solve the following optimization problem,
\[
w^{\mathrm{\textsc{ghm-gpi}}}_t \in \argmax_{w\sim D(\setfont{W})}\,\, \underbrace{(1 - \gamma)^{-1} \E_{X \sim \sm^{\pi_w}(\cdot|s_t,\pi_w(s_t)))}{r(X)}}_{Q^{\pi_w}(s_t, \pi_w(s_t))},
\]
which requires generating samples from $\sm^{\pi_w}$. In our experiments we generate $128$ samples from $\sm^{\pi_w}$.

\subsection{Environments}

\begin{figure}[h]
    \centering
    \hfill
    \includegraphics[width=0.21\textwidth]{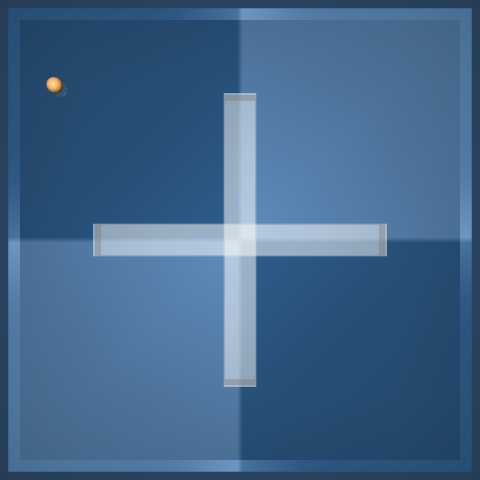}\hfill
    \includegraphics[width=0.21\textwidth]{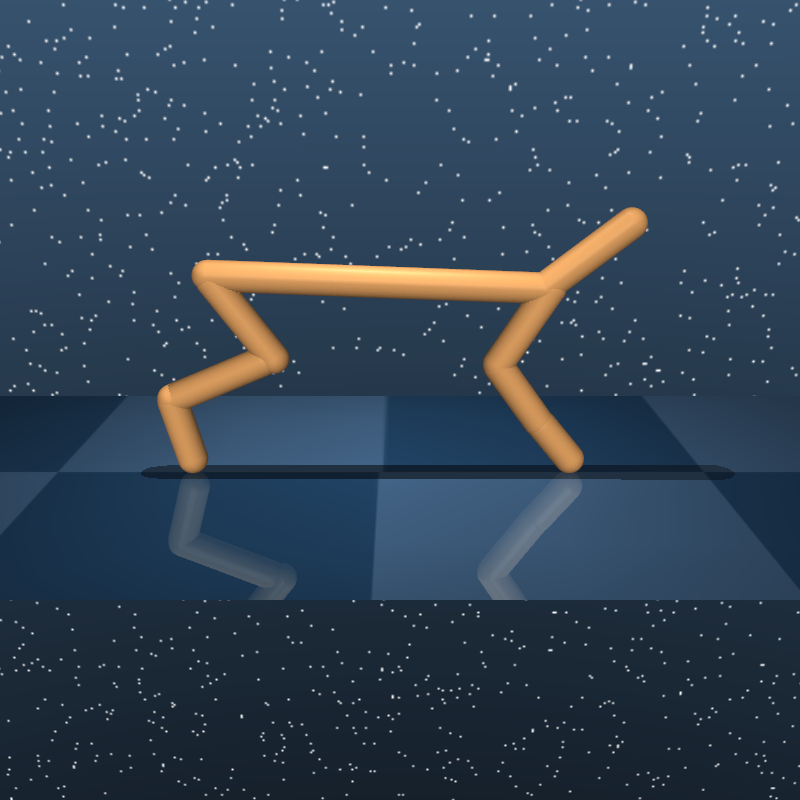}\hfill
    \includegraphics[width=0.21\textwidth,clip]{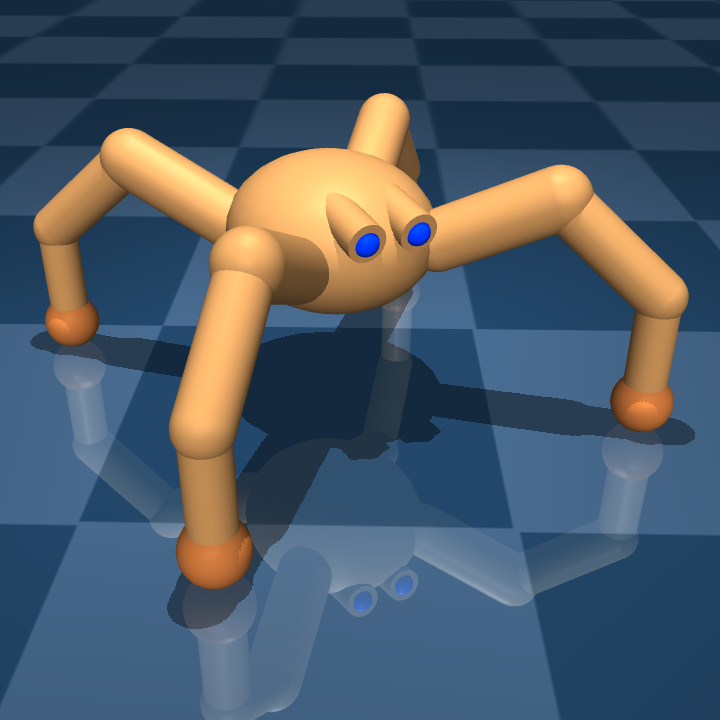}\hfill
    \includegraphics[width=0.21\textwidth]{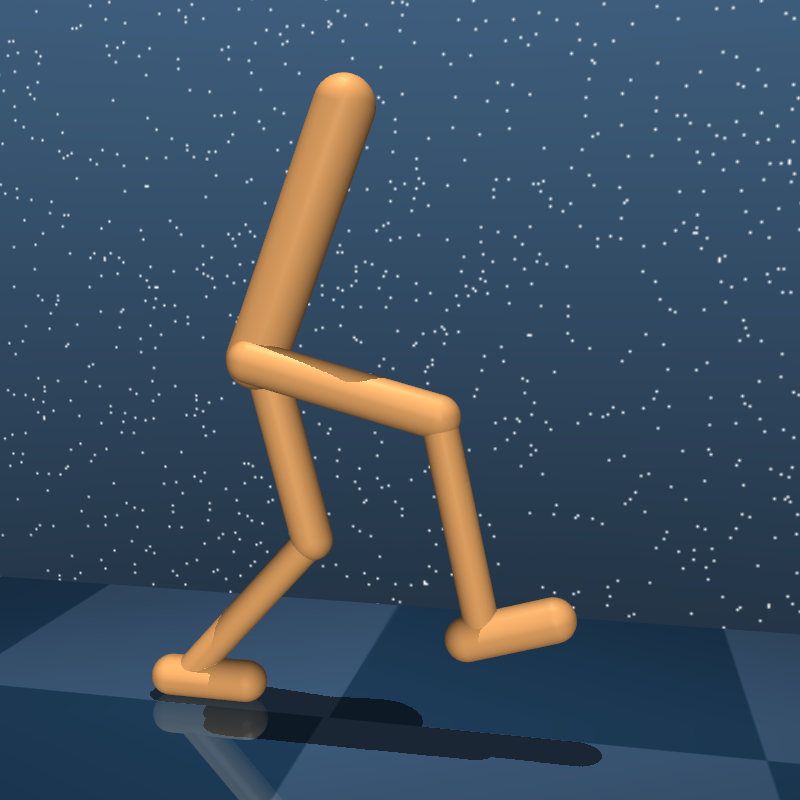}
    \hfill
    \caption{A visual depiction of each domain used in our experiments from the DeepMind Control Suite \citep{tunyasuvunakool20dmc}. From left to right: \textsc{maze}, \textsc{cheetah}, \textsc{quadruped}, \textsc{walker}.}\label{fig:dmc-envs}
\end{figure}

Experiments in this paper were conducted with a subset of domains from the DeepMind Control Suite \citep{tunyasuvunakool20dmc} highlighted in \autoref{fig:dmc-envs}.

\clearpage
\subsection{Geometric Horizon  Models}

This section describes each class of generative model used for our empirical experiments.

\subsubsection{Flow Matching}\label{app:algorithms}

\begin{figure}[H]
    \vspace{-5mm}
    \begin{minipage}[t]{0.575\textwidth}
        \begin{algorithm}[H]
            \caption{Template for TD-Flow algorithms}\label{alg:td-flow}
        \begin{small}
        	\begin{algorithmic}[1]
        \State \textbf{Inputs}: offline dataset $\mathcal{D}$, policy $\pi$, batch size $n$, Polyak coefficient $\zeta$, weight decay $\lambda$, randomly initialized weights $\theta$, discount factor $\gamma$, learning rate $\eta$, one-step conditional path $\onestepterm{p}_{t\mid 1}$ and conditional vector-field $\onestepterm{u}_{t\mid 1}$, bootstrap path $\bootterm{p}_t$ and vector-field $\bootterm{v}_t$.
            \For{$n = 1, \dots$}
            \State Sample mini-batch $\{ (S_k, A_k, S_k')\}_{k=1}^K$ from $\mathcal{D}$
            \For{$k = 1, \dots, K$}
            \State Sample $t_k \sim \mathcal{U}([0,1])$
            \State Sample $\onestepterm{X}_{k} \sim \onestepterm{p}_{t_k\mid 1}(\,\cdot\mid S_{k}')$
            \State $\onestepterm{\ell}_k(\theta) = \big\| v_{t_k}(\onestepterm{X}_{k}\mid S_k, A_k; \theta) - \onestepterm{u}_{t_k\mid 1}(\onestepterm{X}_{k} \mid S_k') \big\|^2$
            \State Sample $\bootterm{X}_{k} \sim \bootterm{p}_{t_k}(\cdot\mid S_{k}', \pi(S_k'); \bar\theta)$
            \State $\bootterm{\ell}_k(\theta) =  
            \big\| v_{t_k}(\bootterm{X}_{k}\mid S_k, A_k; \theta) - \bootterm{v}_{t_k}(\bootterm{X}_{k}\mid S'_k, \pi(S'_k); \bar\theta) \big\|^2$
            \EndFor
            \State {\bf \textcolor{gray!50!blue}{\# \texttt{Compute loss}}}
            \State $\ell(\theta) = \frac{1}{K} \sum_{k=1}^K (1-\gamma) \onestepterm{\ell}_k(\theta) + \gamma \bootterm{\ell}_k(\theta)$
            
            \State {\bf  \textcolor{gray!50!blue}{\# \texttt{Perform gradient step}}}
            \State $\theta \leftarrow \theta - \eta \nabla_{\theta}\left( \ell(\theta) + \lambda \| \theta \|^2 \right)$
            \State {\bf  \textcolor{gray!50!blue}{\# \texttt{Update parameters of target vector field}}}
        	\State  $\bar\theta \leftarrow \zeta \bar\theta + (1-\zeta) \theta$ 
            \EndFor
        \end{algorithmic}
         \end{small}
        \end{algorithm}
    \end{minipage}
    \hfill
    \begin{minipage}[t]{0.4\textwidth}
        \vspace{4mm}
        \captionof{table}{Summary of how different TD-flow algorithms generate the target probability path and vector field. The neural ode $\psi_t$ is defined by the vector field $\overset{\smash{\curvearrowright}}{v}_t$ computed at iteration $n$.}\label{table:algorithms}
        \vspace{3mm}
        \renewcommand{\arraystretch}{2}
        \resizebox{\linewidth}{!}{
        \begin{tabular}{c c c}
             \toprule
             &  $\mathbf{\overset{\smash{\curvearrowright}}{p}_t}$ & $\mathbf{\overset{\smash{\curvearrowright}}{v}_t}$ \\
             \hline
             \multirow{3}{*}{\rotatebox{90}{\vtdfshort}} & $X_0 \sim m_0$ & \multirow{3}{*}{$u_{t\mid 1}(X_t \mid X_1)$} \\
             & $X_1 = \psi_1(X_0\mid S', A'; \bar\theta)$ & \\
             & $X_t \sim p_{t\mid 1}(\cdot\mid X_1)$ \\
             \hline
             \multirow{3}{*}{\rotatebox{90}{\ctdfshort}} & $X_0 \sim m_0$ & \multirow{3}{*}{$u_{t\mid 0,1}(X_t\mid X_0, X_1)$} \\
             & $X_1 = \psi_1(X_0\mid S', A';\bar\theta)$ \\
             & $X_t \sim p_{t\mid 0, 1}(\cdot\mid X_0, X_1)$ \\
             \hline
             \multirow{2}{*}{\rotatebox{90}{\!\!\tdfshort}} & $X_0 \sim m_0$ & \multirow{2}{*}{$v_t(X_t\mid S', A';\bar\theta)$} \\
             & $X_t = \psi_t(X_0\mid S', A';\bar\theta)$ \\
             \bottomrule
        \end{tabular}}
    \end{minipage}
\end{figure}

To discuss the TD-Flow methods introduced herein, we first unify the loss function through defining a general template for the loss as,
\begin{align*}\label{eq:unified.loss}
    \ell(\theta) &= (1-\gamma) 
    \mathbb{E}_{\rho,t,X_t \sim \onestepterm{p}_{t\mid 1}(\cdot\mid S')}
    \Big[
    \big\| v_t(X_t\mid S, A;\theta) - \onestepterm{u}_{t\mid 1}(X_t\mid S')
    \big{\|}^2
    \Big] \\
     \qquad& + \gamma 
    \mathbb{E}_{\rho,t,X_t \sim \bootterm{p}_t^{(n)}(\cdot\mid Z)}
    \Big[
    \big\| v_t(X_t\mid S, A;\theta) - \bootterm{v}_t^{(n)}(X_t\mid Z)
    \big{\|}^2
    \Big] \, .
\end{align*}
We can now recover each algorithm by a specific choice of the target probability path $\bootterm{p}_t^{(n)}$ and vector field $\bootterm{v}_t^{(n)}$ as illustrated in \autoref{table:algorithms}. Based on this unified structure, we present pseudo-code for the TD flow methods in \autoref{alg:td-flow}. In practice, instead of proceeding through full iterations, we use standard mini-batch gradient updates with a target network $\bar\theta$ updated as a moving average of $\theta$.

When employing the conditional probability path $\onestepterm{p}_{t\mid 1}$ and vector field $\onestepterm{u}_{t\mid 1}$ we use the standard Gaussian linear interpolation defined as $\onestepterm{p}_{t\mid 1}(\cdot \mid X_1) = \mathcal{N}(\cdot \mid tX_1, (1 - t)^2 I)$, hence $X_t = t X_1 + (1 - t) X_0 \sim p_{t\mid 1}$, resulting in $\onestepterm{u}_{t\mid 1}(X_t\mid X_1) = (X_1 - X_t) / (1 - t)$ \citep{lipman2022flow}. The source distribution for all experiments is $m_0(\cdot) = \mathcal{N}(\cdot\mid 0, I)$. To sample from the Neural ODE we use the Midpoint method with a constant step size of $\mathrm{d}t = t / 10$ for a total of $10$ steps. We found both coupled and TD${}^2$ methods do not require many solver steps and hypothesize this is due to the reduction in transport cost as analyzed in \autoref{app:sec:transport}.

For all flow and diffusion-based methods, we employ a U-Net-style architecture \citep{ronneberger15unet} that has hierarchical skip connections throughout an MLP. We embed the timestep $t$ by first increasing its dimensionality with a sinusoidal embedding before transforming it through a two-layer MLP with mish activations \citep{misra19mish}.
We further process additional conditioning information, such as the state-action pair and Forward-Backward embedding $z$ through an additional two-layer MLP, whose result then gets concatenated with our time embedding. Finally, the network integrates all prior conditioning information through FiLM modulation \citep{perez18film} that replaces the learned affine transformation for layer normalization \citep{ba2016ln}.

\subsubsection{Denoising Diffusion}

We train a Denoising Diffusion Probabilistic Model \citep[DDPM;][]{ho20ddpm} using the same architecture as our flow matching model above, with the output now being interpreted as a prediction of the noise seed $\epsilon_0$ that began the diffusion process.
We discretize the diffusion process using $1,000$ steps with $\beta_{\mathrm{min}} = 0.1$ and $\beta_{\mathrm{max}} = 20$.
We employ the DDIM sampler\citep{song21ddim} with $50$ sampling steps for both training~and~evaluation.

For evaluating our DDPM model, we compute exact log-likelihoods using the instantaneous change of variables formula \citep{chen18neuralode} along with the probability flow ODE from \citet{song21sde}. That is, we solve the initial value problem in \eqref{eq:llivp} using the vector field,
$$
v_t(x_t \,|\, s, a; \theta) = -\frac{1}{2} \left( \beta_{\mathrm{min}} + t \, (\beta_{\mathrm{max}} - \beta_{\mathrm{min}}) \right) \left( x_t - \frac{1}{\sqrt{1 - \bar{\alpha}_t}} \epsilon_t(x_t \,|\, s, a; \theta) \right) \, .
$$

We now outline the losses for each of the \textsc{td-dpm} experiments in the paper:

\textbf{TD-DD}
To train our vanilla Diffusion GHM we employ the standard DDPM-style objective, that is, we optimize the following loss:
\begin{equation}
    \E_{\substack{\rho,\,t,\,\epsilon \sim \mathcal{N}(\cdot\,|\, 0, I) \\ X_0 \sim \Tpi{\ghmdistbar^{(n)}}(\cdot\mid S, A)}}{
        \norm{\epsilon - \epsilon_t( \sqrt{\bar{\alpha}_t} X_0 + \sqrt{1 - \bar{\alpha}_t} \epsilon \mid S, A; \theta)}^2
    } \,,
\end{equation}
where $\bar{\theta}$ are the target parameters and $\bar{\alpha}$ are the standard diffusion coefficients as seen in \citet{ho20ddpm}.

\textbf{TD\textsuperscript{2}-DD}
As outlined in \autoref{ssec:diffusion} we can split our DDPM loss into two terms, one that will use standard DDPM training on one-step transitions and the second term that will regress to our target networks noise prediction. This materializes as,
\begin{gather}
\onestepterm{\ell}(\theta)
= \E_{\rho,t,\epsilon,X_0}{
\big{||}
 \epsilon_t(\sqrt{\bar{\alpha}_t} X_0 + \sqrt{1 - \bar{\alpha}}_t \epsilon \,|\, S, A; \theta) - \epsilon
\big{||}^2
} 
\tag*{} \\
\text{where}\, X_0 \sim P(\cdot\,|\,S, A) \,, \tag*{} \\[0.5\baselineskip]
\bootterm{\ell}(\theta)
=
\mathbb{E}_{\rho,t,\epsilon,\bootterm{X}_t}
\bigg[ 
\big{||}
 \epsilon_t(\bootterm{X}_t \,|\, S, A; \theta) -
 \epsilon^{(n)}_t(\bootterm{X}_t \,|\, S', \pi(S'))
\big{||}^2
\bigg] \tag*{} \\
\text{where}\, \bootterm{X}_t \sim q^{(n)}_{t|T}(\cdot\,|\,S', \pi(S')) \tag*{} \\[0.5\baselineskip]
\ell_{\tddshort}(\theta) = (1 - \gamma)
\onestepterm{\ell}(\theta)
+ \gamma
\bootterm{\ell}(\theta)
\end{gather}

\subsubsection{Generative Adversarial Network}

We implement a modern Generative Adversarial Network \citep[GAN;][]{goodfellow14gan} baseline based on the recommendations in \citet{huang24gan}. Specifically, we train a relativistic GAN \citep{jolicoeur-martineau19rgan} resulting in the following loss,
\begin{gather*}
    \ell_{\text{\small\textsc{gan}}}(\theta_{\mathrm{G}}, \theta_{\mathrm{D}}) = \E_{\rho, X_0, X_1}{
      f\left(D(G(X_0\mid S, A; \theta_{\mathrm{G}}); \theta_{\mathrm{D}}) - D(X_1\mid S, A; \theta_{\mathrm{D}})  \right)
    } \,, \\
    \text{where}\,\, X_0 \sim \mathcal{N}(\cdot\mid 0, I)\,,
    X_1 \sim \Tpi{\ghmdistbar^{(n)}}(\cdot\mid S, A) \,,
\end{gather*}
We take $f(x) = -\log{(1 + \exp{(-x)})}$ to be the log-sigmoid function \citep{jolicoeur-martineau19rgan} and further add the following zero-centered gradient penalties on the discriminator,
\begin{align*}
R_1(\theta_{\mathrm{D}}) &= \E_{\rho, X \sim \Tpi{\ghmdistbar^{(n)}}(\cdot\mid S, A)}{\norm{\nabla_{X} D(X \mid S, A) }^2} \,, \\
R_2(\theta_{\mathrm{G}}, \theta_{\mathrm{D}}) &= \E_{\rho, X \sim \Tpi{\ghmdistbar}(\cdot\mid S, A; \theta_{\mathrm{G}})}{\norm{\nabla_X D(X\mid S, A)}^2} \, .
\end{align*}
The penalty $R_1$ penalizes the gradient norm of the discriminator $D$ on ``real data'' sampled from our current iterate $\ghmdistbar^{(n)}$, whereas $R_2$ penalizes the gradient norm on ``fake data'' generated directly from the current generator. We experimented with different coefficients and schedules for these gradient penalties and settled on a linear decay schedule from $0.05 \to 0.005$ throughout training. Furthermore, as is common practice, we impose a schedule on the second moment EMA coefficient $\beta_2$ of Adam \citep{kingma15adam} to increase from $0.9 \to 0.99$ throughout training.

The generator and discriminator architecture in our GAN is implemented as a Residual MLP with leaky ReLU activations with the same FiLM-style conditioning \citep{perez18film} as our flow and diffusion models. The input to our generator is random noise sampled from an isotropic Gaussian with dimensionality equal to the number of state dimensions in the environment.

\subsubsection{Variational Auto-Encoder}

We implement a $\beta$-Variational Auto-Encoder \citep{kingma13autoencoding,higgins17beta} following the best practices outlined in \citet{thakoor22ghm}. That is, we train our VAE to minimize the following loss,
\begin{gather*}
\ell_{\text{\small\textsc{vae}}}(\theta_{\mathrm{E}}, \theta_{\mathrm{D}}) = \E_{\rho,X_1}{\E_{X_0 \sim q_{\theta_{\mathrm{E}}}(\cdot\mid S, A, X_1)}{\log{p_{\theta_{\mathrm{D}}}(X_1\mid S, A, X_0)}} - \beta D_{\mathrm{KL}}( q_{\theta_{\mathrm{E}}} \| p_0 )} \,, \\
\text{where}\,\, X_1 \sim \Tpi{\ghmdistbar^{(n)}}(\cdot\mid S, A) \,.
\end{gather*}

We employ a similar architecture to our GAN-GHM and use a residual MLP for the encoder and decoder. We use an isotropic Gaussian latent space with the number of latents equal to the number of state dimensions in the environment. We also swept over $\beta \in \{ 0.1, 0.2, 0.3, 0.4, 0.5, 0.6, 0.7, 0.8, 0.9, 1.0 \}$ on the \textsc{maze} domain and chose $\beta=0.5$ for the rest of our experiments. Overall, we found the $\beta$-VAE-based GHM to be very unstable and likely requires very careful fine-tuning of $\beta$ to get adequate performance at long-horizons.

\clearpage
\subsection{Hyperparameters}

We report the hyper-parameters for training the GHM models used in the single and multi-policy experiments. \autoref{table:fm-dd-hparams} shows the parameters for Flow Matching and Denoising Diffusion. We also report the hyper-parameters for pre-training the Forward-Backward representation \citep{touati21fwdbwd} utilized in the multi-policy GHM experiments in \autoref{table:fb-hparams}.

\begin{table}[H]
\caption{Flow Matching and Denoising Diffusion hyper-parameters used for the single and multi-policy experiments across tasks and domains. We \mhl{highlight} any differences depending on the training context.}
\label{table:fm-dd-hparams}
\renewcommand{\arraystretch}{1.25}
\centering
\resizebox{
\ifdim\width>\columnwidth
    0.975\columnwidth
  \else
    \width
  \fi
}{!}{
\begin{tabular}{@{}clll@{}}
     \toprule
     & {\fontsize{11}{13}\selectfont \textbf{Hyperparameter}}
     & {\fontsize{11}{13}\selectfont \textbf{Single Policy}}
     & {\fontsize{11}{13}\selectfont \textbf{Multi-Policy}} \\
     \midrule
     \multirow{3}{*}{\makecell[c]{Flow Matching\\\citep{lipman2022flow}}} 
     & ODE Solver & Midpoint & Midpoint\\
     & ODE $\mathrm{d}t$ (train) & $0.1$ & $0.1$\\
     & ODE $\mathrm{d}t$ (eval) & $0.1$ & $0.05$ (\mhl{$0.1$ for GPI})\\
     \midrule
    \multirow{6}{*}{
    \makecell[c]{Diffusion (DDPM)\\ \citep{ho20ddpm}}} & $\beta_{\text{min}}$ & $0.1$ & $0.1$ \\
    & $\beta_{\text{max}}$ & $20$ & $20$ \\
    & Discretization Steps & $1,000$ & $1,000$ \\
    & SDE Solver & DDIM \citep{song21ddim} & DDIM \citep{song21ddim} \\
    & SDE Solver Steps (train) & 20 & 20 \\
    & SDE Solver Steps (eval) & 20 & 20 \\
     \midrule
\multirow{5}{*}{\makecell[c]{Network (U-Net)\\\citep{ronneberger15unet}}} & $t$-Positional Embedding Dim. & $256$ & $256$ \\
        & $t$-Positional Embedding MLP & $(256, 256)$ & $(256, 256)$ \\
        &Hidden Activation& mish \citep{misra19mish} & mish \citep{misra19mish} \\
        & Blocks per Stage & $1$ & $1$ \\
        & Block Dimensions & $(512, 512, 512)$ & $(1024, 1024, 1024)$ \\
        \midrule
\multirow{3}{*}{\makecell[c]{Conditional Encoder}} & Encoder Input & $s,a$ & $s,a,z$ \\
& Encoder MLP & $(512, 512, 512)$ & $(1024, 1024, 1024)$ \\
& Encoder Activation& mish \citep{misra19mish} & mish \citep{misra19mish} \\
\midrule
\multirow{5}{*}{\makecell[c]{Optimizer (AdamW)\\\citep{loshchilov19decoupled}}} 
& AdamW $\beta_1$ & $0.9$ & $0.9$ \\
& AdamW $\beta_2$ & $0.999$ & $0.999$ \\
& AdamW $\epsilon$ & $10^{-4}$ & $10^{-4}$ \\
& Learning Rate & $10^{-4}$  & $10^{-4}$ \\
& Weight Decay & $10^{-3}$ & $10^{-2}$ \\
\midrule
\multirow{3}{*}{\makecell[c]{Common}}
& Gradient Steps & $3$M & $8$M \\
& Batch Size & $1024$ & $1024$\\
& Target Network EMA & $10^{-3}$ & $10^{-4}$ \\
\bottomrule
\end{tabular}
}
\end{table}

\begin{figure}[H]
    \centering
    \begin{tabular}{@{}p{0.475\linewidth} p{0.475\linewidth}@{}}
        \begin{minipage}[t]{\linewidth}
        \captionof{table}{$\beta$-VAE \citep{higgins17beta} hyper-parameters for single policy experiments across tasks and domains.}
        \label{table:fm-vae-hparams}
        \resizebox{\linewidth}{!}{
        \renewcommand{\arraystretch}{1.2}
            \begin{tabular}{@{}cll@{}}
     \toprule
     & {\fontsize{11}{13}\selectfont \textbf{Hyperparameter}}
     & {\fontsize{11}{13}\selectfont \textbf{Value}} \\
     \midrule
    \multirow{3}{*}{
    \makecell[c]{$\beta$-VAE\\ \citep{higgins17beta}}} & $\beta$ & 10 \\
    & Latent Prior & $\mathcal{N}(0, I)$ \\
    & Latent Dimension &  $|\setfont{S}|$\\
     \midrule
\multirow{5}{*}{\makecell[c]{Network}} & Encoder  &  Residual MLP \\
        & Decoder &  Residual MLP \\
        & Hidden Activation & mish \citep{misra19mish} \\
        & Blocks per Stage & $1$ \\
        & Block Dimensions & $(512, 512, 512)$ \\
        \midrule
\multirow{3}{*}{\makecell[c]{Conditional Encoder}} & Encoder Input & $s,a$ \\
& Encoder MLP & $(512, 512, 512)$ \\
& Encoder Activation& mish \citep{misra19mish} \\
\midrule
\multirow{5}{*}{\makecell[c]{Optimizer (AdamW)\\\citep{loshchilov19decoupled}}} 
& AdamW $\beta_1$ & $0.9$ \\
& AdamW $\beta_2$ & $0.999$ \\
& AdamW $\epsilon$ & $10^{-4}$ \\
& Learning Rate & $10^{-4}$ \\
& Weight Decay & $10^{-3}$ \\
\midrule
\multirow{3}{*}{\makecell[c]{Common}}
& Gradient Steps & $3$M \\
& Batch Size & $1024$ \\
& Target Network EMA & $10^{-3}$ \\
\bottomrule
\end{tabular}

        }
        \end{minipage}
        &
        \begin{minipage}[t]{\linewidth}
        \captionof{table}{GAN hyper-parameters for single policy experiments across tasks and domains.}
        \label{table:fm-gan-hparams}
        \resizebox{\linewidth}{!}{
        \renewcommand{\arraystretch}{1.2}
            \begin{tabular}{@{}cll@{}}
     \toprule
     & {\fontsize{11}{13}\selectfont \textbf{Hyperparameter}}
     & {\fontsize{11}{13}\selectfont \textbf{Value}} \\
     \midrule
    \multirow{3}{*}{
    \makecell[c]{RGAN\\ \citep{jolicoeur-martineau19rgan}}} & Grad. Penalty Coef  & Linear($0.05 \to 0.005$)  \\
    & Latent Prior & $\mathcal{N}(0, I)$ \\
    & Latent Dimension &  $|\setfont{S}|$\\
     \midrule
\multirow{5}{*}{\makecell[c]{Network}} & Generator &  Residual MLP \\
        & Discriminator &  Residual MLP \\
        & Hidden Activation & Leaky ReLU \\
        & Blocks per Stage & $1$ \\
        & Block Dimensions & $(512, 512, 512)$ \\
        \midrule
\multirow{3}{*}{\makecell[c]{Conditional Encoder}} & Encoder Input & $s,a$ \\
& Encoder MLP & $(512, 512, 512)$ \\
& Encoder Activation& Leaky ReLU \\
\midrule
\multirow{5}{*}{\makecell[c]{Optimizer (AdamW)\\\citep{loshchilov19decoupled}}} 
& AdamW $\beta_1$ & $0.9$ \\
& AdamW $\beta_2$ & Linear($0.9 \to 0.99$) \\
& AdamW $\epsilon$ & $10^{-4}$ \\
& Learning Rate & $10^{-4}$ \\
& Weight Decay & $10^{-3}$ \\
\midrule
\multirow{3}{*}{\makecell[c]{Common}}
& Gradient Steps & $3$M \\
& Batch Size & $1024$ \\
& Target Network EMA & $10^{-3}$ \\
\bottomrule
\end{tabular}

        }
        \end{minipage}
    \end{tabular}
\end{figure}

\begin{table}[H]
\renewcommand{\arraystretch}{1.25}
\centering
\caption{Forward Backward Representation hyper-parameters. We largely reuse the hyper-parameters from \citet{pirotta24bfmil} and \mhl{highlight} any deviations.}\label{table:fb-hparams}
\resizebox{
\ifdim\width>\columnwidth
    0.975\columnwidth
  \else
    \width
  \fi
}{!}{
\begin{tabular}{@{}clllll@{}}
\toprule
& {\fontsize{11}{13}\selectfont \textbf{Hyperparameter}}
& {\fontsize{11}{13}\selectfont \textbf{Walker}}
& {\fontsize{11}{13}\selectfont \textbf{Cheetah}}
& {\fontsize{11}{13}\selectfont \textbf{Quadruped}}
& {\fontsize{11}{13}\selectfont \textbf{Maze}} \\ \midrule
\multirow{5}{*}{\makecell[c]{Forward Backward\\ \citep{touati21fwdbwd}}} &
Embedding Dimension $d$ & $100$ & $50$ & $50$ & $100$ \\
& Embedding Prior & $S^d$ & $S^d$ & $S^d$ & $S^d$ \\
& Embedding Prior Goal Prob.& $0$ & $0$ & $0$ & $\nicefrac{1}{2}$ \\
& $B$ Normalization & $\ell_2$ & \mhl{\,$\ell_2$\,} & $\ell_2$ & \mhl{\,$\ell_2$\,} \\
& \makecell[l]{Orthonormal Loss Coeff.} & $1$ & $1$ & $1$ & $1$ \\ \midrule
\multirow{3}{*}{\makecell[c]{Policy (TD3)\\ \citep{fujimoto18td3}}} & Target Policy Noise & $\mathcal{N}(0, 0.2)$ & $\mathcal{N}(0, 0.2)$ & $\mathcal{N}(0, 0.2)$ & $\mathcal{N}(0, 0.2)$ \\
& Target Policy Clipping & $0.3$ & $0.3$ & $0.3$ & $0.3$ \\
& Policy Update Frequency & $1$ & $1$ & $1$ & $1$ \\ \midrule
\multirow{5}{*}{\makecell[c]{Optimizer (Adam)\\ \citep{kingma15adam}}} 
& Learning Rate (F, B)  & ($10^{-4}$, $10^{-4}$) & ($10^{-4}$, $10^{-4}$) & ($10^{-4}$, $10^{-4}$) & ($10^{-4}$, $10^{-6}$) \\
& Learning Rate ($\pi$)  & $10^{-4}$ & $10^{-4}$ & $10^{-4}$ & $10^{-6}$ \\
& $\mathrm{Adam}$ $\beta_1$ & $0.9$ & $0.9$ & $0.9$ & $0.9$ \\
& $\mathrm{Adam}$ $\beta_2$ & $0.999$ & $0.999$ & $0.999$ & $0.999$ \\
& $\mathrm{Adam}$ $\epsilon$ & $10^{-8}$ & $10^{-8}$ & $10^{-8}$ & $10^{-8}$ \\ \midrule
\multirow{5}{*}{\makecell[c]{\hfill Common \hfill}} & Batch Size     & $2048$ & $1024$ & $2048$ & $1024$ \\
& Gradient Steps & $3$M & $3$M & $3$M & $5$M \\
& Discount Factor $\gamma$ & $0.98$ & $0.98$ & $0.98$ & $0.99$ \\
& Target Network EMA & $0.99$ & $0.99$ & $0.99$ & $0.99$ \\
& \makecell[l]{Reward Inference Samples} & \mhl{$250,000$}& \mhl{$250,000$}& \mhl{$250,000$}& \mhl{$250,000$}\\
\bottomrule
\end{tabular}}
\end{table}

\clearpage

\section{Additional Experimental Results}
\label{app:additional.results}

In this section, we report additional results about the experiments. 

\textbf{Single policy.} 
We report metrics averaged over tasks using a curved conditional path in \autoref{app:results.curved}. We also report the performance per task in \autoref{app:results.curved.task}.
\autoref{tab:singlepolicy} shows the performance of the single-policy experiments (\autoref{sec:experiments.singlepol} in the main paper) expanded for each task. While the performance of TD-based methods is reasonably stable across tasks, VAE and GAN have a large variance across tasks. For example, the EMD of GAN diverges in 2 tasks out of 4 in \textsc{Quadruped}.

\textbf{Multiple policies and planning.}
We report aggregate performance across our full suite of evaluation metrics for the multi-policy experiments in \autoref{tab:fb_policy_value}.
We also report per-task metrics in \autoref{tab:multipolicy}.
We can notice that \tddshort{} achieves quite a high EMD compared to \vtddshort{} while achieving a better MSE(V). By further inspecting the generated samples (see \autoref{fig:pointmass_media}), we found that \vtddshort{} tends to generate highly concentrated samples, while \tddshort{} is more diffuse. However, the samples generated by \vtddshort{} appear to be better at a visual inspection. This may explain the discrepancy between the two metrics.
Finally, we report aggregate planning performance in \autoref{tab:fb_policy_planning} and per-task results in \autoref{tab:gpi.per_task}.

\begin{table}[H]
    \centering
    \renewcommand{\arraystretch}{2}
    \caption{\textbf{Per task} results for the \textbf{single policy} experiments.}
    \label{tab:singlepolicy}
    \vspace{-5mm}

\end{table}

\begin{table}[H]
    \centering
    \caption{Results averaged over tasks for the \textbf{single policy} experiments with a \textbf{curved} conditional path.}
    \label{app:results.curved}
    \resizebox{.5\columnwidth}{!}{
        \renewcommand{\arraystretch}{1.2}
        \begin{tabular}{c c c c c}
\toprule
{\fontsize{11}{13}\selectfont \textbf{Domain}}
& {\fontsize{11}{13}\selectfont \textbf{Method}}
& {\fontsize{11}{13}\selectfont \textbf{EMD}
\raisebox{0.3ex}{\scalebox{0.7}{$\downarrow$}}
}
& {\fontsize{11}{13}\selectfont \textbf{NLL}
\raisebox{0.3ex}{\scalebox{0.7}{$\downarrow$}}
}
& {\fontsize{11}{13}\selectfont \textbf{MSE(V)}
\raisebox{0.3ex}{\scalebox{0.7}{$\downarrow$}}
} \\
\hline
\multirow{3}{*}{
\textsc{Cheetah}
} & \vtdfshort & $13.91$ (\small\textcolor{gray}{$0.73$}) & $1.354$ (\small\textcolor{gray}{$0.017$}) & $477.89$ (\small\textcolor{gray}{$40.53$}) \\
 & \ctdfshort & $25.86$ (\small\textcolor{gray}{$18.91$}) & $1.295$ (\small\textcolor{gray}{$0.067$}) & $189.21$ (\small\textcolor{gray}{$17.69$}) \\
 & \tdfshort & \cellcolor{best} $10.79$ (\small\textcolor{gray}{$0.03$}) & \cellcolor{best} $0.412$ (\small\textcolor{gray}{$0.014$}) & \cellcolor{best} $121.67$ (\small\textcolor{gray}{$5.68$}) \\
 \hline
\multirow{3}{*}{
\makecell[c]{\textsc{Pointmass}\\\textsc{Maze}}
} & \vtdfshort & $0.091$ (\small\textcolor{gray}{$0.003$}) & $1.156$ (\small\textcolor{gray}{$0.081$}) & $758.16$ (\small\textcolor{gray}{$103.54$}) \\
 & \ctdfshort & $0.089$ (\small\textcolor{gray}{$0.008$}) & $4.340$ (\small\textcolor{gray}{$0.456$}) & $679.29$ (\small\textcolor{gray}{$20.11$}) \\
 & \tdfshort & \cellcolor{best} $0.021$ (\small\textcolor{gray}{$0.000$}) & \cellcolor{best} $-0.806$ (\small\textcolor{gray}{$0.017$}) & \cellcolor{best} $9.22$ (\small\textcolor{gray}{$1.40$}) \\
 \hline
\multirow{3}{*}{
\textsc{Quadruped}
} & \vtdfshort & $15.63$ (\small\textcolor{gray}{$0.09$}) & $1.478$ (\small\textcolor{gray}{$0.088$}) & $273.68$ (\small\textcolor{gray}{$34.07$}) \\
\bottomrule
\end{tabular}

    }
\end{table}
\begin{table}[H]
    \centering
    \renewcommand{\arraystretch}{2}
    \caption{\textbf{Per task} results for the \textbf{single policy} experiments with a \textbf{curved} conditional path.}
    \label{app:results.curved.task}
    \vspace{-5mm}
    %
\end{figure}

\begin{table}[H]
    \centering
    \renewcommand{\arraystretch}{1.5}
    \caption{\textbf{Per task} results for the quantitative \textbf{multi-policy} experiments.}
    \label{tab:multipolicy}
    \vspace{-5mm}
    %
\end{table}

\begin{table}[H]
    \centering
    \renewcommand{\arraystretch}{1.5}
    \caption{\textbf{Per task} results for \textbf{planning with GPI}.}
    \label{tab:gpi.per_task}
    \vspace{-5mm}
    %
\caption{Qualitative samples generated with \vtdfshort, \vtddshort, \tdvaeshort, and \tdganshort methods for various discount factors $\gamma$ on the \textsc{Loop} task in the \textsc{Pointmass Maze} domain. The last row depicts ground truth discounted occupancies.}
    \label{fig:pointmass_media}
\end{figure}

\clearpage

\section{Theoretical Results}\label{app:proofs}

\subsection{Proofs of Main Results}

\begin{metaframe}
\lemvfonestepbootstrap*
\end{metaframe}
\begin{proof}
By \autoref{lem:mix-vf-minimization}, we have that
\begin{align*}
    v_t^{(n+1)}(x \mid s, a) = \frac{(1-\gamma) \onestepterm{p}_t(x | s, a) \onestepterm{v}_t(x \mid s, a) + \gamma \bootterm{p}_t^{(n)}(x | s,a) \bootterm{v}_t^{(n)}(x \mid s,a)}{m_t^{(n+1)}(x | s,a)},
\end{align*}
where $m_t^{(n+1)}(x | s,a) = (1-\gamma) \onestepterm{p}_t(x | s, a) + \gamma \bootterm{p}_t^{(n)}(x | s,a)$. \autoref{lem:mix_vf} implies that $m_t^{(n+1)}$ is the probability path generated by $v_t^{(n+1)}$. It is easy to see that $m_0^{(n+1)} = m_0$ since $\onestepterm{p}_0 = \bootterm{p}_0^{(n)} = m_0$. Moreover, since $\onestepterm{p}_1 = P$ and $\bootterm{p}_1^{(n)} = P^\pi m_1^{(n)}$ by assumption, $m_1^{(n+1)} = (1-\gamma) P + \gamma P^\pi m_1^{(n)} = \mathcal{T}^\pi m_1^{(n)}$, which proves the result.
\end{proof}

\begin{metaframe}
\thmprobpathoperator*
\end{metaframe}
\begin{proof}
To prove that the iterates of the three algorithms satisfy a Bellman-like update through the operator $\mathcal{B}^\pi_t$ we only need to apply \autoref{prop:tdf-bellman} for \tdfshort, \autoref{th:vtdf-bellman} for \vtdfshort, and \autoref{th:ctdf-bellman} for \ctdfshort. That $\mathcal{B}_t$ is a $\gamma$-contraction in 1-Wasserstein distance can be seen by applying \autoref{th:contraction} with $k=1$.
\end{proof}

\begin{metaframe}
\corconvergence*
\end{metaframe}
\begin{proof}
That $\mathcal{B}^\pi_t$ has a unique fixed point $\bar m_t$ to which every sequence $m_t^{(n)}$ converges to is a consequence of the Banach fixed point theorem applied on the space of all probability paths $m_t : \setfont{S} \times \setfont{A} \rightarrow \mathscr{P}(\mathbb{R}^d)$ equipped with the sup-1-Wasserstein metric. By inspecting the definition of $\mathcal{B}^\pi_t$, it is easy to see that $\bar m_t = (I - \gamma P^\pi)^{-1} P_t$. Since $P_t(x | s,a) = \int p_{t|1}(x|x_1) P(x_1|s,a) \mathrm{d}x_1$,
\begin{align*}
    \bar m_t (x | s,a) = [(I - \gamma P^\pi)^{-1} P_t](x | s,a) = \int p_{t|1}(x|x_1) \underbrace{[(I - \gamma P^\pi)^{-1} P](x_1 | s,a)}_{= m^\pi(x_1 | s,a)} \mathrm{d}x_1 = m^{\text{MC}}_t(x|s,a).
\end{align*}
\end{proof}

\begin{metaframe}
\thmvartdfvtdfmain*
\end{metaframe}
\begin{proof}
See \autoref{th:var-tdf-vtdf}.
\end{proof}

\begin{metaframe}
\thmvarctdfmain*
\end{metaframe}
\begin{proof}
See \autoref{app:thm.variance.tdvscoupled}.
\end{proof}

\subsection{General Results}

\begin{metaframe}
\begin{lemma}\label{lem:mix_vf}
    Let $v^1_t$ and $v^2_t$ be vector fields that generate the probability paths $p^1_t$ and $p^2_t$, respectively. Then, for any $\gamma \in [0,1]$, the mixture probability path $p_t = (1-\gamma) p^1_t+ \gamma p^2_t$ is generated by the vector field
    \begin{align}\label{eq:mix-vf}
        v_t := \frac{(1-\gamma) p^1_t v^1_t + \gamma p^2_t v^2_t }{(1-\gamma) p^1_t + \gamma p^2_t}.
    \end{align}
\end{lemma}
\end{metaframe}
    
    \begin{proof}
        Since $v_1^t$ (resp. $v_2^t$) generates $p^1_t$ (resp. $p^2_t$), we know from the continuity equation that:
        \begin{align*}
            \frac{\partial p^1_t}{\partial t} = \text{div} (p^1_t v^1_t), \quad
            \frac{\partial p^2_t}{\partial t} = \text{div} (p^2_t v^2_t),
        \end{align*}
        where $\text{div}$ denotes the divergence operator. Then, by linearity of $\text{div}$,
        \begin{align*}
             \frac{\partial p_t }{\partial t}& = \frac{\partial \left( (1-\gamma) p^1_t+ \gamma p^2_t \right)}{\partial t} \\
             & = (1-\gamma) \text{div} (p^1_t v^1_t) + \gamma \text{div} (p^2_t v^2_t) \\
            & = \text{div} \left( (1-\gamma) p^1_t v^1_t + \gamma p^2_t v^2_t  \right) \\
            & = \text{div} \left( \frac{(1-\gamma) p^1_t v^1_t + \gamma p^2_t v^2_t}{(1-\gamma) p^1_t+ \gamma p^2_t}  \left((1-\gamma) p^1_t+ \gamma p^2_t \right) \right) \\
            & = \text{div} \left( \frac{(1-\gamma) p^1_t v^1_t + \gamma p^2_t v^2_t}{(1-\gamma) p^1_t+ \gamma p^2_t}  p_t \right)
            \\ &= \text{div} (v_t p_t).
        \end{align*}
        Hence, $(v_t, p_t)$ satisfies the continuity equation, which implies that $v_t$ generates $p_t$.
\end{proof}

\begin{metaframe}
\begin{lemma}\label{lem:mix-vf-minimization}
    Let $v^1_t$ and $v^2_t$ be vector fields that generate the probability paths $p^1_t$ and $p^2_t$, respectively. For $\gamma \in [0,1]$, the vector field $v_t = \frac{(1-\gamma) p^1_t v^1_t + \gamma p^2_t v^2_t }{(1-\gamma) p^1_t + \gamma p^2_t}$ satisfies
    \begin{align*}
        v_t = \argmin_{v: \mathbb{R}^d\rightarrow \mathbb{R}^d} \Big \{ (1-\gamma) \E_{x_t \sim p^1_t}{\| v_t(x_t) - v^1_t(x_t)\|^2} + \gamma \E_{x_t \sim p^2_t}{\| v_t(x_t) - v^2_t(x_t)\|^2} \Big \}.
    \end{align*}
\end{lemma}
\end{metaframe}
\begin{proof}
    Let $\ell_t(v) := (1-\gamma) \E_{x_t \sim p^1_t}{\| v_t(x_t) - v^1_t(x_t)\|^2} + \gamma \E_{x_t \sim p^2_t}{\| v_t(x_t) - v^2_t(x_t)\|^2}$. The functional derivative of this quantity wrt $v$ evaluated at some point $x$ is
    \begin{align*}
        \nabla_v \ell_t(v)(x) = (1-\gamma) p_1^t(x) ( v_t(x) - v^1_t(x) ) + \gamma  p_2^t(x) ( v_t(x) - v^2_t(x) ).
    \end{align*}
    Setting this to zero and solving for $v_t(x)$ yields the result.
\end{proof}

\subsection{Analysis of TD\texorpdfstring{\textsuperscript{2}}{2}-CFM}

We study the learning dynamics of an idealized variant of \tdfshort~ which minimizes the flow-matching loss exactly. Starting from an arbitrary vector field $v_t^{(0)}$, at each iteration $n \geq 0$ we compute
\begin{align}\label{eq:tdf-ideal}
    v_t^{(n+1)}(\cdot | s,a) \in \argmin_{v :  \mathbb{R}^d \rightarrow \mathbb{R}^d} \ell_{\text{\tdfshort}}^{(n)}(t,s,a),
\end{align}
where
\begin{gather*}
    \ell_{\text{\tdfshort}}^{(n)}(t,s,a) := (1-\gamma) \onestepterm{\ell}(t,s,a) + \gamma \bootterm{\ell}(t,s,a)  \\
    \onestepterm{\ell}(t,s,a) := 
    \mathbb{E}_{S'\sim P(\cdot | s,a), X_t \sim p_{t|1}(\cdot|S')}\Big[\big \| v(X_t| s, a) - u_t(X_t |S')\big \|^2\Big] \\
      \bootterm{\ell}(t,s,a) := \mathbb{E}_{S'\sim P(\cdot | s,a), X_t \sim m_t^{(n)}(\cdot|s',\pi(s'))}\Big[\big \| v(X_t| s, a) - v_t^{(n)}(X_t | S',\pi(S'))\big \|^2\Big],
\end{gather*}
and $m_t^{(n)}(x | s,a)$ is the probability path generated by $v_t^{(n)}(x | s,a)$.

\begin{metaframe}
\begin{lemma}\label{lem:tdf-vn}
    For any $n \geq 0$, the vector field minimizing \eqref{eq:tdf-ideal} is
    \begin{align*}
        &v^{(n+1)}_t(x \mid s, a)  =\\ &\quad\quad\frac{(1-\gamma) \int u_{t \mid 1}(x \mid x_1) p_{t\mid 1}( x \mid x_1) P(x_1 | s,a)\mathrm{d}x_1 + \gamma \mathbb{E}_{S'\sim P(\cdot | s,a)}[m_t^{(n)}(x|S',\pi(S')) v_t^{(n)}(x | S',\pi(S'))]}{m_t^{(n+1)}(x | s,a)}
    \end{align*}
    where we define $m_t^{(n+1)}(x | s,a) := (1-\gamma) P_t(x|s,a) + \gamma \mathbb{E}_{S'\sim P(\cdot | s,a)}[m_t^{(n)}(x|S',\pi(S'))]$ and $P_t(x | s,a) := \int p_{t\mid 1}( x \mid x_1) P(x_1 | s,a)\mathrm{d}x_1$. Moreover $v^{(n+1)}_t$ generates $m_t^{(n+1)}$.
\end{lemma}
\end{metaframe}
\begin{proof}
    By Theorem 2 of \cite{lipman2022flow}, we have for the first term in $\ell_{\text{\tdfshort}}$
    \begin{align*}
        \nabla_\theta \onestepterm{\ell}(t, s, a) & = \nabla_\theta  
    \mathbb{E}_{X_t \sim P_{t}(\cdot|s,a)}\Big[\big \| v_t(X_t| s, a) - \onestepterm{v}_t(X_t | s,a)\big \|^2\Big],
    \end{align*}
    where $P_t(x | s,a) := \int p_{t\mid 1}( x \mid x_1) P(x_1 | s,a)\mathrm{d}x_1$, $\onestepterm{v}_t(x | s,a) = \frac{\int u_{t \mid 1}(x \mid x_1) p_{t\mid 1}( x \mid x_1) P(x_1 | s,a)\mathrm{d}x_1}{P_t(x | s,a)}$. 
    Similarly, we have for the second term:
    \begin{align*}
        \nabla_\theta \bootterm{\ell}(t, s, a) & = \nabla_\theta  
    \mathbb{E}_{X_t \sim \bootterm{p}^{(n)}_{t}(\cdot|s,a)}\Big[\big \| v_t(X_t| s, a) - \bootterm{v}_t(X_t | s,a)\big \|^2\Big],
    \end{align*}
    where $\bootterm{p}^{(n)}_t = P^\pi m^{(n)}_t $ and $\bootterm{v}_t= \frac{P^\pi( m^{(n)}_t v^{(n)}_t)}{P^\pi m^{(n)}_t}$.
    
    Therefore, $\ell_{\text{\vtdfshort}}^{(n)}(t,s,a)$ is equivalent, in term of gradient, to a mixture of two marginal flow-matching losses, which implies that $v^{(n+1)}_t$ has the stated expression by \autoref{lem:mix-vf-minimization}. The fact that it generates $m_t^{(n+1)}$ is a consequence of \autoref{lem:mix_vf}.
\end{proof}

We then define the following operator to characterize the iterates of \tdfshort.

\begin{metaframe}
\begin{definition}[Bellman operator for probability paths]\label{def:Lt-operator}
    For any $t\in[0,1]$, we define the operator $\mathcal{B}_t^\pi m := (1-\gamma) P_t + \gamma P^\pi m$, where $P_t(x | s,a) := \int p_{t\mid 1}( x \mid x_1) P(x_1 | s,a)\mathrm{d}x_1$.
\end{definition}
\end{metaframe}

The following observation is then immediate from \autoref{lem:tdf-vn}.

\begin{metaframe}
\begin{proposition}\label{prop:tdf-bellman}
    For any $n \geq 0$, the probability path generated by \tdfshort~ satisfies $m_t^{(n+1)}(x | s,a) = \big(\mathcal{B}_t^\pi m_t^{(n)}\big)(x\mid s,a)$, where $\mathcal{B}_t^\pi$ is the operator of \autoref{def:Lt-operator}. Moreover, $m_1^{(n+1)}(x|s,a) = \big(\mathcal{T}^\pi m_1^{(n)}\big)(x\mid s,a)$.
\end{proposition}
\end{metaframe}

\begin{metaframe}
\begin{theorem}\label{th:contraction}
    For any $t \in [0,1]$, the operator $\mathcal{B}_t^\pi$ of \autoref{def:Lt-operator} is a $\gamma^{1/k}$-contraction in Wasserstein k-distance, i.e., for any couple of probability paths $p_t, q_t$ and $k\in[1,\infty)$,
    \begin{align*}
        \sup_{s,a} W_k\left( \big(\mathcal{B}_t^\pi p_t\big)(\cdot \mid s, a), \big(\mathcal{B}_t^\pi q_t\big)(\cdot \mid s, a)\right) \leq \gamma^{1/k} \sup_{s, a} W_k\left(p_t(\cdot \mid s, a), q_t(\cdot \mid s, a)\right).
    \end{align*}
\end{theorem}
\end{metaframe}
    \begin{proof}
        Recall that the Wasserstein k-distance between $p_t$ and $q_t$ induced by a metric $d$ is defined as
        \begin{align*}
            W_k(p_t(\cdot | s,a), q_t(\cdot | s,a)) := \inf_{\Gamma(\cdot | s,a) \in \mathcal{C}(p_t(\cdot | s,a), q_t(\cdot | s,a))} \mathbb{E}_{(X,Y) \sim \Gamma(\cdot | s,a)} [d(X,Y)^k]^{1/k},
        \end{align*}
        where $\mathcal{C}(p_t(\cdot | s,a), q_t(\cdot | s,a))$ is the set of all couplings between the two measures. Now take any coupling $\tilde\Gamma(\cdot | s,a) \in \mathcal{C}(p_t(\cdot | s,a), q_t(\cdot | s,a))$ for any $s,a$. Then, the following quantity
        \begin{align*}
            \Theta(x,y | s,a) = (1-\gamma) P(x | s,a) \delta(x - y) + \gamma \big(P^\pi \tilde\Gamma\big)(x,y | s,a)
        \end{align*}
        is a valid coupling between $\big(\mathcal{B}_t^\pi p_t\big)(\cdot \mid s, a)$ and $\big(\mathcal{B}_t^\pi q_t\big)(\cdot \mid s, a)$. In fact,
        \begin{align*}
            \int \Theta(x, y | s,a) \mathrm{d}x 
            & = (1-\gamma) \int P(x | s,a) \delta(x - y) \mathrm{d}x + \gamma \int \big(P^\pi \tilde\Gamma\big)(x,y \mid s,a) \mathrm{d}x
            \\ & = (1-\gamma) P(y | s,a) + \gamma \int \E_{s'\sim P(\cdot|s,a)}{\tilde\Gamma(x,y|s',\pi(s'))} \mathrm{d}x
            \\ &= (1-\gamma) P(y | s,a) + \gamma \E_{s'\sim P(\cdot|s,a)}{\int \tilde\Gamma(x,y|s',\pi(s')) \mathrm{d}x}
            \\ &= (1-\gamma) P(y | s,a) + \gamma \E_{s'\sim P(\cdot|s,a)}{q_t(y|s',\pi(s'))}
            \\ &= \big(\mathcal{T}^\pi q_t\big)(y|s,a).
        \end{align*}
        Analogously, we can prove that $\int \Theta(x, y | s,a) \mathrm{d}y = \big(\mathcal{B}^\pi p_t\big)(x|s,a)$. Then,
        \begin{align*}
            W_k\left(\big(\mathcal{B}_t^\pi p_t\big)(\cdot \mid s, a), \big(\mathcal{B}_t^\pi q_t\big)(\cdot \mid s, a)\right)
            &= \inf_{\Gamma(\cdot | s,a) \in \mathcal{C}([\mathcal{L}_t^\pi p_t](\cdot | s,a), [\mathcal{L}_t^\pi q_t](\cdot | s,a))} \mathbb{E}_{(X,Y) \sim \Gamma(\cdot | s,a)} [d(X,Y)^k]^{1/k}
            \\ &\leq \mathbb{E}_{(X,Y) \sim \Theta(\cdot | s,a)} [d(X,Y)^k]^{1/k}
            \\ &=  \left( (1-\gamma) \mathbb{E}_{(X \sim P(\cdot | s,a), Y \sim \delta_X)} [d(X,Y)^k] + \gamma  \mathbb{E}_{(X,Y) \sim [P^\pi \tilde\Gamma](\cdot | s,a)} [d(X,Y)^k] \right)^{1/k}
            \\ &= \gamma^{1/k} \mathbb{E}_{s'\sim P(\cdot | s,a), (X,Y) \sim \tilde\Gamma(\cdot | s',\pi(s'))} [d(X,Y)^k]^{1/k}.
        \end{align*}
        Since this holds for any coupling $\tilde\Gamma(\cdot | s,a) \in \mathcal{C}(p_t(\cdot | s,a), q_t(\cdot | s,a))$, we can take the infimum over all such couplings on the right-hand side, so that
        \begin{align*}
            W_k\left(\big(\mathcal{B}_t^\pi p_t\big)(\cdot \mid s, a), \big(\mathcal{B}_t^\pi q_t\big)(\cdot \mid s, a)\right)  
            &\leq \gamma^{1/k} \left( \mathbb{E}_{s'\sim P(\cdot | s,a)} \left[ \inf_{\Gamma \in \mathcal{C}(p_t(\cdot | s',\pi(s')), q_t(\cdot | s',\pi(s')))} \mathbb{E}_{(X,Y) \sim \Gamma} [d(X,Y)^k] \right] \right)^{1/k}
            \\ &= \gamma^{1/k} \left( \mathbb{E}_{s'\sim P(\cdot | s,a)} \left[ W_k(p_t(\cdot | s',\pi(s')), q_t(\cdot | s',\pi(s')))^k \right] \right)^{1/k}
            \\ &\leq \gamma^{1/k} \sup_{s, a} W_k(p_t(\cdot \mid s, a), q_t(\cdot \mid s, a)).
        \end{align*}
        Taking the supremum over $(s,a)$ of the left-hand side concludes the proof.
    \end{proof}

\subsection{Analysis of TD-CFM}\label{app:analysis.vtdf}

We study the learning dynamics of an idealized variant of \vtdfshort~ which minimizes the flow-matching loss exactly. Starting from an arbitrary vector field $v_t^{(0)}$, at each iteration $n \geq 0$ we compute
\begin{align}\label{eq:vtdf-ideal}
    v_t^{(n+1)}(\cdot | s,a) \in \argmin_{v_t(\cdot) :  \mathbb{R}^d \rightarrow \mathbb{R}^d} \ell_{\text{\vtdfshort}}^{(n)}(t,s,a) := \mathbb{E}_{X_1\sim \big(\mathcal{T}^\pi m_1^{(n)}\big)(s,a), X_t \sim p_{t|1}(\cdot | X_1)} \Big[ \|v_t(X_t) - u_{t|1}(X_t|X_1)\|^2\Big],
\end{align}
where $m_t^{(n)}(x | s,a)$ is the probability path generated by $v_t^{(n)}(x | s,a)$.

\begin{metaframe}
\begin{lemma}\label{lem:vtdf-vn}
    For any $n \geq 0$, the vector field minimizing \eqref{eq:vtdf-ideal} is
    \begin{align*}
        v^{(n+1)}_t(x \mid s, a)  & = \int u_{t|1}(x | x_1) \frac{ p_{t\mid 1}( x \mid x_1)  \big(\mathcal{T}^\pi m_1^{(n)}\big)(x_1 \mid s, a) }{ m_t^{(n+1)}(x | s,a)  } \mathrm{d}x_1,
    \end{align*}
    where $m_t^{(n+1)}(x | s,a) := \int p_{t\mid 1}( x \mid x_1) \big(\mathcal{T}^\pi m_1^{(n)}\big)(x_1 \mid s, a) \mathrm{d}x_1$. Moreover $v^{(n+1)}_t$ generates $m_t^{(n+1)}$.
\end{lemma}
\end{metaframe}
\begin{proof}
    Note that \eqref{eq:vtdf-ideal} is a standard flow matching loss for the target distribution $\mathcal{T}^\pi m_1^{(n)}$. The expression of $v^{(n+1)}_t(x \mid s, a)$ given in the statement is exactly the vector field obtained by marginalization of the conditional vector field $u_{t|1}$, which we know to be the minimizer of the loss from Theorem 2 of \cite{lipman2022flow}. The fact that $v^{(n+1)}_t$ generates $m_t^{(n+1)}$ is a consequence of Theorem 1 of \cite{lipman2022flow}.
\end{proof}

\begin{metaframe}
\begin{lemma}\label{lem:vtdf-m1}
    For any $n \geq 0$, the probability path generated by \eqref{eq:vtdf-ideal} satisfies $m_1^{(n+1)}(x|s,a) = \big(\mathcal{T}^\pi m_1^{(n)}\big)(x|s,a)$.
\end{lemma}
\end{metaframe}
\begin{proof}
    This is immediate from the definition of conditional probability path, as we set $p_{1\mid 1}( x \mid x_1) = \delta(x - x_1)$ by construction, where $\delta(\cdot)$ is the Dirac's delta function.
\end{proof}

\begin{metaframe}
\begin{theorem}\label{th:vtdf-bellman}
    For any $n \geq 1$, the probability path generated by \eqref{eq:vtdf-ideal} satisfies 
    $$m_t^{(n+1)}(x | s,a) = \big(\mathcal{B}_t^\pi m_t^{(n)}\big)(x|s,a),$$
    where $\mathcal{B}_t^\pi$ is the operator of \autoref{def:Lt-operator}. Moreover, if the initial vector field $v_t^{(0)}$ satisfies
    $$ v^{(0)}_t(x \mid s, a) = \int u_{t|1}(x | x_1) \frac{ p_{t\mid 1}( x \mid x_1)  m_1^{(0)}(x_1 \mid s, a) }{ m_t^{{(0)}}(x | s,a)  } \mathrm{d}x_1,$$
    with $m_t^{{(0)}}$ being its generated proability path, then this result is valid at all $n \geq 0$.
\end{theorem}
\end{metaframe}
\begin{proof}
    We know that, for all $n \geq 0$, $v^{n+1}_t$ generates $m_t^{(n+1)}$ (\autoref{lem:vtdf-vn}) and that $m_1^{(n+1)} = \mathcal{T}^\pi m_1^{(n)}$ (\autoref{lem:vtdf-m1}). Note that $m_t^{(n+1)}$ is written as a function of $m_1^{(n)}$ only, i.e., at each iteration we keep only the distribution generated at time $t=1$ ($m_1^{(n)}$) and discard the associated probability path ($m_t^{(n)}$ for $t < 1$). We can however express $m_t^{(n+1)}$ as a function of $m_t^{(n)}$ thanks to the linearity of the Bellman operator and the definition of marginal paths. For any $n \geq 1$,
    \begin{align*}
        m_t^{(n+1)}(x \mid s,a) 
        :&= \int p_{t\mid 1}( x \mid x_1) \big(\mathcal{T}^\pi m_1^{(n)}\big)(x_1 \mid s, a) \mathrm{d}x_1
        \\ &= \int p_{t\mid 1}( x \mid x_1) \left((1-\gamma) P(x_1 \mid s,a) + \gamma \E_{s'\sim P(\cdot \mid s,a)}{m_1^{(n)}(x_1 \mid s', \pi(s'))} \right) \mathrm{d}x_1
        \\ &= (1-\gamma) \int p_{t\mid 1}( x \mid x_1) P(x_1 \mid s,a)\mathrm{d}x_1 + \gamma \E_{s'\sim P(\cdot \mid s,a)}{\int p_{t\mid 1}( x \mid x_1) m_1^{(n)}(x_1 \mid s', \pi(s')) \mathrm{d}x_1}
        \\ &= (1-\gamma) \int p_{t\mid 1}( x \mid x_1) P(x_1 \mid s,a)\mathrm{d}x_1 + \gamma \E_{s'\sim P(\cdot \mid s,a)}{ \int p_{t\mid 1}( x \mid x_1) \big(\mathcal{T}^\pi m_1^{(n-1)}\big)(x_1 \mid s', \pi(s')) \mathrm{d}x_1}
        \\ &= (1-\gamma) \int p_{t\mid 1}( x \mid x_1) P(x_1 \mid s,a)\mathrm{d}x_1 + \gamma \E_{s'\sim P(\cdot \mid s,a)}{ m_t^{(n)}(x \mid s', \pi(s'))}
        \\ &= (1-\gamma) P_t(x | s,a) + \gamma \E_{s'\sim P(\cdot \mid s,a)}{ m_t^{(n)}(x \mid s', \pi(s')) }
        = \big(\mathcal{B}_t^\pi m_t^{(n)}\big)(x \mid s,a).
    \end{align*}
    This proves the first part of the statement.
    For the second part, we only need to prove that the result also holds at $n=0$. Note that the assumption on $v_t^{(0)}$ implies that $m_t^{(0)}(x \mid s,a) := \int p_{t\mid 1}( x \mid x_1) m_1^{(0)}(x_1 \mid s, a) \mathrm{d}x_1$. Thus,
    \begin{align*}
        m_t^{(1)}(x \mid s,a) 
        :&= \int p_{t\mid 1}( x \mid x_1) \big(\mathcal{T}^\pi m_1^{(0)}\big)(x_1 \mid s, a) \mathrm{d}x_1
        \\ &= \int p_{t\mid 1}( x \mid x_1) \left((1-\gamma) P(x_1 \mid s,a) + \gamma \E_{s'\sim P(\cdot | s,a)}{m_1^{(0)}(x_1 \mid s', \pi(s'))} \right) \mathrm{d}x_1
        \\ &= (1-\gamma) \int p_{t\mid 1}( x \mid x_1) P(x_1 \mid s,a)\mathrm{d}x_1 + \gamma \E_{s'\sim P(\cdot \mid s,a)}{\int p_{t\mid 1}( x \mid x_1) m_1^{(0)}(x_1 \mid s', \pi(s')) \mathrm{d}x_1}
        \\ &= (1-\gamma) \int p_{t\mid 1}( x \mid x_1) P(x_1 \mid s,a)\mathrm{d}x_1 + \gamma \E_{s'\sim P(\cdot \mid s,a)}{ m_t^{(0)}(x \mid s', \pi(s'))}
        = \big(\mathcal{B}_t^\pi m_t^{(0)}\big) (x \mid s,a).
    \end{align*}
\end{proof}

\subsection{Analysis of TD-CFM(C)}\label{app:analysis.ctd}

The idealized update of \ctdfshort~ is, for any $n \geq 0$,
\begin{equation}\label{eq:ctdf-ideal}
\begin{gathered}
    v_t^{(n+1)}(\cdot | s,a) \in \argmin_{v_t(\cdot) :  \mathbb{R}^d \rightarrow \mathbb{R}^d} \ell_{\text{\ctdfshort}}^{(n)}(t,s,a)\,, \text{where} \\
    \ell_{\text{\ctdfshort}}^{(n)}(t,s,a) := \E_{(X_0,X_1) \sim \Gamma_{0,1}^{(n)}(\cdot | s,a), X_t \sim p_{t\mid 0,1}(\cdot \mid X_0,X_1)}{ \|v_t(X_t) - u_{t\mid0,1}(X_t\mid X_0,X_1)\|^2},
\end{gathered}
\end{equation}
and $\Gamma_{0,1}^{(n)}(\cdot \mid s,a)$ is the coupling between $m_0$ and $\mathcal{T}^\pi m_1^{(n)}$, while $p_{t\mid 0,1}, u_{t\mid 0,1}$ are such that $u_{t\mid 0,1}(x \mid x_0, x_1)$ generates $p_{t\mid 0,1}( x \mid x_0, x_1)$, $p_{1\mid 0,1}( x \mid x_0, x_1) = \delta_{x_1}(x)$, and
\begin{align}\label{eq:ctdf-pt01}
    p_{t\mid 1}( x \mid x_1) = \int p_{t\mid 0,1}( x \mid x_0, x_1) m_0(x_0) \mathrm{d}x_0.
\end{align}

\begin{metaframe}
\begin{lemma}\label{lem:ctdf-coupling}
    The coupling $\Gamma_{0,1}^{(n)}(\cdot \mid s,a)$ satisfies
    \begin{align*}
        \Gamma_{0,1}^{(n)}(x_0, x_1 \mid s,a) &= (1-\gamma)P(x_1 \mid s,a) m_0(x_0) + \gamma \E_{S' \sim P(\cdot \mid s,a)}{m_{0,1}^{(n)}(x_0, x_1 \mid S',\pi(S'))},
    \end{align*}
    where $m_{0,1}^{(n)}(x_0, x_1 \mid s,a) = m_{0}(x_0) \delta_{\psi^{(n)}_1(x_0 \mid s, a)}(x_1)$ is the joint distribution of $(X_0, X_1)$, \textit{i.e} the endpoints of the ODE.
\end{lemma}
\end{metaframe}
\begin{proof}
    For any $x_0,x_1$, we can write $\Gamma_{0,1}^{(n)}(x_0, x_1 \mid s,a) = \Gamma_{1}^{(n)}(x_1 \mid s,a, x_0) m_0(x_0)$, where $\Gamma_{1}^{(n)}$ is the corresponding conditional distribution. By definition, we have
    \begin{align*}
        \Gamma_{1}^{(n)}(x_1 \mid s,a, x_0) 
        &= (1-\gamma)P(x_1 \mid s,a) + \gamma \E_{s' \sim P(\cdot \mid s,a)}{ \delta_{\psi_1^{(n)}(x_0 \mid s',\pi(s'))}(x_1)}
    \end{align*}
    where $\psi_1^{(n)}$ is the flow that generates $m_1^{(n)}$. Multiplying both sides by $m_0(x_0)$ and using that $m_{0,1}^{(n)}(x_0, x_1 \mid s,a) = m_{0}(x_0) \delta_{\psi^{(n)}_1(x_0 \mid s, a)}(x_1)$ concludes the proof.
\end{proof}

\begin{metaframe}
\begin{lemma}\label{lem:ctdf-vn}
    For any $n \geq 0$, the vector field minimizing \eqref{eq:ctdf-ideal} is
    \begin{align*}
        v^{(n+1)}_t(x \mid s, a)  & = \int \int u_{t\mid 0,1}(x \mid x_0, x_1) \frac{ p_{t\mid 0,1}( x \mid x_0, x_1)  \Gamma_{0,1}^{(n)}(x_0, x_1 \mid s,a) }{ m_t^{(n+1)}(x \mid s,a)  } \mathrm{d}x_0 \mathrm{d}x_1,
    \end{align*}
    where $m_t^{(n+1)}(x \mid s,a) := \int\int p_{t\mid 0,1}( x \mid x_0,x_1) \Gamma_{0,1}^{(n)}(x_0, x_1 \mid s,a) \mathrm{d}x_0 \mathrm{d}x_1$. Moreover $v^{(n+1)}_t$ generates $m_t^{(n+1)}$.
\end{lemma}
\end{metaframe}
\begin{proof}
    Note that \eqref{eq:ctdf-ideal} is a standard conditional flow matching loss since $u_{t\mid 0,1}(x \mid x_0, x_1)$ generates $p_{t\mid 0,1}( x \mid x_0, x_1)$ and $p_{1\mid 0,1}( x \mid x_0, x_1) = \delta_{x_1}(x)$. The expression of $v^{(n+1)}_t(x \mid s, a)$ given in the statement is exactly the vector field obtained by marginalization of the conditional vector field $u_{t\mid 0,1}$, which we know to be the minimizer of the loss from Theorem 2 of \cite{lipman2022flow}. The fact that $v^{(n+1)}_t$ generates $m_t^{(n+1)}$ is a consequence of Theorem 1 of \cite{lipman2022flow}.
\end{proof}\

\begin{metaframe}
\begin{lemma}\label{lem:ctdf-m1}
    For any $n \geq 0$, the probability path generated by \eqref{eq:vtdf-ideal} satisfies $m_1^{(n+1)}(x\mid s,a) = \big(\mathcal{T}^\pi m_1^{(n)}\big)(x\mid s,a)$.
\end{lemma}
\end{metaframe}
\begin{proof}
    By \autoref{lem:ctdf-vn} and the fact that $p_{1\mid 0,1}( x \mid x_0, x_1) = \delta_{x_1}(x)$,
    \begin{align*}
        m_1^{(n+1)}(x \mid s,a) &:= \int\int p_{1\mid 0,1}( x \mid x_0,x_1) \Gamma_{0,1}^{(n)}(x_0, x_1 \mid s,a) \mathrm{d}x_0 \mathrm{d}x_1
        \\ &= \int \Gamma_{0,1}^{(n)}(x_0, x \mid s,a) \mathrm{d}x_0
        \\ &= \big(\mathcal{T}^\pi m_1^{(n)}\big)(x|s,a).
    \end{align*}
\end{proof}

\begin{metaframe}
\begin{theorem}\label{th:ctdf-bellman}
    For any $n \geq 1$, the probability path generated by \eqref{eq:vtdf-ideal} satisfies 
    $$m_t^{(n+1)}(x \mid s,a) = \big(\mathcal{B}_t^\pi m_t^{(n)}\big)(x\mid s,a),$$
    where $\mathcal{B}_t^\pi$ is the operator of \autoref{def:Lt-operator}. Moreover, if the initial vector field $v_t^{(0)}$ satisfies
    $$v^{(0)}_t(x \mid s, a) = \int\int u_{t\mid 0,1}(x | x_0, x_1) \frac{ p_{t\mid 0,1}( x \mid x_0, x_1)  m_{0,1}^{(0)}(x_0, x_1 \mid s,a) }{ m_t^{{(0)}}(x \mid s,a)  } \mathrm{d}x_0 \mathrm{d}x_1,$$
    with $m_t^{{(0)}}$ being its generated probability path, then this result is valid at all $n \geq 0$.
\end{theorem}
\end{metaframe}
\begin{proof}
    We know that, for all $n \geq 0$, $v^{n+1}_t$ generates $m_t^{(n+1)}$ (\autoref{lem:ctdf-vn}) and that $m_1^{(n+1)} = \mathcal{T}^\pi m_1^{(n)}$ (\autoref{lem:ctdf-m1}). While $m_t^{(n+1)}$ is written as a function of $\Gamma_{0,1}^{(n)}$ only, we can rewrite it as a function of $m_t^{(n)}$ thanks to the linearity of the Bellman operator and the definition of marginal paths. For any $n \geq 1$, By \autoref{lem:ctdf-coupling},
    \begin{align*}
        m_t^{(n+1)}(x \mid s,a) 
        &:= \int\int p_{t\mid 0,1}( x \mid x_0,x_1) \Gamma_{0,1}^{(n)}(x_0, x_1 \mid s,a) \mathrm{d}x_0 \mathrm{d}x_1
        \\ &= \int\int p_{t\mid 0,1}( x \mid x_0,x_1) \left( (1-\gamma)P(x_1 \mid s,a)m_0(x_0) + \gamma \E_{S' \sim P(\cdot \mid s,a)}{m_{0,1}^{(n)}(x_0, x_1 \mid S',\pi(S'))} \right) \mathrm{d}x_0 \mathrm{d}x_1
        \\ &= (1-\gamma) \underbrace{\int\int p_{t\mid 0,1}( x \mid x_0,x_1) P(x_1 \mid s,a)m_0(x_0)\mathrm{d}x_0 \mathrm{d}x_1}_{(i)}
        \\ & \qquad + \gamma \underbrace{\E_{s' \sim P(\cdot | s,a)} { \int\int p_{t\mid 0,1}( x \mid x_0,x_1) m_{0,1}^{(n)}(x_0, x_1 \mid S',\pi(S'))  \mathrm{d}x_0 \mathrm{d}x_1 }}_{(ii)}.
    \end{align*}
    By \eqref{eq:ctdf-pt01},
    \begin{align*}
        (i) = \int p_{t\mid 1}( x \mid x_1) P(x_1 \mid s,a)\mathrm{d}x_1 = P_t(x \mid s,a).
    \end{align*}
    For (ii), by Lemma~\ref{lem:ctdf-vn}, we have $m_t^{(n)}(x \mid s,a) = \int\int p_{t\mid 0,1}( x \mid x_0,x_1) \Gamma_{0,1}^{(n-1)}(x_0, x_1 \mid s,a) \mathrm{d}x_0 \mathrm{d}x_1, \forall n \geq 0$, which implies
    \begin{align*}
        m_{0,1}^{(n)}(x_0, x_1 \mid s', \pi(s')) = \Gamma_{0,1}^{(n-1)}(x_0, x_1 \mid s', \pi(s')).
    \end{align*}
    Therefore, again by definition of $m_t^{(n)}$ (\autoref{lem:ctdf-vn}),
    \begin{align*}
        (ii) &= \E_{s' \sim P(\cdot \mid s,a)} { \int\int p_{t\mid 0,1}( x \mid x_0,x_1) \Gamma_{0,1}^{(n-1)}(x_0, x_1 \mid s', \pi(s'))  \mathrm{d}x_0 \mathrm{d}x_1}
        \\ &= \E_{s' \sim P(\cdot \mid s,a)}{ m_t^{(n)}(x \mid s', \pi(s'))}.
    \end{align*}
    Plugging the expressions of (i) and (ii) into the one of $m_t^{(n+1)}(x \mid s,a)$ yields the first part of the statement.

    For the second part, we only need to prove that the result also holds at $n=0$. Note that the assumption on $v_t^{(0)}$ implies that $m_t^{(0)}(x \mid s,a) = \int\int p_{t\mid 0,1}( x \mid x_0,x_1) m_{0,1}^{(0)}(x_0, x_1 \mid s', \pi(s'))  \mathrm{d}x_0 \mathrm{d}x_1$. Thus, using the same decomposition above, we have
    \begin{align*}
        m_t^{(1)}(x \mid s,a) &= (1-\gamma) P_t(x \mid s,a) + \gamma \E_{s' \sim P(\cdot \mid s,a)}{\int\int p_{t\mid 0,1}( x \mid x_0,x_1) m_{0,1}^{(0)}(x_0, x_1 \mid s',\pi(s'))  \mathrm{d}x_0 \mathrm{d}x_1}
        \\ &= (1-\gamma) P_t(x \mid s,a) + \gamma \E_{s' \sim P(\cdot \mid s,a)}{ m_t^{(0)}(x \mid s',\pi(s'))},
    \end{align*}
    which proves the result.
\end{proof}

\subsection{Variance Analysis}\label{app:variance.analysis}

\begin{metaframe}
\begin{theorem}\label{th:var-tdf-vtdf}
Let us define the random variables
\begin{align*}
    g_{\text{\tdfshort}}(t, s, a, s', \onestepterm{X}_t, X^{(n)}_t) & :=  (1-\gamma) \nabla_\theta v_t(\onestepterm{X}_t| s, a;\theta)^\top \big ( v_t(\onestepterm{X}_t| s, a;\theta) -  u_{t|1}(\onestepterm{X}_t | s') \big )\\
    & \qquad + \gamma \nabla_\theta v_t(X^{(n)}_t| s, a;\theta)^\top \big( v_t(X^{(n)}_t| s, a;\theta) - v_t^{(n)}(X^{(n)}_t | s', \pi(s')) \big ) \\
    g_{\text{\vtdfshort}}(t, s, a, s', \onestepterm{X}_t, X_1, X_t) & := (1-\gamma) \nabla_\theta v_t(\onestepterm{X}_t| s, a;\theta)^\top \big ( v_t(\onestepterm{X}_t| s, a;\theta) -  u_{t|1}(\onestepterm{X}_t | s') \big) \\
    & \qquad + \gamma \nabla_\theta v_t(X_t| s, a;\theta)^\top \big ( v_t(X_t| s, a;\theta) -  u_{t|1}(X_t | X_1) \big)
\end{align*}
where $t \sim \mathcal{U}([0,1]), (s, a) \sim \rho, s'\sim P(\cdot| s, a)$, $\onestepterm{X}_t \sim p_{t|1}(\cdot | s'), X^{(n)}_t \sim m^{(n)}_t(\cdot \mid s', \pi(s')), X_1 \sim m^{(n)}_1(\cdot \mid s', \pi(s'))$, and $X_t \sim p_{t|1}(\cdot | X_1)$.
Then, $g_{\text{\tdfshort}}$ and $g_{\text{\vtdfshort}}$ are 
respectively unbiased estimates of the gradients $\nabla_\theta \ell_{\text{\tdfshort}}(\theta)$ and $\nabla_\theta \ell_{\text{\vtdfshort}}(\theta)$.

Moreover, if we consider their respective total variations defined as:
\begin{align*}
    \sigma^2_{\text{\tdfshort}} & = \text{Trace}\left( \text{Cov}_{t, s, a, s', \onestepterm{X}_t, X^{(n)}_t} \left [ g_{\text{\tdfshort}}(t, s, a, s', \onestepterm{X}_t, X^{(n)}_t) \right] \right) \\
    \sigma^2_{\text{\vtdfshort}} & = \text{Trace}\left( \text{Cov}_{t, s, a, s', \onestepterm{X}_t, X_1, X_t} \left [ g_{\text{\vtdfshort}}(t, s, a, s', \onestepterm{X}_t, X_1, X_t) \right] \right)
\end{align*}
and we assume that $m^{(n)}_t(x \mid s, a) = \int p_{t|1}(x \mid x_1) m^{(n)}_1(x_1 \mid s, a) \mathrm{d}x_1$, then we obtain 
\begin{equation*}
    \sigma^2_{\text{\vtdfshort}} =  \sigma^2_{\text{\tdfshort}} + \gamma^2 \mathbb{E}_{t, s, a, X_t} \left[ \text{Trace}\left( \text{Cov}_{X_1 \mid s, a, X_t} \left[ \nabla_\theta v_t(X_t| s, a;\theta)^\top u_{t|1}(X_t \mid X_1)\right] \right)\right].
\end{equation*}    
\end{theorem}
\end{metaframe}

\begin{proof}
Recall the {\tdfshort} and {\vtdfshort} objectives:
\begin{align*}
\ell_{\text{\tdfshort}}(\theta) & = (1-\gamma) \mathbb{E}_{\substack{t, s,a, s', X_t \sim p_{t|1}(\cdot | s')}} \left[ \Big \| v_t(X_t| s, a;\theta) - u_{t|1}(X_t | s') \Big \|^2\right]
\\ & \qquad\qquad + \gamma \mathbb{E}_{\substack{t, s,a,s', X_t \sim  m_t^{(n)}(\cdot|s',\pi(s'))}}\Big[\big \| v_t(X_t| s, a;\theta) - v_t^{(n)}(X_t | s', \pi(s'))\big \|^2\Big], \\
\ell_{\text{\vtdfshort}}(\theta) & = (1-\gamma) \mathbb{E}_{\substack{t, s,a, s', X_t \sim p_{t|1}(\cdot | s')}} \left[ \Big \| v_t(X_t| s, a;\theta) - u_{t|1}(X_t | s') \Big \|^2\right]
\\ & \qquad\qquad + \gamma \mathbb{E}_{{t, s,a,s', X_1 \sim  m_1^{(n)}(\cdot|s',\pi(s')), X_t \sim p_{t|1}(\cdot | X_1)}}\Big[\big \| v_t(X_t| s, a;\theta) - u_{t|1}(X_t | X_1)\big \|^2\Big].
\end{align*}
Computing the gradients of these quantities w.r.t. $\theta$, it is easy to check that $g_{\text{\tdfshort}}$ and $g_{\text{\vtdfshort}}$ are their unbiased estimates.

Let us now analyze the total variation of these estimators. By assumption, we have $m^{(n)}_t(x \mid s, a) = \int p_{t|1}(x \mid x_1) m^{(n)}_1(x_1 \mid s, a) \mathrm{d}x_1$, which implies that $X^{(n)}_t$ and $X_t$ follow the same law. Moreover, we obtain the following identities:
\begin{align*}
 v^{(n)}_t(x \mid s', \pi(s')) & = \mathbb{E}_{X_1 \mid x, s'}\left[ u_{t | 1}(x \mid X_1)\right], \\
    g_{\text{\tdfshort}}(t, s, a, s', \onestepterm{X}_t, X_t) & = \mathbb{E}_{X_1 \mid X_t, s'}\left[ g_{\text{\vtdfshort}}(t, s, a, s', X^{o}_t, X_1, X_t) \right], \\ 
    \mathbb{E}_{X_t \sim m^{(n)}_t(\cdot \mid s', \pi(s'))} \Big [ g_{\text{\tdfshort}}(t, s, a, s', \onestepterm{X}_t, X_t) \Big] & = \mathbb{E}_{\substack{X_1 \sim m^{(n)}_1(\cdot \mid s', \pi(s'))\\ X_t \sim p_{t|1}(\cdot | X_1)}} \Big [ g_{\text{\vtdfshort}}(t, s, a, s', \onestepterm{X}_t, X_1, X_t)\Big],
\end{align*}
where $X_1 \mid x, s'  \sim \frac{p_{t|1}(x| X_1) m_1^{(n)}(X_1 | s', \pi(s'))}{m^{(n)}_t(x \mid s, a)}$ is the posterior distribution of $X_1$ given $x$ and $s'$.

To simplify notation, we denote by $Y$ the random variable $(t, s, a, s', \onestepterm{X}_t)$. Using the decomposition of variance into conditional variance, $\mathrm{Var}(X) = \mathbb{E}[\mathrm{Var}(X | Y)]) + \mathrm{Var(\mathbb{E}[X | Y])}$, we conclude that
\begin{align*}
    \sigma_{\text{\vtdfshort}} &= \text{Trace}\left( \text{Cov}_{Y, X_1, X_t} \left[ g_{\text{\vtdfshort}}(Y, X_1, X_t) \right] \right) \\
    & = \mathbb{E}_{Y, X_1, X_t} \left[  \Big \| g_{\text{\vtdfshort}}(Y, X_1, X_t) - \mathbb{E}_{Y, X_1, X_t} \left[ g_{\text{\vtdfshort}}(Y, X_1, X_t)\right] \Big\|^2\right] \\
    & = \mathbb{E}_{Y, X_t} \left[  \Big \| \mathbb{E}_{X_1 \mid Y, X_t} \left[ g_{\text{\vtdfshort}}(Y, X_1, X_t)\right] - \mathbb{E}_{Y, X_1, X_t} \left[ g_{\text{\vtdfshort}}(Y, X_1, X_t)\right] \Big\|^2 \right]  \\
    & \qquad + \mathbb{E}_{Y, X_t} \left[  \mathbb{E}_{X_1 \mid Y, X_t} \Big [ \Big \| g_{\text{\vtdfshort}}(Y, X_1, X_t) - \mathbb{E}_{X_1 \mid Y, X_t} \left[ g_{\text{\vtdfshort}}(Y, X_1, X_t)\right] \Big\|^2 \Big ] \right]  \\
    & = \mathbb{E}_{Y, X_t} \left[  \Big \|  g_{\text{\tdfshort}}(Y, X_t) - \mathbb{E}_{Y, X_t} \left[ g_{\text{\tdfshort}}(Y, X_t)\right] \Big\|^2 \right]  \\
    & \qquad + \gamma^2 \mathbb{E}_{Y, X_t} \left[  \mathbb{E}_{X_1 \mid Y, X_t} \Big [ \Big \| \nabla_\theta v_t(X_t| s, a;\theta)^\top u_{t|1}(X_t \mid X_1) - \mathbb{E}_{X_1 \mid Y, X_t} \left[ \nabla_\theta v_t(X_t| s, a;\theta)^\top u_{t|1}(X_t \mid X_1) \right] \Big\|^2 \Big ] \right]  \\
    & =  \sigma_{\text{\tdfshort}} + \gamma^2 \mathbb{E}_{Y, X_t} \left[ \text{Trace}\left( \text{Cov}_{X_1 \mid Y, X_t} \left[ \nabla_\theta v_t(X_t| s, a;\theta)^\top u_{t|1}(X_t \mid X_1)\right] \right)\right] \\
    & =  \sigma_{\text{\tdfshort}} + \gamma^2 \mathbb{E}_{t, s, a, X_t} \left[ \text{Trace}\left( \text{Cov}_{X_1 \mid s, a, X_t} \left[ \nabla_\theta v_t(X_t| s, a;\theta)^\top u_{t|1}(X_t \mid X_1)\right] \right)\right].
\end{align*}
\end{proof}

\begin{metaframe}
\begin{theorem} \label{app:thm.variance.tdvscoupled}
Let us define the random variable 
\begin{align*}
    g_{\text{\ctdfshort}}(t, s, a, s', \onestepterm{X}_t, X_0, X_1, X_t) & := (1-\gamma) \nabla_\theta v_t(\onestepterm{X}_t| s, a;\theta)^\top \big ( v_t(\onestepterm{X}_t| s, a;\theta) -  u_{t|0,1}(\onestepterm{X}_t | X_0, s') \big) \\
    & \qquad + \gamma  \nabla_\theta v_t(X_t| s, a;\theta)^\top \big ( v_t(X_t| s, a;\theta) -  u_{t|0,1}(X_t |X_0, X_1) \big)
\end{align*}
where $t \sim \mathcal{U}([0,1]), (s, a) \sim \rho, s'\sim P(\cdot| s, a), \onestepterm{X}_t \sim p_{t|1}(\cdot | s'), (X_0, X_1) \sim m^{(n)}_{0,1}(\cdot \mid s', \pi(s')) \text{ and } X_t \sim p_{t|0,1}(\cdot | X_0, X_1, )$.
Then $g_{\text{\ctdfshort}}$ is an unbiased estimate of the gradient $\nabla_\theta \ell_{\text{\ctdfshort}}(\theta)$.

Moreover, if we consider its total variation defined as:
\begin{align*}
    \sigma_{\text{\ctdfshort}} & = \text{Trace}\left( \text{Cov}_{t, s, a, s', \onestepterm{X}_t, X_0, X_1, X_t} \left[ g_{\text{\ctdfshort}}(t, s, a, s', \onestepterm{X}_t, X_0, X_1, X_t) \right] \right)
\end{align*}
and we assume that $m^{(n)}_t(x \mid s, a) = \int\int p_{t|0,1}(x \mid x_0, x_1) m^{(n)}_{0,1}(x_0, x_1 \mid s, a) \mathrm{d}x_0 \mathrm{d}x_1$, then we obtain 
\begin{align*}
    \sigma_{\text{\ctdfshort}} & =  \sigma_{\text{\tdfshort}} + \gamma^2 \mathbb{E}_{t, s, a, X_t} \left[ \text{Trace}\left( \text{Cov}_{(X_0, X_1) \mid s, a, X_t} \left[ \nabla_\theta v_t(X_t| s, a;\theta)^\top u_{t|0,1}(X_t \mid X_0, X_1)\right] \right)\right].
\end{align*}  
Furthermore, if we use straight conditional paths, \textit{i.e.}, $p_{t | 0, 1}(x | x_0, x_1) = \delta(tx_1 + (1-t)x_0 - x)$, then
\begin{align*}
     &\sigma_{\text{\ctdfshort}} \leq \sigma_{\text{\tdfshort}} \\
     &\qquad\qquad\,\,+\, \gamma^2 \sup_{t, s,a,x} \Big \| \nabla_\theta v_t(x| s, a;\theta) \Big \|^2  \mathbb{E}_{t, s,a, s', X_0, X_1, X_t} \left[ \| X_1 - X_0 - \mathbb{E}_{(X_1, X_0) \mid s,a,s', X_t} \left[ X_1 - X_0 \right]\|^2 \right].
\end{align*}
In particular, when the paths of the linear interpolation $X_t$ do not intersect for any $s,a,s'$, we have $\mathbb{E}_{t, s,a,s', X_0, X_1, X_t} \left[ \| X_1 - X_0 - \mathbb{E}_{(X_1, X_0) \mid s,a,s', X_t} \left[ X_1 - X_0 \right]\|^2 \right] = 0$ and $\sigma_{\text{\ctdfshort}} = \sigma_{\text{\tdfshort}}$.
\end{theorem}
\end{metaframe}
\begin{proof}
    The first two statements can be checked by repeating the proof of \autoref{th:var-tdf-vtdf} with conditional paths $p_{t|0,1}$ and vector fields $u_{t|0,1}$. Let us thus prove the second part. We know that the flow $\phi_t(x_0, x_1)$ that generates the the conditonal path $p_{t | 0, 1}(x | x_0, x_1) = \delta_{tx_1 + (1-t)x_0}(x)$ is $\phi_t(x_0, x_1) = tx_1 + (1-t)x_0$. Its associated vector field $u_{t|0,1}$ is thus
    \begin{align*}
        u_{t|0,1}(\phi_t(x_0, x_1) | x_0, x_1) = \frac{d}{dt} \phi_t(x_0, x_1) = x_1 - x_0.
    \end{align*}
    Theorefore, denoting $Y = (t, s, a)$, we can bound the second term in the decomposition of $\sigma_{\text{\ctdfshort}}$ as
    \begin{align*}
        & \mathbb{E}_{Y, X_t} \left[ \text{Trace}\left( \text{Cov}_{(X_0, X_1) \mid Y, X_t} \left[ \nabla_\theta v_t(X_t| s, a;\theta)^\top u_{t|1}(X_t \mid X_0, X_1)\right] \right)\right]
        \\ &= \mathbb{E}_{Y, X_t} \left[ \mathbb{E}_{X_0,X_1 | Y, X_t}\left[\Big\| \nabla_\theta v_t(X_t| s, a;\theta)^\top u_{t|0,1}(X_t \mid X_0, X_1) - \mathbb{E}_{X_0,X_1 | Y, X_t}\left[ \nabla_\theta v_t(X_t| s, a;\theta)^\top u_{t|0,1}(X_t \mid X_0, X_1) \right] \Big\|^2\right] \right]
        \\ &\leq \mathbb{E}_{Y, X_t} \left[ \|\nabla_\theta v_t(X_t| s, a;\theta)\|^2 \mathbb{E}_{X_0,X_1 | Y, X_t}\left[ \Big\| u_{t|0,1}(X_t \mid X_0, X_1) - \mathbb{E}_{X_0,X_1 | Y, X_t}\left[ u_{t|0,1}(X_t \mid X_0, X_1) \right] \Big\|^2\right] \right]
        \\ &= \mathbb{E}_{Y, X_t} \left[ \|\nabla_\theta v_t(X_t| s, a;\theta)\|^2 \mathbb{E}_{X_0,X_1 | Y, X_t}\left[ \Big\| X_0 - X_1 - \mathbb{E}_{X_0,X_1 | Y, X_t}\left[ X_1 - X_0 \right] \Big\|^2\right] \right]
        \\ &\leq \sup_{t, s,a,x} \|\nabla_\theta v_t(x| s, a;\theta)\|^2 \mathbb{E}_{Y, X_t} \left[ \mathbb{E}_{X_0,X_1 | Y, X_t}\left[ \Big\| X_0 - X_1 - \mathbb{E}_{X_0,X_1 | Y, X_t}\left[ X_1 - X_0 \right] \Big\|^2\right] \right].
    \end{align*}
    This proves the third statement.

    To this the last point, simply note that if the paths generating $X_t$ do not cross, then the distribution of $X_0,X_1 | Y, X_t$ is supported over a single couple $(X_0, X_1)$, which means that its variance is zero.
\end{proof}

\clearpage
\subsection{Transport Cost Analysis}\label{app:sec:transport}

\begin{metaframe}
\begin{theorem}
    Assume that $m^{(n)}_t(x \mid s, a) = \int p_{t|1}(x \mid x_1) m^{(n)}_{1}(x_1 \mid s, a) \mathrm{d}x_1$, where $p_{t|1}(\cdot \mid x_1) = \mathcal{N}(t x_1, (1-t)^2 I)$ is a Gaussian path. Then, the conditional paths\,\footnotemark built by {\ctdfshort} and {\tdfshort} to generate $m^{(n+1)}_1 = \mathcal{T}^\pi m^{(n)}_1$ induce a smaller transport cost than those built by {\vtdfshort}. Formally, for every $t,s,a$,
    \begin{equation*}
        \mathbb{E}_{t,s,a,s', X_0 \sim m_0, X_1 \sim (1-\gamma) \delta_{s'} + \gamma \delta_{\psi^{(n)}_1(X_0 | s', \pi(s'))} } \left[ \| X_1 - X_0\|^2 \right] \leq \mathbb{E}_{t,s,a,s', X_0 \sim m_0, X_1 \sim [\mathcal{T}^\pi m^{(n)}_1](\cdot\mid s,a)} \left[ \| X_1 - X_0\|^2\right].
    \end{equation*}
\end{theorem}
\end{metaframe}
\footnotetext{Recall that, given a marginal probability path $m^{(n)}_t(x \mid s, a)$, the conditional probability path built by {\ctdfshort} and {\tdfshort} to generate $\mathcal{T}^\pi m^{(n)}_1$ is a linear interpolation between noise $X_0 \sim m_0$ and $X_1 \sim (1-\gamma) \delta_{s'} + \gamma \psi^{(n)}_1(X_0 | s', \pi(s'))$, while the one built by {\vtdfshort} is a linear interpolation between noise $X_0 \sim m_0$ and a sample $X_1 \sim [\mathcal{T}^\pi m^{(n)}_1](\cdot\mid s,a)$ from the target distribution.}

\begin{proof}
    
The paths generated by {\ctdfshort} and {\tdfshort} induce the same transport cost since both algorithms connect the endpoints of the ODE path $m^{(n)}_t$ in the bootstrapped term. Hence,
\begin{align*}
& \mathbb{E}_{t,s,a,s', X_0 \sim m_0, X_1 \sim (1-\gamma) \delta_{s'} + \gamma \delta_{\psi^{(n)}_1(X_0 | s', \pi(s'))} } \left[ \| X_1 - X_0\|^2 \right] \\
& = (1-\gamma) \mathbb{E}_{t,s,a,s', X_0} \left[ \| s' - X_0\|^2\right] + \gamma  
\mathbb{E}_{t,s,a,s', X_0} \left[ \| \psi_1^{(n)}(X_0 \mid s', \pi(s')) - X_0\|^2\right] \\
& \stackrel{(a)}{=} (1-\gamma) \mathbb{E}_{t,s,a,s', X_0} \left[ \| s' - X_0\|^2\right] + \gamma  
\mathbb{E}_{t,s,a,s', X_0} \left[ \Big \| \int v_t^{(n)}(\psi_t^{(n)}(X_0 \mid s', \pi(s'))) \mathrm{d}t \Big \|^2\right] \\
& \stackrel{(b)}{\leq} (1-\gamma) \mathbb{E}_{t,s,a,s', X_0} \left[ \| s' - X_0\|^2\right] + \gamma  
\int  \mathbb{E}_{t,s,a,s', X_0} \left[ \Big \| v_t^{(n)}(\psi_t^{(n)}(X_0 \mid s', \pi(s'))) \Big \|^2\right]  \mathrm{d}t\\
& \stackrel{(c)}{=} (1-\gamma) \mathbb{E}_{t,s,a,s', X_0} \left[ \| s' - X_0\|^2\right] + \gamma  
\int  \mathbb{E}_{t,s,a,s', X_t \sim m^{(n)}_t(\cdot \mid s', \pi(s'))} \left[ \Big \| v_t^{(n)}(X_t \mid s', \pi(s')) \Big \|^2\right]  \mathrm{d}t \\
& \stackrel{(d)}{=} (1-\gamma) \mathbb{E}_{t,s,a,s', X_0} \left[ \| s' - X_0\|^2\right] + \gamma  
\int  \mathbb{E}_{t,s,a,s', X_t \sim m^{(n)}_t(\cdot \mid s', \pi(s'))} \left[ \Big \| \mathbb{E}_{X_1 \mid s', X_t} \left[ u_{t|1}(X_t | X_1)\right] \Big \|^2\right]  \mathrm{d}t \\
& \stackrel{(e)}{\leq} (1-\gamma) \mathbb{E}_{t,s,a,s', X_0} \left[ \| s' - X_0\|^2\right] + \gamma  
\int  \mathbb{E}_{t,s,a,s', X_t \sim m^{(n)}_t(\cdot \mid s', \pi(s'))} \left[ \mathbb{E}_{X_1 \mid s', X_t} \left[ \Big \| u_{t|1}(X_t | X_1) \Big \|^2 \right]  \right] \mathrm{d}t \\
& \stackrel{(f)}{=} (1-\gamma) \mathbb{E}_{t,s,a,s', X_0} \left[ \| s' - X_0\|^2\right] + \gamma  
\int  \mathbb{E}_{t,s,a,s', X_1 \sim m^{(n)}_1(\cdot \mid s', \pi(s')), X_t \sim p_{t|1}(\cdot \mid X_1)} \left[  \Big \| u_{t|1}(X_t | X_1) \Big \|^2 \right]  \mathrm{d}t \\
& \stackrel{(g)}{=} (1-\gamma) \mathbb{E}_{t,s,a,s', X_0} \left[ \| s' - X_0\|^2\right] + \gamma  
\int  \mathbb{E}_{t,s,a,s', X_1 \sim m^{(n)}_1(\cdot \mid s', \pi(s')), X_0} \left[  \Big \| u_{t|1}(tX_1 + (1-t) X_0 | X_1) \Big \|^2 \right]  \mathrm{d}t \\
& \stackrel{(h)}{=} (1-\gamma) \mathbb{E}_{t,s,a,s', X_0} \left[ \| s' - X_0\|^2\right] + \gamma  
\int  \mathbb{E}_{t,s,a,s', X_1 \sim m^{(n)}_1(\cdot \mid s', \pi(s')), X_0} \left[  \Big \| X_1 - X_0 \Big \|^2 \right]  \mathrm{d}t \\
& \stackrel{(i)}{=} (1-\gamma) \mathbb{E}_{t,s,a,s', X_0} \left[ \| s' - X_0\|^2\right] + \gamma   \mathbb{E}_{t,s,a,s', X_1 \sim m^{(n)}_1(\cdot \mid s', \pi(s')), X_0} \left[  \Big \| X_1 - X_0 \Big \|^2 \right]  \\
& \stackrel{(j)}{=} \mathbb{E}_{t,s,a,s', X_0 \sim m_0, X_1 \sim [\mathcal{T}^\pi m^{(n)}_1](\cdot\mid s,a)} \left[ \| X_1 - X_0\|^2\right],
\end{align*}
where (a) uses the definition of flow as integration of a vector field, (b) uses Cauchy-Schwarz inequality, (c) uses that $m_0 * \psi_t^{(n)}$ is the pushforward measure generating $m^{(n)}_t$, (d) defines $X_1 \mid x, s'  \sim \frac{p_{t|1}(x| X_1) m_1^{(n)}(X_1 | s', \pi(s'))}{m^{(n)}_t(x \mid s, a)}$ as the posterior distribution of $X_1$ given $x,s'$ and uses that $v_t^{(n)}$ is in marginal form by assumption, (e) uses Jensen's inequality, (f) uses the Tower property of expectations, (g) uses the definition of $p_{t|1}$ and the corresponding linear-interpolation flow, (h) uses the definition of $u_{t|1}$, (i) is trivial, and (j) simply combines the two terms using the definition of Bellman operator $\mathcal{T}^\pi$.
\end{proof}

\end{appendices}

\end{document}